\pdfoutput=1

\documentclass[]{paper}
\usepackage{helvet}

\usepackage[toc,page,header]{appendix}
\usepackage{titletoc}
\usepackage{minitoc}

\usepackage{algorithm}
\usepackage{algpseudocode}

\usepackage{wasysym}
\usepackage{fontawesome}

\usepackage{enumitem}

\usepackage{listings}

\usepackage{pgfplots}
\pgfplotsset{compat=1.18}
\usetikzlibrary{arrows.meta, positioning, calc, fit, backgrounds, shapes.geometric, decorations.pathmorphing}

\usepackage{arydshln}
\usepackage{adjustbox}
\usepackage{colortbl}
\usepackage{multirow}

\usepackage{xspace}
\usepackage{nicefrac}
\usepackage{float}
\usepackage{comment}
\usepackage{wrapfig}
\usepackage{pifont}
\usepackage{dsfont}

\usepackage{amsmath, amssymb}
\usepackage{amsthm}

\usepackage{tabularx}
\usepackage{xcolor}

\newcommand{\camyla}{\textbf{Camyla}\xspace}
\newcommand{\camylabench}{\textbf{CamylaBench}\xspace}
\newcommand{\camylanet}{\textbf{CamylaNet}\xspace}
\newcommand{\camylaD}{\textbf{Camyla}$_\text{D}$\xspace}
\newcommand{\camylaS}{\textbf{Camyla}$_\text{S}$\xspace}

\title{Camyla: Scaling Autonomous Research in Medical Image Segmentation}

\author{
    Yifan Gao\textsuperscript{1,2},
    Haoyue Li\textsuperscript{1},
    Feng Yuan\textsuperscript{1},
    Xin Gao\textsuperscript{*1},
    Weiran Huang\textsuperscript{*2,3},
    Xiaosong Wang\textsuperscript{*2,4}
}
\affiliation[1]{\mbox{University of Science and Technology of China}}
\affiliation[2]{\mbox{Shanghai Innovation Institute}}
\affiliation[3]{\mbox{Shanghai Jiao Tong University}}
\affiliation[4]{\mbox{Shanghai Artificial Intelligence Laboratory}}

\contribution{* Corresponding authors}

\abstract{
We present \camyla, a system for fully autonomous research within the scientific domain of medical image segmentation. \camyla transforms raw datasets into literature-grounded research proposals, executable experiments, and complete manuscripts without human intervention. Autonomous experimentation over long horizons poses three interrelated challenges: search effort drifts toward unpromising directions, knowledge from earlier trials degrades as context accumulates, and recovery from failures collapses into repetitive incremental fixes. To address these challenges, the system combines three coupled mechanisms: Quality-Weighted Branch Exploration for allocating effort across competing proposals, Layered Reflective Memory for retaining and compressing cross-trial knowledge at multiple granularities, and Divergent Diagnostic Feedback for diversifying recovery after underperforming trials. The system is evaluated on \camylabench, a contamination-free benchmark of 31 datasets constructed exclusively from 2025 publications, under a strict zero-intervention protocol across two independent runs within a total of 28 days on an 8-GPU cluster. Across the two runs, \camyla generates more than 2{,}700 novel model implementations and 40 complete manuscripts, and surpasses the strongest per-dataset baseline selected from 14 established architectures, including nnU-Net, on 22 and 18 of 31 datasets under identical training budgets, respectively (union: 24/31). Senior human reviewers score the generated manuscripts at the T1/T2 boundary of contemporary medical imaging journals. Relative to automated baselines, \camyla outperforms AutoML and NAS systems on aggregate segmentation performance and exceeds six open-ended research agents on both task completion and baseline-surpassing frequency. These results suggest that domain-scale autonomous research is achievable in medical image segmentation.
}

\contact{yifangao@mail.ustc.edu.cn, xingaosam@163.com, weiran.huang@outlook.com, wangxiaosong@pjlab.org.cn}
\coderepository{\url{https://github.com/yifangao112/camyla}}
\project{\url{https://yifangao112.github.io/camyla-page/}}

\begin{document}
\maketitle
\vspace{-6mm}

\begin{figure*}[!htbp]
\centering
\includegraphics[width=\linewidth,height=0.35\textheight,keepaspectratio]{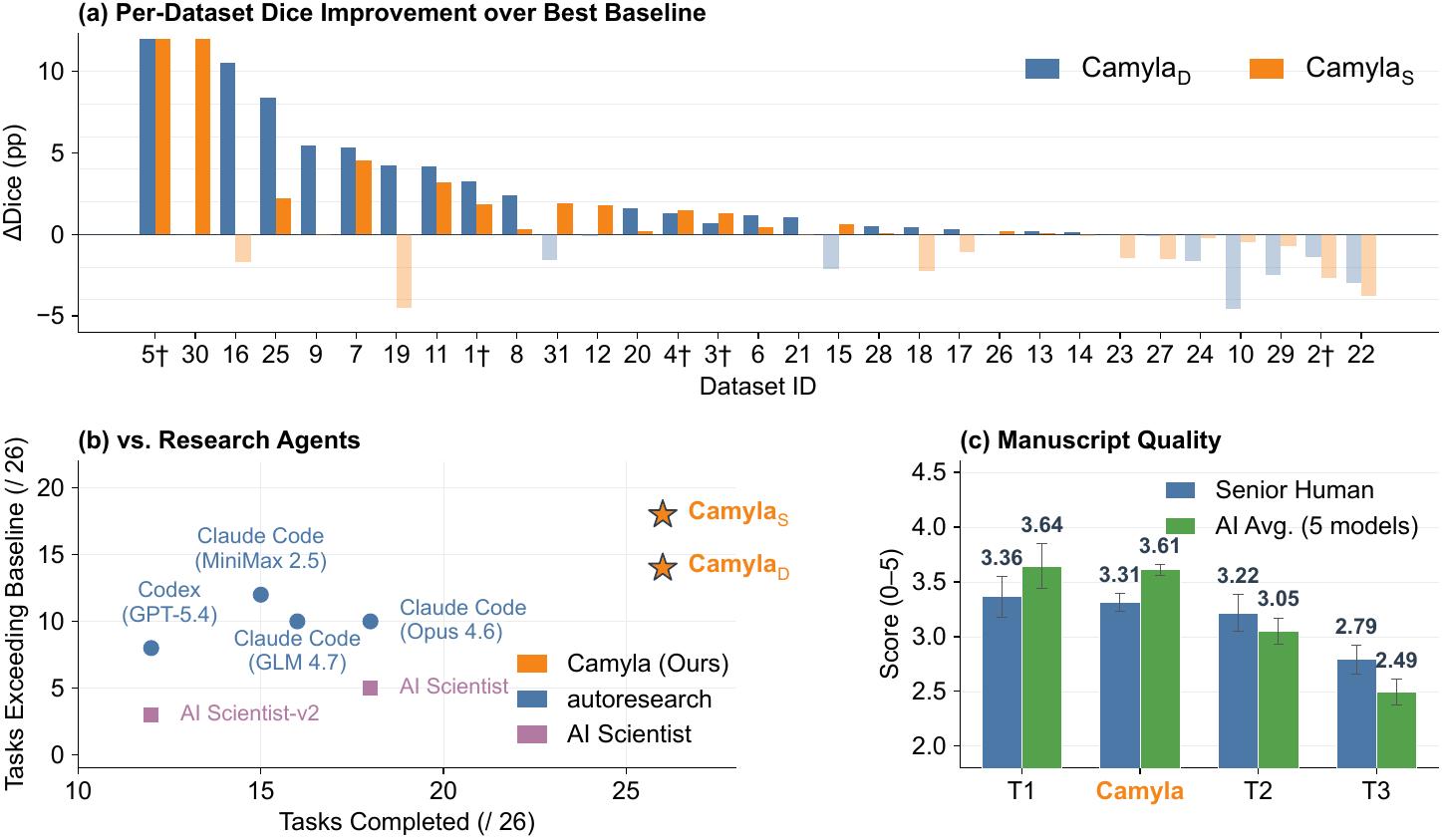}
\caption{Main results of \camyla across two independent runs on \camylabench. \textbf{(a)}~Per-dataset Dice improvement over the strongest baseline for both runs (\camylaD and \camylaS). Datasets marked with $\dagger$ are validation; the remainder are blind-test. \textbf{(b)}~Comparison with six open-ended research agents on 26 blind-test datasets: \camyla achieves 100\% task completion with zero proposal drift. \textbf{(c)}~Manuscript quality evaluation: senior human reviewers and AI evaluators both score \camyla's manuscripts at the T1/T2 boundary of contemporary medical image segmentation venues. Tier-defining journals: T1 = IEEE Transactions on Medical Imaging and Medical Image Analysis (2 venues); T2 = IEEE Journal of Biomedical and Health Informatics, \textit{et al.} (7 venues); T3 = Biomedical Physics \& Engineering Express, \textit{et al.} (9 venues).}
\label{fig:hero}
\end{figure*}

\section{Introduction}
\label{sec:intro}

Recent autonomous research systems have demonstrated promising capabilities on individual scientific tasks~\citep{lu2024aiscientist,yamada2025aiscientistv2,schmidgall2025agentlab}, but they have been validated at small experimental scale, typically one to three datasets per run. Whether such systems can operate reliably at domain scale, conducting hypothesis generation, experimentation, and paper writing across dozens of diverse tasks under a single protocol, remains an open question.

Medical image segmentation is a demanding scientific domain for this challenge: it offers standardized metrics, strong and diverse baselines, and a broad spectrum of tasks spanning different anatomical structures, imaging modalities, and clinical contexts. A system that performs effectively across such tasks must overcome three interrelated difficulties. First, search effort drifts toward unpromising directions when the system lacks a principled mechanism for reallocating experimental budget across competing proposals. Second, knowledge from earlier trials degrades as context accumulates over long horizons, causing the system to repeat past mistakes or lose hard-won insights. Third, recovery from failures collapses into repetitive incremental fixes when only a single corrective signal is available, trapping the system in local optima.

Existing automation approaches cover parts of this pipeline but not the whole. Neural architecture search~\citep{he2021dints,yu2020c2fnas} and AutoML methods~\citep{becktepe2025autonnunet,hutter2019automl} optimize models within fixed, human-designed search spaces and produce no transferable research insights. General-purpose research agents~\citep{karpathy2026autoresearch} can plan and edit code freely but lack the domain-specific infrastructure to enforce experimental rigor: in our evaluation, the best such agent completes only 69\% of assigned tasks, with 54\% exhibiting proposal drift. End-to-end research systems~\citep{lu2024aiscientist,yamada2025aiscientistv2} broaden the scope to idea generation, coding, and writing, but have not been tested at the scale needed to establish whether their discoveries generalize across diverse problem settings.

We introduce \camyla, an autonomous research system for fully autonomous, domain-scale research in medical image segmentation. \camyla transforms raw segmentation datasets into literature-grounded research proposals, iterative experiments, and complete manuscripts through a four-stage pipeline that operates without any human intervention. The system rests on two infrastructure components: \camylabench, a 31-dataset benchmark constructed exclusively from 2025 publications to eliminate data contamination, and \camylanet, a three-function programmatic workbench that wraps nnU-Net into an agent-friendly interface while exposing full architectural flexibility.

Three coupled mechanisms address these challenges directly. \textbf{Quality-Weighted Branch Exploration (QWBE)} counters search drift by modeling the multi-proposal search as a bandit problem with a risk-averse prior, concentrating experimental budget on promising research directions while dynamically creating new proposals when existing ones stagnate. \textbf{Layered Reflective Memory (LRM)} counters knowledge degradation by organizing experimental knowledge into trial-level, cycle-level, and global tiers, each progressively compressed, so that coding agents receive only decision-relevant information at the appropriate abstraction level. \textbf{Divergent Diagnostic Feedback (DDF)} counters incremental-fix traps by replacing single-point error corrections with five-suggestion diagnostic portfolios spanning architecture, hyperparameters, code fixes, and proposal-implementation gaps. The three mechanisms are mutually conditioning: DDF generates the diagnostic diversity that LRM compresses and QWBE exploits.

We evaluate \camyla under a strict zero-intervention protocol across two independent runs on all 31 \camylabench datasets. The two runs differ only in the language model used for idea generation; all other components remain identical. Under identical training budgets, the system surpasses the strongest per-dataset baseline selected from 14 established architectures, including nnU-Net, on 22 and 18 of 31 datasets, respectively, and produces 40 complete manuscripts that senior human reviewers score at the T1/T2 boundary of contemporary medical imaging journals. Relative to automated baselines, both runs outperform AutoML and NAS systems on aggregate Dice and substantially exceed all six open-ended research agents on both task completion and baseline-surpassing frequency.

In summary, this paper presents the following contributions:
\begin{itemize}
\item We introduce \camyla and its three coupled mechanisms for long-horizon research orchestration: QWBE for resource allocation, LRM for cross-trial knowledge retention, and DDF for recovery diversity after setbacks.
\item We present \camylabench, a 31-dataset benchmark with 26 blind-test datasets constructed exclusively from 2025 publications to eliminate data contamination.
\item We present \camylanet, a domain-specific research workbench that compresses dataset preparation, training, and evaluation into a three-function interface while preserving full architectural flexibility.
\item We release \textbf{CamylaTrace-232k}, the experimental-discovery trajectories from both runs, covering over 232,000 agent events, 2,865 coding sessions, and 1,343 model implementations.
\item Across two independent runs and comparisons against nine automated baselines, \camyla surpasses the strongest per-dataset baseline on 24 of 31 datasets and produces manuscripts scored at the T1/T2 boundary of medical imaging journals by senior human reviewers.
\end{itemize}

\section{Research Setting and Infrastructure}
\label{sec:infrastructure}

This section introduces the two foundational components that underpin \camyla: the benchmark on which the system is evaluated and the computational workbench through which it conducts experiments. Together, \camylabench and \camylanet define the research setting---what the system must accomplish and the tools it uses to do so.

\subsection{Research Task and Evaluation Protocol}
\label{sec:task}

The research task is defined as follows. Given a medical image segmentation dataset, the system must autonomously generate a complete research manuscript that includes a novel methodological contribution, comprehensive experimental evaluation, and a clear scientific narrative. The system operates under a zero-intervention protocol: once launched with a dataset identifier, it produces all outputs---research proposals, trained models, ablation studies, and a compiled manuscript---without any human input.

Each dataset is preprocessed and evaluated following the same protocol as nnU-Net~\citep{isensee2021nnu}. The primary metric is the mean Dice coefficient~\citep{milletari2016vnet} computed over all foreground classes. The secondary metric is the 95th-percentile Hausdorff distance (HD95)~\citep{taha2015metrics}, which captures boundary accuracy. A system is considered to have exceeded the baseline on a given dataset if the Dice improvement over the baseline exceeds 0.5 percentage points, or if the Dice difference is within 0.5 percentage points and the HD95 is strictly lower than the baseline HD95. This two-tier criterion prevents the system from claiming success based on negligible Dice fluctuations while still recognizing meaningful improvements in boundary quality when volumetric overlap is already near its ceiling.

\subsection{CamylaBench}
\label{sec:camylabench}

Evaluating an autonomous research system requires a benchmark that the system has never encountered during development and that no existing method has been specifically tuned on. Established segmentation benchmarks such as the Medical Segmentation Decathlon~\citep{antonelli2022msd} have been widely used for years, creating a risk of implicit data contamination: an LLM-based research agent may generate methods that exploit memorized knowledge about these benchmarks rather than discovering genuinely new solutions. To eliminate this risk, we construct \camylabench entirely from datasets that did not exist in the public domain until 2025.

We adopt an exhaustive, zero-filtering collection protocol. \camylabench is constructed from the complete set of medical image segmentation datasets published in \emph{Scientific Data} during 2025. We collected every article in this journal whose primary contribution is a segmentation dataset with publicly available images and annotations. No sampling, no difficulty-based filtering, and no modality restriction was applied beyond the requirement that the dataset must contain pixel-level segmentation labels. This protocol means that the benchmark composition is determined entirely by what the research community published in a given year, not by our expectations about which tasks the system can or cannot solve.

The resulting benchmark comprises 31 datasets spanning 12 anatomical regions and 10 imaging modalities, with training set sizes ranging from 24 to 10{,}662 cases and foreground class counts ranging from 1 to 10. Table~\ref{tab:camylabench} provides the complete dataset inventory, and the source publications are cited in full~\citep{ds900_gdmrict,ds901_mrebsa,ds902_pnpc,ds903_smrifb,ds904_lmdbm,ds905_amsmchtm,ds906_bonbid,ds907_btxrd,ds908_busuclm,ds909_cirrmri,ds910_cpaisd,ds911_denpar,ds912_dermaocta,ds913_endoscapes,ds914_fovea,ds915_fundusavseg,ds916_hrusmbt,ds918_lapgc,ds919_longciu,ds920_mslesseg,ds921_muglioma,ds922_nlstseg,ds923_oct5k,ds924_pedims,ds925_plccect,ds926_pwbalfc,ds927_seasis,ds928_ststooth,ds930_tom500,ds931_trusted,ds932_endomri}.

The 31 datasets span 12 anatomical regions (brain, abdomen, ophthalmology, head and neck, lung, dental, breast, skin, cytology, musculoskeletal, soft tissue, and gynecology) and a wide difficulty spectrum: baseline Dice ranges from 14.2\% (endoscopic instrument segmentation) to 96.2\% (dental panoramic radiography), with a median of 71.4\%. A detailed breakdown of anatomical coverage and task difficulty is provided in Appendix~\ref{app:dataset-diversity}. Among the 31 datasets, five (Datasets~1--5) were randomly designated as validation datasets used during system development; the remaining 26 were held out as blind-test datasets. All reported results distinguish between these two subsets.

\begin{table*}[!htbp]
    \centering
    \scriptsize
    \setlength{\tabcolsep}{2pt}
    \renewcommand{\arraystretch}{0.82}
    \caption{Complete inventory of the 31 datasets in \camylabench, sourced from \emph{Scientific Data} 2025. $\dagger$\,=\,validation. Cfg.\,=\,2D/3D. Cls.\,=\,foreground classes. Baseline\,=\,best of 14 pretrained architectures.}
    \label{tab:camylabench}
    \resizebox{\textwidth}{!}{%
    \begin{tabular}{r l l l r r c l r @{\hskip 5pt}!{\vrule width 0.3pt}@{\hskip 7pt} r l l l r r c l r}
    \toprule
    \textbf{ID} & \textbf{Dataset} & \textbf{Region} & \textbf{Modality} & \textbf{N} & \textbf{C} & \textbf{Cfg} & \textbf{Baseline} & \textbf{Dice}
    & \textbf{ID} & \textbf{Dataset} & \textbf{Region} & \textbf{Modality} & \textbf{N} & \textbf{C} & \textbf{Cfg} & \textbf{Baseline} & \textbf{Dice} \\
    \midrule
    1$^\dagger$ & GDMRI-CT      & Brain        & MRI        &    50 &  1 & 3D & STU-Net   & 70.6
    & 17 & Fundus-AVSeg  & Ophthal.     & Fundus     &   100 &  3 & 2D & U-Mamba   & 77.6 \\
    2$^\dagger$ & MRE-BSA       & Abdomen      & MRI        &   114 & 10 & 3D & U-Mamba   & 71.8
    & 18 & HRUS-MBT      & Soft Tiss.   & Ultrasound &    34 &  1 & 3D & nnU-Net   & 50.1 \\
    3$^\dagger$ & PNPC          & Head\&Neck   & MRI        &   277 &  1 & 3D & nnU-Net   & 60.6
    & 19 & LapGC-KVAD    & Abdomen      & Laparoscopy & 1252 &  4 & 2D & U-Mamba   & 52.8 \\
    4$^\dagger$ & AMSMC-HTM     & Head\&Neck   & MRI        &    24 &  4 & 3D & nnU-Net   & 81.4
    & 20 & LongCIU       & Lung         & CT         &    90 &  2 & 2D & nnU-Net   & 73.5 \\
    5$^\dagger$ & NLSTseg       & Lung         & CT         &   604 &  1 & 3D & nnU-Net   & 23.2
    & 21 & MSLesSeg      & Brain        & MRI        &   115 &  1 & 3D & U-Mamba   & 69.6 \\
    6  & SMRI-FB       & Brain        & MRI        &    70 &  8 & 3D & MedNeXt   & 85.3
    & 22 & MU-Glioma     & Brain        & MRI        &   594 &  4 & 3D & nnU-Net   & 71.6 \\
    7  & LMD-BM        & Brain        & MRI        &    44 &  3 & 3D & nnU-Net   & 46.2
    & 23 & OCT5k         & Ophthal.     & OCT        & 1672  &  5 & 2D & U-Mamba   & 65.4 \\
    8  & BONBID2023    & Brain        & MRI        &    89 &  1 & 3D & MedNeXt   & 55.9
    & 24 & PediMS        & Brain        & MRI        &    28 &  1 & 3D & nnU-Net   & 79.5 \\
    9  & BTXRD         & Musculoskel. & X-ray      & 1867  &  2 & 2D & UTNet     & 44.5
    & 25 & PLC-CECT      & Abdomen      & CT         &   278 &  1 & 3D & U-Net++   & 53.2 \\
    10 & BUS-UCLM      & Breast       & Ultrasound &   670 &  1 & 2D & nnU-Net   & 36.7
    & 26 & PW-BALFC      & Cytology     & Microscopy & 2105  &  1 & 2D & nnU-Net   & 76.7 \\
    11 & CirrMRI600+   & Abdomen      & MRI        &   373 &  1 & 3D & nnU-Net   & 81.8
    & 27 & SEA-SIS       & Abdomen      & Endoscopy  & 10662 &  6 & 2D & nnU-Net   & 14.2 \\
    12 & CPAISD        & Brain        & CT         &   112 &  1 & 3D & U-Mamba   & 27.3
    & 28 & STS-Tooth     & Dental       & X-ray      &   850 &  1 & 2D & U-Mamba   & 93.7 \\
    13 & DenPAR        & Dental       & X-ray      &   800 &  1 & 2D & nnU-Net   & 96.2
    & 29 & TOM500        & Ophthal.     & MRI        &   500 &  9 & 3D & U-Mamba   & 92.2 \\
    14 & DERMA-OCTA    & Skin         & OCTA       &   331 &  1 & 2D & U-Mamba   & 83.9
    & 30 & TRUSTED       & Abdomen      & Ultrasound &    59 &  1 & 3D & U-Mamba   & 62.1 \\
    15 & Endoscapes    & Abdomen      & Laparoscopy &  343 &  6 & 2D & nnU-Net   & 47.5
    & 31 & UT-EndoMRI    & Gynecology   & MRI        &    57 &  1 & 3D & U-Mamba   & 75.1 \\
    16 & FOVEA         & Ophthal.     & Fundus     &    40 &  1 & 2D & UKAN      & 71.4
    &    &               &              &            &       &    &    &           &      \\
    \bottomrule
    \end{tabular}%
    }
\end{table*}

\subsection{CamylaNet}
\label{sec:camylanet}

An autonomous research system requires a computational workbench that satisfies two competing constraints: it must encapsulate the full experimental lifecycle so that an LLM agent can run a complete experiment through a small number of function calls, and it must expose sufficient architectural flexibility so that the agent can implement genuinely novel designs rather than merely selecting from a fixed model menu. \camylanet is a domain-specific instantiation of this workbench pattern for medical image segmentation, wrapping nnU-Net~v2~\citep{isensee2021nnu} into a three-function programmatic interface while preserving full control over the network architecture. Analogous substrates could be constructed for other domains with established training and evaluation pipelines; \camylanet's role is to expose a minimal, agent-executable interface over a standardized experimental stack.

The three entry points correspond to the three stages of a segmentation experiment:
\begin{equation}
\texttt{plan\_and\_preprocess}(d, \mathcal{C}) \;\rightarrow\; \pi,
\label{eq:preprocess}
\end{equation}
\begin{equation}
\texttt{training\_network}(d, c, \mathcal{T}, \pi) \;\rightarrow\; (\mathcal{R}, \ell),
\label{eq:train}
\end{equation}
\begin{equation}
\texttt{evaluate}(d, \mathcal{R}) \;\rightarrow\; \mathcal{M},
\label{eq:eval}
\end{equation}
where $d$ is a dataset identifier, $\mathcal{C}$ is a set of target configurations (2D, 3D full-resolution, or both), $\pi$ is a plans object that records the fingerprint-derived preprocessing recipe, $\mathcal{T}$ is a trainer class, $c \in \mathcal{C}$ is the selected configuration, $\mathcal{R}$ is the result folder containing checkpoints and predictions, $\ell$ is a training log, and $\mathcal{M}$ is a dictionary of evaluation metrics including Dice and HD95. Internally, \texttt{plan\_and\_preprocess} chains fingerprint extraction, experiment planning, and data preprocessing into a single call. \texttt{training\_network} instantiates the trainer, configures the optimizer and learning rate schedule, executes the training loop with mixed-precision gradient scaling, and returns the output path. \texttt{evaluate} runs inference, computes per-case and aggregate metrics, and writes a structured summary. The entire pipeline runs in isolated subprocesses to prevent state leakage between successive experiments.

\paragraph{Architecture Extension Interface.}
The key design decision that makes \camylanet suitable for autonomous architectural research is the separation between the training infrastructure and the network definition. A new architecture is introduced by subclassing a base trainer and overriding a single static method, \texttt{build\_network\_architecture}, which receives the number of input channels, the number of output classes, and a dictionary of architecture parameters derived from the dataset fingerprint. The method returns an \texttt{nn.Module} that maps an input tensor to a segmentation logit map. All other components---data loading, augmentation, loss computation (soft Dice plus cross-entropy), learning rate scheduling (polynomial decay), checkpointing, and validation---are inherited from the base trainer. This design means that an LLM agent can implement an arbitrary PyTorch network, wrap it in a six-line trainer subclass, and obtain a fully functional experiment with the same preprocessing, augmentation, and evaluation protocol used by every other architecture in the system. The separation between infrastructure and network definition is the core abstraction that makes this workbench pattern transferable: the agent-facing contract (override one method, return an \texttt{nn.Module}) is independent of the underlying training framework.

\paragraph{One-Epoch Verification.}
A common failure mode in LLM-generated code is that the network definition contains shape mismatches, numerical instabilities, or memory overflows that only manifest at runtime. \camylanet provides a dedicated one-epoch verification function that executes a single training epoch with full data loading, forward and backward passes, and validation inference. Passing this check guarantees that the data pipeline, the network forward pass, the loss computation, the gradient update, and the inference export all execute without error. Every candidate architecture is validated through this function before committing to full training, reducing wasted computation from implementation errors.

\paragraph{Precomputed Baseline Bank.}
To establish a strong per-dataset reference point, we pretrain a bank of 14 baseline architectures spanning convolutional, transformer-based, and state-space model families (full list in Appendix~\ref{app:baseline-bank-details}). For each dataset, every compatible architecture is trained and the one achieving the highest Dice is selected as the baseline. This per-dataset selection ensures that the baseline is the strongest available model for each task: nnU-Net~\citep{isensee2021nnu} is the best baseline on 16 of 31 datasets, U-Mamba~\citep{ma2024umamba} on 10, and the remaining 5 are won by other architectures.

\section{The Camyla System}
\label{sec:system}

\begin{figure*}[!htbp]\centering
\includegraphics[width=\linewidth,height=0.45\textheight,keepaspectratio]{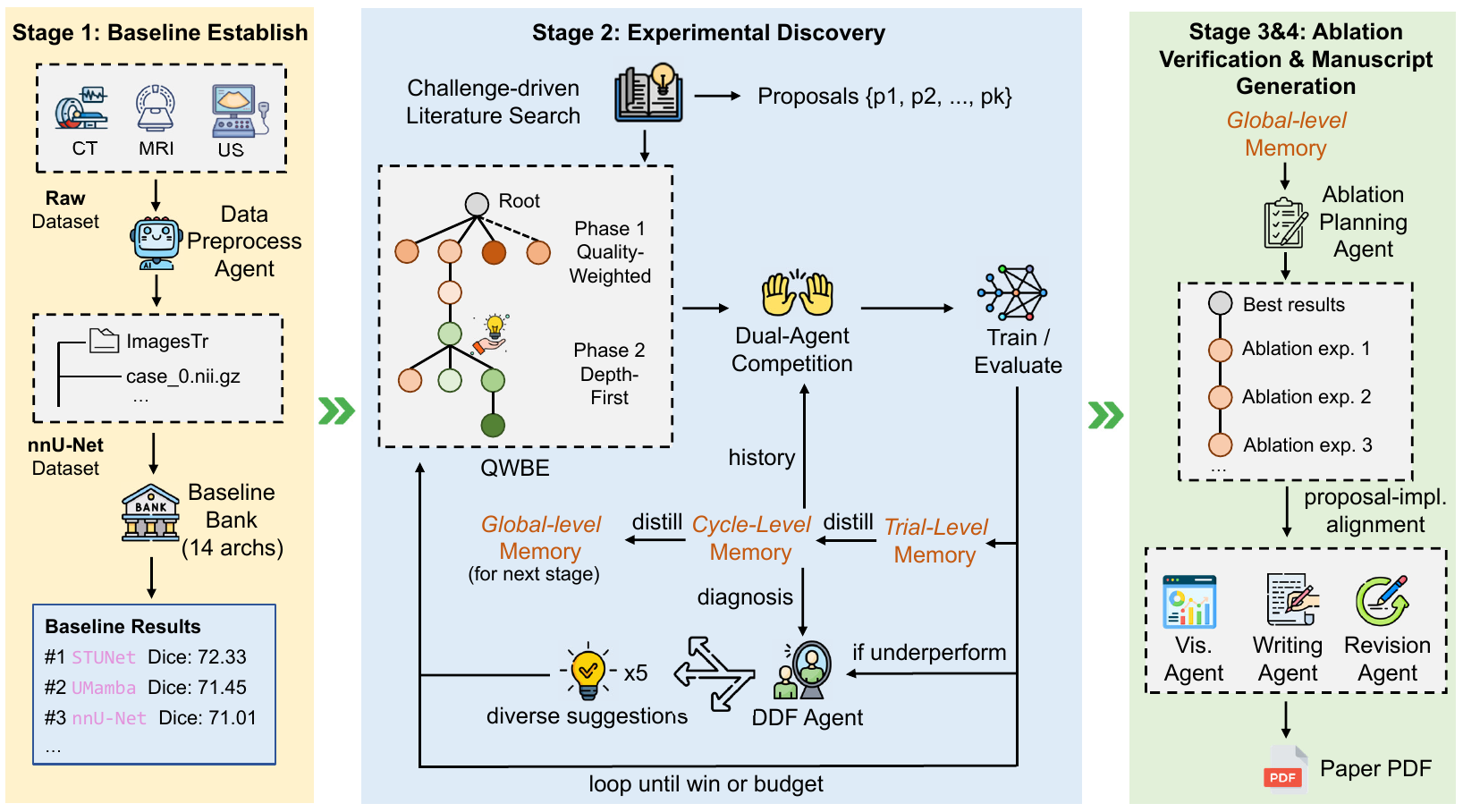}
\caption{Architecture of the \camyla system. The four-stage pipeline transforms a raw segmentation dataset into a complete research manuscript. Stage~1 establishes baselines via the precomputed baseline bank in \camylanet. Stage~2 executes experimental discovery through literature-grounded proposal generation and three coupled mechanisms for long-horizon experimentation: Quality-Weighted Branch Exploration (QWBE), Layered Reflective Memory (LRM), and Divergent Diagnostic Feedback (DDF). Stage~3 conducts ablation verification of the winning method. Stage~4 synthesizes the final manuscript through a multi-agent writing pipeline.}
\label{fig:overview}
\end{figure*}

The \camyla system transforms a raw segmentation dataset into a complete research manuscript through a four-stage pipeline: baseline establishment, experimental discovery, ablation verification, and manuscript generation. Each stage produces artifacts that feed the next, forming a closed loop from data ingestion to paper output. The entire pipeline executes without human intervention once the dataset identifier is provided.

\subsection{Overview}
\label{sec:overview}

Algorithm~\ref{alg:camyla} summarizes the end-to-end procedure. In Stage~1, the system loads the target dataset into \camylanet (\S\ref{sec:camylanet}), applies fingerprint-based preprocessing, and retrieves the precomputed results of 14 baseline architectures from the baseline bank. The architecture achieving the highest Dice on the evaluation fold is selected as the reference baseline $b^*$, and a structured summary of all baseline results is injected into the agent context so that subsequent stages can condition on baseline strengths and weaknesses. In Stage~2, the system generates a set of literature-grounded research proposals (\S\ref{sec:proposal}) and explores them through an iterative experimental search. At each iteration, Quality-Weighted Branch Exploration (\S\ref{sec:planning}) selects which proposal branch to expand, a pair of competing coding agents implements the next modification, and the resulting model is trained and evaluated. Layered Reflective Memory (\S\ref{sec:memory}) maintains a compressed record of all prior trials across proposals, while Divergent Diagnostic Feedback (\S\ref{sec:diagnostic}) generates categorically diverse improvement suggestions after each underperforming trial. Stage~2 terminates when a trial exceeds $b^*$ or the computational budget is exhausted. If a winning proposal is found, Stage~3 conducts ablation studies by systematically removing each proposed module and recording the resulting performance drop. Finally, Stage~4 assembles the research proposal, experimental results, and ablation evidence into a manuscript through a multi-agent writing pipeline (\S\ref{sec:manuscripts}). Stage-transition rules, subagent competition, and the ablation construction procedure are detailed in Appendix~\ref{app:pipeline-stage-details}.

Proposal generation (\S\ref{sec:proposal}) determines \emph{what} to investigate by grounding ideas in the recent literature. The remaining three components form a coupled mechanism for long-horizon experimentation: QWBE (\S\ref{sec:planning}) determines \emph{where} to allocate the finite experimental budget across competing proposals, LRM (\S\ref{sec:memory}) determines \emph{what} experimental knowledge is preserved at each decision point, and DDF (\S\ref{sec:diagnostic}) determines \emph{how} the system recovers and diversifies after setbacks. The ablation study in Section~\ref{sec:ablation} confirms that these three mechanisms are complementary: removing any one degrades either search efficiency, end-state quality, or recovery from failure.

\begin{algorithm}[!htbp]
\caption{The \camyla Pipeline}
\label{alg:camyla}
\begin{algorithmic}[1]
\Require Dataset identifier $d$, proposal budget $K$, iteration budget $N$ per proposal
\Ensure Manuscript PDF, trained model checkpoint

\Statex \textbf{Stage 1: Baseline Establishment}
\State Preprocess $d$ via \camylanet; extract fingerprint $\pi$
\State $b^* \gets \arg\max_{a \in \text{BaselineBank}} \mathrm{Dice}(d, a, \pi)$

\Statex \textbf{Stage 2: Experimental Discovery}
\State Generate proposals $\{P_1, \ldots, P_K\}$ from literature search \Comment{\S\ref{sec:proposal}}
\State Initialize global memory $M \gets \emptyset$
\For{each QWBE selection step} \Comment{\S\ref{sec:planning}}
    \State Select branch $b_i$ (or create new branch) via Eq.~\eqref{eq:puct-score}
    \State Two coding agents independently implement next modification
    \State Train and evaluate; record result in memory (\S\ref{sec:memory})
    \If{trial underperforms}
        \State Generate diagnostic report via DDF (\S\ref{sec:diagnostic})
    \EndIf
    \If{any trial exceeds $b^*$}
        \State \textbf{break} \Comment{Early stop}
    \EndIf
\EndFor

\Statex \textbf{Stage 3: Ablation Verification}
\For{each module $m$ in winning proposal}
    \State Train and evaluate model with $m$ removed
\EndFor

\Statex \textbf{Stage 4: Manuscript Generation}
\State Assemble evidence and generate paper (\S\ref{sec:manuscripts})
\end{algorithmic}
\end{algorithm}

\subsection{Literature-Grounded Proposal Generation}
\label{sec:proposal}

The starting point of every experimental campaign is a set of research proposals that specify what architectural innovations to pursue. A na\"ive approach---prompting a language model directly for improvement ideas---tends to produce generic suggestions that recombine well-known modules without grounding in the current state of the literature. We address this with a multi-agent pipeline that transforms a raw dataset description into a ranked set of self-contained research proposals through two phases.

\paragraph{Phase 1: Challenge Discovery.}
The pipeline constructs a search query from the dataset metadata (target anatomy, imaging modality, and task type) and retrieves recent papers from multiple academic databases, including arXiv, OpenAlex~\citep{priem2022openalex}, and PubMed. A challenge discovery agent analyzes the retrieved abstracts to extract concrete architectural challenges that remain unresolved. This extraction is repeated over multiple iterations: after each round, the accumulated challenge memory steers the next search toward underexplored directions. The collected challenges are consolidated into a small set of distinct research themes (typically three), each pairing an architectural challenge with a core research direction.

\paragraph{Phase 2: Theme-Specific Literature Search and Proposal Tournament.}
For each theme, the system conducts an independent literature search. A review agent retrieves and analyzes candidate papers; a citation network agent expands the keyword pool by examining references; and a method extraction agent produces structured summaries of architectural innovations from the full text. Each theme's search begins with a clean state so that the method libraries for different themes reflect genuinely different regions of the literature.

Given the literature review, the system generates proposals through an iterative best-of-$N$ tournament. In each round, three generator agents with distinct reasoning profiles (creative, rigorous, and domain-specific) independently produce a candidate proposal. An assessment agent evaluates them along six dimensions---coherence, credibility, verifiability, novelty, alignment, and modularity---and the winner is selected. Accepted proposals are added to a negative constraint list that enforces diversity across subsequent rounds. Each proposal specifies a title, a literature-grounded motivation, proposed modules with mathematical formulations, an integration plan, and expected contributions.

\paragraph{Proposal Refinement on Failure.}
When a proposal fails during experimentation (e.g., the generated code does not compile or the model diverges), a refinement step produces a simplified variant. The strategy is deliberately conservative: it may remove problematic modules or reduce complexity, but does not alter the core research direction.

\subsection{Quality-Weighted Branch Exploration}
\label{sec:planning}

Stage~2 explores a sequence of distinct research proposals, each forming an independent experimental subtree. The fundamental challenge is an exploration--exploitation tradeoff~\citep{auer2002ucb}: the system must decide at every step whether to broaden its search by trying a new proposal or to deepen its investment in a partially explored one. The value of each proposal branch is unknown at inception and must be estimated incrementally from noisy trial results, while the per-trial budget is finite. Two na\"ive policies fail in complementary ways. Pure exploitation---a greedy best-first policy---collapses onto whichever proposal shows the earliest improvement, abandoning potentially superior alternatives after insufficient evaluation. Pure exploration---uniform allocation across all branches---squanders resources on directions whose empirical quality consistently falls below the baseline, never concentrating enough effort to refine a promising idea into a genuine improvement.

We address this with \textbf{Quality-Weighted Branch Exploration (QWBE)}, a hierarchical search strategy that dynamically balances exploration and exploitation through quality-modulated resource allocation. The key insight is that empirical quality measurements from prior trials within a branch should \emph{modulate} the exploration bonus itself: a branch that has consistently underperformed the baseline warrants only sparse continued exploration, while a branch showing moderate improvement deserves aggressive deepening. QWBE operationalizes this through a risk-averse prior that suppresses the exploration budget for low-quality arms, combined with a two-phase switching rule that transitions from diversified exploration to focused exploitation once a winning direction is identified.

\paragraph{Quality-Weighted Selection.}
The selection rule builds on PUCT (Predictor + Upper Confidence bound applied to Trees)~\citep{rosin2011puct}, a variant of UCB that modulates the exploration bonus with a prior over arm quality. In AlphaGo Zero~\citep{silver2017alphagozero}, the prior is supplied by a policy network trained from self-play; in QWBE, we replace it with a quality-dependent function derived from empirical trial performance, so that the exploration bonus itself is suppressed for consistently underperforming branches. Specifically, at each selection step QWBE scores each existing branch $b_i$ as:
\begin{equation}
    \mathrm{Score}_i = Q_i + c_{\mathrm{puct}} \cdot P(Q_i) \cdot \frac{\sqrt{N_{\mathrm{total}}}}{1 + N_i},
    \label{eq:puct-score}
\end{equation}
where $Q_i \in [-1, +1]$ is the mean normalized quality of branch $b_i$, $N_i$ is the number of trials allocated to that branch, $N_{\mathrm{total}} = K + \sum_{j=1}^{K} N_j$ incorporates both the branch count and all expansions, and $c_{\mathrm{puct}}$ is the exploration coefficient (default 1.5). The term $P(Q_i)$ is a risk-averse prior:
\begin{equation}
    P(Q_i) = \max\!\left(0,\; 1 + Q_i\right)^{p},
    \label{eq:prior}
\end{equation}
with $p = 3$ by default. This prior scales the exploration bonus as a strongly nonlinear function of empirical quality: a branch with $Q_i = -1$ receives zero exploration pressure, effectively suspending further investment. A branch at baseline quality ($Q_i = 0$) retains unit prior, and one with moderate improvement ($Q_i = 0.5$) receives a prior of $(1.5)^3 = 3.375$. The procedure for normalizing raw evaluation metrics into $Q_i \in [-1,+1]$, including the handling of execution errors and leaf selection within a branch, is detailed in Appendix~\ref{app:qwbe-quality-normalization}. The system also maintains a virtual new-branch action with score:
\begin{equation}
    \mathrm{Score}_{\mathrm{new}} = c_{\mathrm{puct}} \cdot \frac{\sqrt{N_{\mathrm{total}}}}{1 + K}.
    \label{eq:score-new}
\end{equation}
The denominator $1 + K$ penalizes new-branch creation as existing branches accumulate, so the system diversifies aggressively early and converges toward deepening promising directions as the search progresses.

\paragraph{Phase Transition to Depth-First Exploitation.}
QWBE operates in two phases. During Phase~1, the PUCT-based scoring rule governs all selection decisions, allocating trials across proposals in proportion to quality-modulated exploration bonuses. As soon as any trial achieves a metric strictly above the baseline $m_0$, the system transitions to Phase~2: the multi-arm selection rule is suspended, and the globally best-performing node is selected for expansion at every subsequent step. This depth-first exploitation strategy concentrates resources on refining the most successful direction once it has been identified.

\subsection{Layered Reflective Memory}
\label{sec:memory}

Drawing on ideas from reflective agent architectures~\citep{shinn2023reflexion}, a key challenge in deploying LLM agents for iterative experimentation is \emph{context contamination}: as the number of trials grows, raw execution logs---spanning debugging traces, compiler errors, training curves, and diagnostic outputs---accumulate rapidly and pollute the agent's context window. Worse, heterogeneous failure modes become entangled: a low-level syntax error from a prior trial may sit alongside a high-level conclusion that an entire research direction is unviable, forcing the agent to spend capacity disentangling signal from noise. This mirrors how human researchers maintain separate mental registers for different granularities of experimental knowledge---a researcher does not re-read every failed training log before deciding the next architectural modification, but instead consults a compressed understanding of what has been tried and what has worked.

Motivated by this observation, we propose \textbf{Layered Reflective Memory (LRM)}, a three-tier memory architecture that isolates, compresses, and selectively relays experimental knowledge. The core design principle is that each coding agent receives only the information necessary for its current decision, structured at the appropriate level of abstraction.

\paragraph{Trial-Level Memory.}
Upon completion of each trial, a reflective summarization step distills the result into a compact modification record: a 2--3 sentence summary describing what was changed and why the outcome differed from the parent trial. The raw execution logs are then discarded from subsequent agent contexts, retaining only this compressed delta.

\paragraph{Cycle-Level Memory.}
Each research cycle (corresponding to one proposal) maintains a structured experimental history assembled from the modification records and evaluation metrics of all trials within it. Each trial entry contains its outcome status (baseline, success, underperforming, or error), quantitative metrics, the modification record, and a truncated diagnostic analysis. This structured format separates implementation-level failures from scientific-level conclusions, enabling the agent to distinguish between ``this approach crashed due to a fixable bug'' and ``this approach was correctly implemented but does not improve over the baseline.''

Table~\ref{tab:lrm-example} illustrates the three-layer interaction with a curated 12-row excerpt from a complete experimental trajectory on optic disc segmentation (Dataset~16). Across three research cycles, the system improves from a baseline Dice of 0.7142 to 0.8191 (+10.5 Dice points) as global-level insights accumulate and guide subsequent cycles. The complete 33-trial record---including the losing agent's Dice score at each trial and the cross-cycle summary---is provided in Appendix~\ref{app:full-trajectory-dataset16}.

\begin{table}[!htbp]
\centering
\caption{Complete experimental trajectory across three research cycles on optic disc segmentation (Dataset~16, baseline UKAN Dice = 0.7142). Each cycle explores a different research proposal; the best artifact and a compressed global memory are relayed to the next cycle. Trials marked with $\star$ achieved the cycle-best result. The system progressively improves from 0.7142 to 0.8191 (+10.5 Dice points) as global-level insights accumulate and guide subsequent cycles.}
\label{tab:lrm-example}
\small
\renewcommand{\arraystretch}{1.12}
\setlength{\tabcolsep}{4pt}
\begin{tabular}{@{}cclccp{5.8cm}@{}}
\toprule
\textbf{Cycle} & \textbf{Trial} & \textbf{Dice}$\uparrow$ & \textbf{Status} & \textbf{Diagnostic} & \textbf{Modification Record (abridged)} \\
\midrule
\multicolumn{6}{@{}l}{\textit{Cycle 1: HCP-Net --- Hierarchical Token Diffusion replacing skip connections}} \\
\midrule
 1 & 1  & 0.7142          & Baseline   & ---                      & Precomputed UKAN baseline. \\
 1 & 2  & 0.6835          & Underperf. & impl.\ shortcut  & HTD with cross-attention; hierarchy embedding omitted. \\
 1 & 3$\star$  & \textbf{0.7682} & Success    & ---               & HTD at 64${\times}$64; learnable hierarchy embeddings as query bias. \\
 1 & 6  & 0.7610          & Success    & ---                      & Reverted to query-bias (guided by Trial~3 in cycle memory). \\
 1 & 7  & 0.6976          & Underperf. & regression      & Added Sobel edge-aware refinement; hurt Dice. \\
\midrule
\multicolumn{6}{@{}l}{\textit{Cycle 2: BAFNet --- Boundary-Guided Instance Normalization (seeded from Cycle 1 best)}} \\
\midrule
 2 & 2  & 0.5653          & Underperf. & impl.\ shortcut  & Boundary normalization reduced to scalar; fusion simplified. \\
 2 & 4  & 0.7386          & Success    & ---                      & Added residual connection + dropout for gradient flow. \\
 2 & 8$\star$  & \textbf{0.7829} & Success    & ---               & Fixed NaN loss ($\epsilon$: $10^{-5}{\to}10^{-8}$); numerical stability was the key. \\
 2 & 9  & 0.7413          & Underperf. & regression      & Added multi-scale detection; regression. \\
\midrule
\multicolumn{6}{@{}l}{\textit{Cycle 3: ARTNet --- Adaptive Resolution Tokenization (seeded from Cycle 2 best)}} \\
\midrule
 3 & 2  & 0.6095          & Underperf. & impl.\ shortcut  & Adaptive tokenization reduced to fixed-grid pooling. \\
 3 & 5  & 0.7988          & Success    & ---                      & Gradient-magnitude complexity estimator; adaptive tokenization working. \\
 3 & 6$\star$  & \textbf{0.8191} & Success    & ---               & Minor fix; architecture from Trial~5 preserved. Overall best. \\
 3 & 10 & 0.7033          & Underperf. & regression      & Added boundary refinement + increased channels; both hurt. \\
\bottomrule
\end{tabular}
\end{table}

\paragraph{Global Memory.}
When a research cycle completes, its structured history is summarized into a scientific digest that extracts key findings, identifies cross-cutting insights, and discards implementation minutiae. The global memory is then updated by merging the new digest with the accumulated narrative. This progressive compression ensures that the global memory grows sublinearly with the number of cycles. The global memory is injected into the context of each new cycle's agents, providing high-level awareness of which directions proved fruitful, which were dead ends, and which findings may transfer---without burdening them with trial-level details from prior cycles. When a cycle concludes, its best-performing artifact is propagated forward as the seed for the next cycle, accompanied by the updated global memory. This dual relay---artifact plus compressed narrative---ensures continuity of both code and scientific understanding.

\subsection{Divergent Diagnostic Feedback}
\label{sec:diagnostic}

A common failure pattern in LLM-based iterative experimentation is \emph{feedback convergence}: when a trial underperforms, conventional systems generate a single-point diagnosis (e.g., ``the learning rate is too high'') and feed it back as a prescriptive fix. This creates a self-reinforcing loop in which the agent repeatedly pursues the same narrow correction, effectively reducing the search tree to a single chain. The problem is compounded by two behavioral tendencies of LLM agents. First, \emph{premature commitment}: once a plausible explanation is articulated, subsequent iterations anchor on that explanation even when the evidence is ambiguous. Second, \emph{regression to simplicity}: after a sequence of failed trials, agents tend to fall back on trivial solutions---replacing sophisticated proposed modules with standard convolutions, reverting to default hyperparameters, or abandoning the research direction entirely in favor of minimal modifications that merely ensure the code runs. In automated scientific experimentation, where the true cause of underperformance may be a subtle interaction among architecture, hyperparameters, and implementation fidelity, these tendencies dramatically narrow the explored solution space and trap the system in local optima.

We introduce \textbf{Divergent Diagnostic Feedback (DDF)}, a mechanism that counteracts both tendencies by replacing single-point prescriptions with structured, multi-dimensional diagnostic reports. The central idea is that after each underperforming trial, a dedicated diagnostic agent produces not one corrective instruction, but a portfolio of five categorically distinct improvement suggestions---spanning architecture modifications, hyperparameter adjustments, implementation bug fixes, and proposal--implementation gaps. By presenting the agent with multiple qualitatively different recovery paths simultaneously, DDF prevents the search from collapsing onto a single correction trajectory and actively broadens the agent's horizon after setbacks, enabling it to escape local optima through directions it would not have considered under single-point feedback.

\paragraph{Diagnostic Procedure.}
When a trial underperforms, a dedicated diagnostic agent analyzes the research proposal, the current implementation, the cycle-level history, and the performance gap relative to the best-known result. It produces a structured report containing a reasoning trace that diagnoses the root cause and five categorically distinct improvement suggestions. Each suggestion specifies a category, an actionable description, and an expected impact assessment.

The category taxonomy spans four dimensions that together cover the space of possible interventions: \emph{architecture} (structural changes to the network), \emph{hyperparameter} (tuning numerical settings), \emph{code fix} (correcting implementation errors), and \emph{proposal--implementation gap} (identifying components described in the proposal but missing or oversimplified in the actual code). The diagnostic agent distributes its five suggestions across multiple categories, with at least one suggestion required to audit the proposal--implementation gap.

\paragraph{Proposal--Implementation Gap Audit.}
The gap audit addresses a particularly important failure mode: implementation shortcuts. When faced with a complex proposal, the coding agent may substitute a sophisticated module with a trivial placeholder (e.g., replacing a proposed adaptive tokenization mechanism with a fixed-grid pooling operation). Standard execution feedback---which only reports training loss and evaluation metrics---cannot distinguish between ``the proposed approach does not work'' and ``the proposed approach was never actually implemented.'' The gap audit forces the diagnostic agent to explicitly compare the proposal against the implementation and identify such discrepancies.

\paragraph{Dual-Mode Operation.}
DDF operates in two modes depending on the trial outcome. When a trial underperforms, it produces the five-suggestion failure analysis described above. When a trial exceeds the current best, it switches to an optimization mode that performs a systematic completeness audit of the proposal, labeling each proposed component as fully implemented, simplified, or missing, and generating prioritized suggestions for recovering the missing innovations. This ensures that DDF contributes to exploration even when the system is already succeeding---rather than declaring victory after beating the baseline, it identifies the remaining gap between what was proposed and what was built.

\paragraph{Amplification via Dual-Agent Competition.}
The divergent nature of DDF is amplified by the system's dual-agent competition mechanism. When a diagnostic report is produced, each of two competing coding agents independently selects which 1--2 suggestions to pursue. Because the agents operate with different sampling trajectories, they naturally gravitate toward different subsets, converting the five-suggestion portfolio into parallel exploration branches. The winner is selected based on evaluation metrics, and the loser's approach is recorded in the cycle-level memory as a counterfactual observation. Appendix~\ref{app:ddf-examples} presents two real diagnostic reports---one failure-analysis and one post-baseline optimization---together with the five-suggestion portfolios and the divergent choices made by the two competing agents.

\subsection{From Experimental Evidence to Manuscripts}
\label{sec:manuscripts}

The final stage transforms the accumulated experimental evidence into a self-contained research manuscript. Every claim in the generated paper must be traceable to a specific experimental result: the system synthesizes text from structured evidence rather than generating prose from a general description.

\paragraph{Evidence Assembly and Reconciliation.}
When the experimental search and ablation stages complete, the system collects four categories of evidence: the research proposal, the quantitative experimental record, the ablation record, and the implementation code of the best configuration. A reconciliation step compares the original proposal against the final implementation, identifies discrepancies (omitted components, emergent additions, and modified modules), and produces a corrected methodology description that accurately reflects what was actually built.

\paragraph{Analysis, Visualization, and Writing.}
An analysis agent identifies the principal findings from the experimental and ablation records. A visualization agent generates result figures (performance comparison charts, ablation bar plots) directly from the numerical data, and produces architecture diagrams after the paper text has been drafted so that they match the notation used in the methodology section. A writing agent then produces a complete manuscript organized into standard sections. A citation agent identifies technical claims requiring references, queries academic search engines, and inserts verified citations. The manuscript undergoes two rounds of automated revision: the first targets structural issues (duplicate sections, method--experiment inconsistencies); the second removes phrasing patterns characteristic of language-model-generated text. The final output is a compiled PDF together with a complete LaTeX project. Agent-level details for each of the six steps---reconciliation, analysis, figure generation, writing, citation management, and revision---together with the complete list of output artifacts are provided in Appendix~\ref{app:manuscript-pipeline-details}.

\section{Main Results}
\label{sec:results}

\subsection{End-to-End Autonomous Research at Scale}
\label{sec:e2e}

We evaluate \camyla on the full 31-dataset \camylabench through two independent end-to-end runs. The two runs differ only in the large language model used for proposal generation: DeepSeek~V3.2~\citep{deepseek2024v3} for \camylaD and Claude Sonnet~4.6 for \camylaS. All other system components---QWBE, LRM, DDF, and the \camylanet training framework---remain identical. Both runs operate under a zero-intervention protocol: once launched, the system autonomously generates proposals, executes experiments, and produces manuscripts without any human input.

\paragraph{Study Design.}
Of the 31 datasets, 5 serve as validation datasets used during system development, and the remaining 26 are held out for blind testing. A dataset is considered successful if the system's best result exceeds the baseline Dice by more than 0.5 percentage points, or achieves a lower HD95 when the Dice difference is within 0.5~pp.

\paragraph{Results.}
Table~\ref{tab:end_to_end_summary} summarizes the main outcomes. On the full benchmark, \camylaD succeeds on 18 out of 31 datasets (58.1\%) and \camylaS on 22 out of 31 (71.0\%). On the 26 blind-test datasets alone, the success rates are 53.8\% (14/26) and 69.2\% (18/26), respectively, while both runs achieve 4 out of 5 on validation. \camylaD achieves a mean Dice of 65.93\% across all 31 datasets compared to the 64.04\% baseline, an average improvement of +1.89~pp. \camylaS reaches 65.22\%, an average improvement of +1.18~pp. The largest single-dataset gain is +18.58~pp on NLSTseg (Dataset~5), where both runs improve the baseline from 23.20\% to above 41\%.

The two runs display distinct success profiles. Of the 18 wins for \camylaD, 16 are achieved through Dice improvement exceeding the 0.5~pp threshold, and 2 through HD95 tiebreak. \camylaS wins 11 datasets on Dice and 11 on HD95 tiebreak, suggesting that the Sonnet-based generator produces methods that improve boundary precision more consistently. Table~\ref{tab:cross_run} shows that 16 datasets are won by both runs, 2 by \camylaD alone, 6 by \camylaS alone, and 7 by neither. The union of both runs covers 24 out of 31 datasets (77.4\%), substantially more than either run individually.

In terms of search efficiency, \camylaD explores 863 total experimental nodes (mean 27.8 per dataset), while \camylaS explores 514 nodes (mean 16.6). Among successful datasets, the first node exceeding the baseline appears at mean positions of 5.7 and 7.6, respectively, indicating that both runs typically identify a successful method within the first proposal. The total computational cost is approximately 28 days on an 8-GPU cluster across all 62 experiments. The LLM API cost averages \$23 per dataset for the DeepSeek run and \$26 for the Sonnet run. Per-dataset GPU-hour breakdowns and a cost decomposition across pipeline stages are provided in Appendix~\ref{app:computational-cost}. For each successful dataset, \camyla produces a full manuscript, yielding 18 and 22 papers for a total of 40.

\begin{table}[!htbp]
    \centering
    \small
    \caption{Summary of two independent \camyla runs on \camylabench (31 datasets). Success is defined as exceeding the baseline Dice by more than 0.5~pp, or achieving lower HD95 when Dice is within 0.5~pp. HD95 is measured in millimeters.}
    \label{tab:end_to_end_summary}
    \begin{tabular}{l r r}
    \toprule
    & \camylaD & \camylaS \\
    \midrule
    Idea generator model & DeepSeek V3.2 & Claude Sonnet 4.6 \\
    \midrule
    Success rate (all 31) & 18 / 31 (58.1\%) & 22 / 31 (71.0\%) \\
    \quad Blind-test (26) & 14 / 26 (53.8\%) & 18 / 26 (69.2\%) \\
    \quad Validation (5) & 4 / 5 & 4 / 5 \\
    \midrule
    Win via Dice & 16 & 11 \\
    Win via HD95 tiebreak & 2 & 11 \\
    \midrule
    Mean Dice (\%) & 65.93 & 65.22 \\
    Median Dice (\%) & 70.40 & 69.56 \\
    Mean Dice improvement (pp) & +1.89 & +1.18 \\
    Mean HD95 (mm) & 30.63 & 32.18 \\
    \midrule
    Total nodes explored & 863 & 514 \\
    Mean nodes per dataset & 27.8 & 16.6 \\
    First success position (mean / med.) & 5.7 / 4.5 & 7.6 / 6.0 \\
    \midrule
    Total wall time (8-GPU cluster) & \multicolumn{2}{c}{$\sim$28 days} \\
    LLM API cost per dataset (mean) & \$23 & \$26 \\
    Papers generated & 18 & 22 \\
    \bottomrule
    \end{tabular}
\end{table}

\begin{table}[!htbp]
    \centering
    \small
    \caption{Cross-run consistency between \camylaD and \camylaS. The union of both runs covers substantially more datasets than either run alone.}
    \label{tab:cross_run}
    \begin{tabular}{l c c}
    \toprule
    & All (31) & Blind-test (26) \\
    \midrule
    Both runs succeed & 16 & 12 \\
    \camylaD only & 2 & 2 \\
    \camylaS only & 6 & 6 \\
    Neither & 7 & 6 \\
    \midrule
    Union & 24 (77.4\%) & 20 (76.9\%) \\
    \bottomrule
    \end{tabular}
\end{table}

\begin{figure*}[!htbp]
    \centering
    \includegraphics[width=\linewidth]{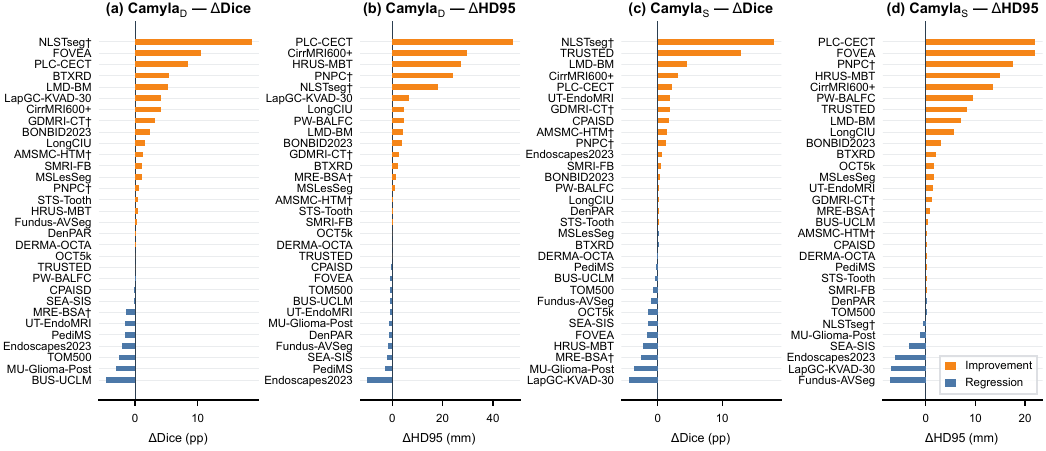}
    \caption{Per-dataset performance change over the best baseline for both independent runs. Four panels show (a)~$\Delta$Dice for \camylaD, (b)~$\Delta$Dice for \camylaS, (c)~$\Delta$HD95 for \camylaD, and (d)~$\Delta$HD95 for \camylaS. Orange bars indicate improvement; blue bars indicate regression. For HD95, positive values indicate reduced boundary error (improvement).}
    \label{fig:e2e-improvement}
\end{figure*}

\paragraph{Statistical Significance.}
Binomial tests confirm that both the \camylaS win rate (22/31, $p{=}0.015$) and the union rate (24/31, $p{=}0.002$) are significant under a conservative null ($p_0{=}0.5$). Per-sample Wilcoxon signed-rank tests show that 70\% of winning experiments are significant on at least one metric, with 20 surviving Bonferroni correction. Full statistical details are provided in Appendix~\ref{app:statistical-significance}.

\subsection{Comparison with Automated Baselines}
\label{sec:baselines}

\paragraph{Study Design.}
We compare \camyla against three categories of automated segmentation methods and two categories of open-ended research agents. Table~\ref{tab:baseline-characterization} characterizes the baselines. The automated segmentation methods are: \textsc{AutoNNUNet}~\citep{becktepe2025autonnunet}, which applies multi-fidelity Bayesian optimization to nnU-Net's design decisions; \textsc{DiNTS}~\citep{he2021dints}, which performs differentiable topology-then-cell search; and \textsc{AutoSeg3D}~\citep{myronenko2023auto3dseg}, integrated within MONAI~\citep{cardoso2022monai}, which selects among multiple algorithm families and ensembles their predictions. These methods operate within fixed, human-designed search spaces and do not produce research manuscripts. The open-ended research agents include four \texttt{autoresearch}~\citep{karpathy2026autoresearch} configurations (Claude Code + Opus~4.6, Claude Code + MiniMax~2.5, Claude Code + GLM~4.7, and Codex + GPT-5.4) and two AI Scientist systems~\citep{lu2024aiscientist,yamada2025aiscientistv2}. All research agents receive the same \camylanet documentation and operate on the 26 blind-test datasets.

\begin{table}[!htbp]
    \centering
    \small
    \caption{Characterization of automated baselines compared against \camyla. ``Search space'' indicates whether the method explores a predefined space (closed) or can propose arbitrary code modifications (open). ``Paper output'' indicates whether the method produces a manuscript.}
    \label{tab:baseline-characterization}
    \begin{tabular}{l l c c}
    \toprule
    \textbf{Method} & \textbf{Type} & \textbf{Search Space} & \textbf{Paper} \\
    \midrule
    AutoNNUNet & AutoML & Closed & No \\
    DiNTS & NAS & Closed & No \\
    AutoSeg3D & Ensemble & Closed & No \\
    \texttt{autoresearch} & Research agent & Open & No \\
    AI Scientist & Research agent & Open & Yes \\
    AI Scientist-v2 & Research agent & Open & Yes \\
    \camyla & Research agent & Open & Yes \\
    \bottomrule
    \end{tabular}
\end{table}

\paragraph{Results.}
Against automated segmentation methods, both \camyla runs outperform all baselines on aggregate Dice. Table~\ref{tab:baseline-summary} reports that the DeepSeek run achieves a mean Dice of 65.93\% across 31 datasets, compared to 64.27\% for \textsc{AutoNNUNet}, 58.03\% for \textsc{DiNTS}, and 63.43\% for \textsc{AutoSeg3D} (evaluated on the 18 volumetric datasets). At the dataset level, the DeepSeek run is Pareto-superior to \textsc{AutoNNUNet} on 15/31 datasets and to \textsc{AutoSeg3D} on 11/18 datasets, while these baselines are Pareto-superior to the DeepSeek run on only 9/31 and 5/18 datasets, respectively. Detailed per-dataset comparisons are provided in Appendix~\ref{app:comparison-tables} (Tables~\ref{tab:autonnunet_detailed_comparison}--\ref{tab:autoseg3d_detailed_comparison}).

\begin{table}[!htbp]
    \centering
    \small
    \caption{Aggregate comparison between \camyla and automated segmentation baselines. For each method, we report the mean Dice across evaluated datasets and the number of datasets on which the method is Pareto-superior to \camylaD. AutoSeg3D is evaluated only on the 18 volumetric (3D) datasets.}
    \label{tab:baseline-summary}
    \begin{tabular}{l c r c}
    \toprule
    \textbf{Method} & \textbf{Datasets} & \textbf{Mean Dice} & \textbf{Pareto vs \camylaD} \\
    \midrule
    AutoNNUNet & 31 & 64.27\% & 9/31 \\
    DiNTS (retrained) & 31 & 58.03\% & 1/31 \\
    AutoSeg3D & 18 & 63.43\% & 5/18 \\
    \camylaD & 31 & \textbf{65.93\%} & --- \\
    \camylaS & 31 & 65.22\% & --- \\
    \bottomrule
    \end{tabular}
\end{table}

Against open-ended research agents, Table~\ref{tab:research_agent_baselines} summarizes the comparison on the 26 blind-test datasets. Both \camyla runs complete all 26 datasets (100\% completion rate), with the DeepSeek run exceeding the baseline on 14/26 datasets (mean Dice 65.91\%, mean HD95 28.43~mm) and the Sonnet run on 18/26 (mean Dice 65.16\%, mean HD95 29.21~mm). Among the \texttt{autoresearch} variants, Claude Code + Opus~4.6 provides the best overall balance: it completes 18/26 tasks and exceeds the baseline on 10/26, while exhibiting the lowest proposal-drift rate (4/26) and the strongest quantitative profile among the \texttt{autoresearch} variants with a mean Dice of 63.14\% and a mean HD95 of 34.72~mm. Claude Code + MiniMax~2.5 achieves the highest baseline-surpassing rate among the \texttt{autoresearch} variants (12/26), but with substantial instability: it drifts from the proposal on 14/26 datasets and yields a weaker quantitative profile. The \textsc{AI Scientist} family is weaker on this benchmark: the original completes 18/26 experiments and exceeds the baseline on 5/26, reaching a mean Dice of only 60.24\% over completed runs; \textsc{AI Scientist-v2} completes only 12/26 with 3/26 exceeding the baseline. \camyla achieves zero proposal drift by construction, since the proposals are self-generated and the system architecture is specifically designed to maintain coherence between proposal and implementation.

\begin{table*}[!htbp]
    \centering
    \scriptsize
    \setlength{\tabcolsep}{4pt}
    \caption{Comparison with open-ended research agents on the 26 blind-test datasets. ``Completed'' denotes successful experiment execution; ``$>$Baseline'' counts datasets on which the system exceeds the baseline (Dice improvement $>$0.5~pp, or HD95 improvement when Dice is within 0.5~pp). ``Proposal Drift'' denotes clear deviation from the original plan. Dice and HD95 are summarized over completed runs only. \camylaD = DeepSeek~V3.2; \camylaS = Claude Sonnet~4.6.}
    \label{tab:research_agent_baselines}
    \resizebox{\textwidth}{!}{%
    \begin{tabular}{l l c c c r r r r}
    \toprule
    \textbf{Family} & \textbf{Configuration} & \textbf{Completed} & \textbf{$>$Baseline} & \textbf{Proposal Drift} & \textbf{Mean Dice} & \textbf{Mean HD95} & \textbf{Median Dice} & \textbf{Median HD95} \\
    \midrule
    \texttt{autoresearch} & Claude Code + Opus~4.6 & 18 / 26 & 10 / 26 & 4 / 26 & 63.14\% & 34.72 & 67.25\% & 20.83 \\
    \texttt{autoresearch} & Claude Code + MiniMax~2.5 & 15 / 26 & 12 / 26 & 14 / 26 & 62.58\% & 36.41 & 66.43\% & 22.15 \\
    \texttt{autoresearch} & Claude Code + GLM~4.7 & 16 / 26 & 10 / 26 & 12 / 26 & 62.21\% & 37.06 & 65.89\% & 23.04 \\
    \texttt{autoresearch} & Codex + GPT-5.4 (\texttt{xhigh}) & 12 / 26 & 8 / 26 & 10 / 26 & 61.35\% & 38.92 & 64.71\% & 24.67 \\
    \midrule
    \textsc{AI Scientist} & Default workflow & 18 / 26 & 5 / 26 & -- & 60.24\% & 40.55 & 63.18\% & 26.31 \\
    \textsc{AI Scientist-v2} & BFTS workflow & 12 / 26 & 3 / 26 & -- & 58.73\% & 43.18 & 61.52\% & 28.94 \\
    \midrule
    \camyla & DeepSeek~V3.2 (\camylaD) & 26 / 26 & 14 / 26 & 0 / 26 & 65.91\% & 28.43 & 69.62\% & 15.57 \\
    \camyla & Claude Sonnet~4.6 (\camylaS) & 26 / 26 & 18 / 26 & 0 / 26 & 65.16\% & 29.21 & 69.62\% & 16.01 \\
    \bottomrule
    \end{tabular}
    }
\end{table*}

\subsection{Manuscript Quality}
\label{sec:quality}

\paragraph{Study Design.}
We deliberately do not submit system-generated manuscripts to real venues or use live peer review as an evaluation instrument. Submitting autonomously generated papers would be inconsistent with the responsible-use policies of all target venues and would impose an unnecessary burden on volunteer reviewers. Instead, we construct a controlled, offline, and reproducible evaluation suite that combines human structured-summary review, multi-model AI summary review, and full-manuscript agentic review, benchmarked against 90 contemporaneous external papers. This design constitutes the best available controlled proxy for manuscript quality under the constraint that live submission is neither appropriate nor ethical.

To instantiate this evaluation, we construct an external benchmark of 90 contemporary medical image segmentation papers sampled from 18 journals indexed in Web of Science during 2025. The journals are organized into three tiers, summarized in Table~\ref{tab:benchmark-tiers}, with five papers sampled per journal; the full venue list is in Appendix~\ref{app:benchmark-composition}.

\begin{table}[!htbp]
    \centering
    \small
    \setlength{\tabcolsep}{4pt}
    \caption{Tier composition of the external evaluation benchmark (90 papers across 18 journals, five papers per journal). Representative venues are listed; see Appendix~\ref{app:benchmark-composition} for the complete list.}
    \label{tab:benchmark-tiers}
    \begin{tabular}{l c c p{0.5\linewidth}}
    \toprule
    \textbf{Tier} & \textbf{Journals} & \textbf{Papers} & \textbf{Representative venues} \\
    \midrule
    T1 & 2 & 10 & IEEE Transactions on Medical Imaging; Medical Image Analysis \\
    T2 & 7 & 35 & IEEE Journal of Biomedical and Health Informatics; Artificial Intelligence in Medicine; \textit{et al.} \\
    T3 & 9 & 45 & International Journal of Computer Assisted Radiology and Surgery; Biomedical Physics \& Engineering Express; \textit{et al.} \\
    \midrule
    \textbf{Total} & \textbf{18} & \textbf{90} & \\
    \bottomrule
    \end{tabular}
\end{table}

All 130 papers (90 external + 40 internal) are converted into structured paper summaries---standardized cards containing title, problem formulation, contributions, method overview, experimental findings, and conclusions. This representation removes superficial formatting differences and allows both human and AI evaluators to operate on a common interface. The evaluation proceeds on two complementary tracks: a structured-summary track assessed by 15 human reviewers (10 junior, 5 senior) and 5 frontier AI models, and a full-manuscript track assessed by Stanford Agentic Reviewer.

\paragraph{Card-Based Blind Review by Humans.}
Each reviewer assesses 30 papers, producing 450 review assignments in total. Each review scores four dimensions on a 0--5 scale: clarity of problem formulation and motivation, methodological novelty, experimental completeness, and overall recommendation. Table~\ref{tab:human-ai-tier-comparison} reports the aggregate results. Senior human reviewers place \camyla's internal papers (3.311) above the T2 mean (3.217) and close to the T1 mean (3.364). Junior reviewers exhibit a similar pattern, scoring internal papers at 3.053 compared to T1's 3.208 and T2's 3.109. Both reviewer groups place internal papers clearly above the T3 mean (2.792 for senior, 2.433 for junior).

\paragraph{Card-Based AI Review.}
Five frontier AI models (Claude Opus~4.6, DeepSeek~V3.2, Gemini~3 Flash Preview, Gemini~3.1 Pro Preview, and GPT-5.4) evaluate the same structured summaries. The AI average scores internal papers at 3.610, placing them on par with T1 papers (3.644) and well above T2 (3.051) and T3 (2.492). This pattern---internal papers scoring between T1 and T2 by human evaluation and at the T1 level by AI evaluation---suggests that the manuscripts generated by \camyla reach the quality level of strong clinical AI venues. Per-dimension breakdowns across the four rubric axes (problem formulation, methodological novelty, experimental completeness, and overall recommendation) are reported in Appendix~\ref{app:quality-dimensions}.

\begin{table}[!htbp]
    \centering
    \small
    \setlength{\tabcolsep}{5pt}
    \caption{Aggregate overall-recommendation scores by reviewer group and paper tier on the structured-summary track. Human rows correspond to the provisional reviewer matrix; the AI row reports the average over five frontier model evaluators.}
    \label{tab:human-ai-tier-comparison}
    \begin{tabular}{l c c c c c}
    \toprule
    \textbf{Source} & \textbf{T1} & \textbf{T2} & \textbf{T3} & \textbf{Internal} & \textbf{Ext.\ Overall} \\
    \midrule
    Junior reviewers & 3.208 & 3.109 & 2.433 & 3.053 & 2.954 \\
    Senior reviewers & 3.364 & 3.217 & 2.792 & 3.311 & 3.038 \\
    All human & 3.257 & 3.145 & 2.558 & 3.137 & 2.972 \\
    \midrule
    AI average (5 models) & 3.644 & 3.051 & 2.492 & 3.610 & 2.862 \\
    \bottomrule
    \end{tabular}
\end{table}

\paragraph{Full-Paper AI Review.}
To complement the summary-based evaluation, Stanford Agentic Reviewer scores complete manuscripts. Table~\ref{tab:stanford-agentic-review-comparison} reports the results. The reviewer assigns blind-test internal papers a mean score of 4.766, above the external overall mean (4.633), the T2 mean (4.603), and the T3 mean (4.422), while remaining below the T1 mean (5.690). Validation-derived internal papers score lower (4.363), consistent with the expectation that system development artifacts receive less favorable evaluation. These results confirm that the manuscripts produced by \camyla on unseen datasets approach the quality of T1/T2 venues in the contemporary medical image segmentation literature.

\begin{table}[!htbp]
    \centering
    \small
    \setlength{\tabcolsep}{4pt}
    \caption{Stanford Agentic Reviewer results on full manuscripts.}
    \label{tab:stanford-agentic-review-comparison}
    \resizebox{0.75\linewidth}{!}{%
    \begin{tabular}{l c c c c c c c}
    \toprule
    & \textbf{T1} & \textbf{T2} & \textbf{T3} & \textbf{Int.\ Blind} & \textbf{Int.\ Val.} & \textbf{Int.\ All} & \textbf{Ext.\ All} \\
    \midrule
    Stanford Agentic & 5.690 & 4.603 & 4.422 & 4.766 & 4.363 & 4.685 & 4.633 \\
    \bottomrule
    \end{tabular}%
    }
\end{table}

\begin{figure*}[!htbp]
    \centering
    \includegraphics[width=\linewidth]{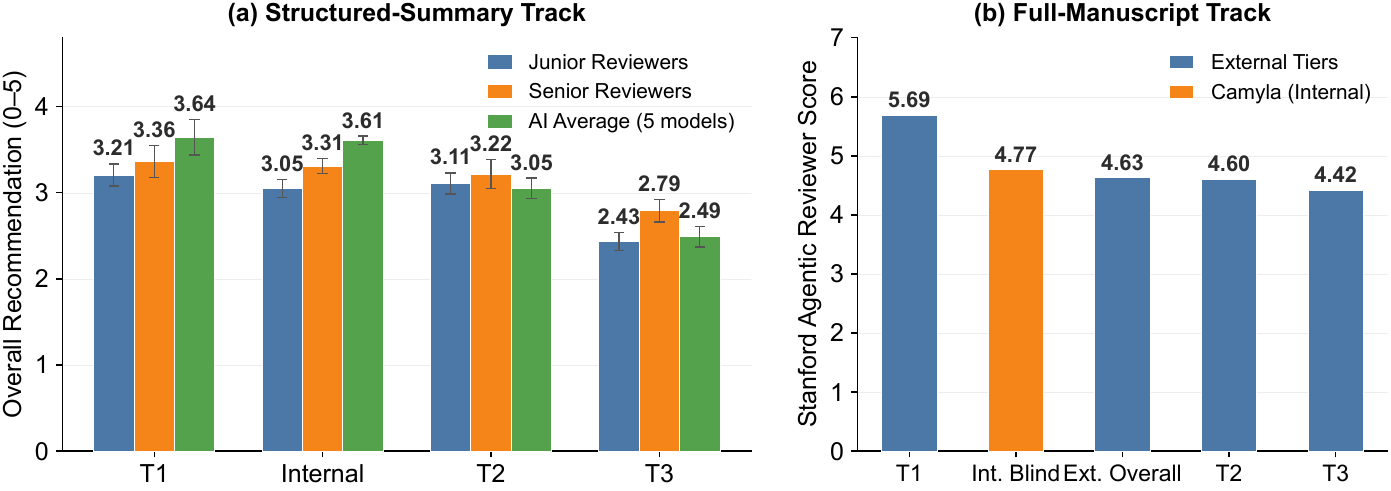}
    \caption{Manuscript quality scores across evaluation tracks. (a)~Structured-summary track: mean overall-recommendation scores from senior human reviewers and the five-model AI average, stratified by paper tier (T1, T2, T3) and \camyla-generated internal papers. The dashed line marks the external overall mean for senior reviewers. (b)~Full-manuscript track: Stanford Agentic Reviewer scores for external tiers and \camyla internal blind-test papers.}
    \label{fig:manuscript-quality}
\end{figure*}

Representative manuscripts are shown in Appendix~\ref{app:qualitative-examples}, including full page layouts for papers on neonatal brain lesion segmentation (Dataset~8, +4.25~pp Dice) and liver segmentation from cirrhotic MRI (Dataset~11, +5.14\% Dice, $-$33.85\% HD95). Across all 40 generated papers, each manuscript contains a self-contained literature review, a method section with architectural diagrams and mathematical formulations, and an experimental section with ablation studies and multi-baseline comparisons.

\section{Analysis}
\label{sec:analysis}

\subsection{Ablation of System Components}
\label{sec:ablation}

\paragraph{Experimental Protocol.}
We conduct ablation experiments on the five validation datasets (1, 2, 3, 4, and 5) to quantify the contribution of each core system component. Starting from the full \camyla configuration, we remove or replace one component at a time:
\begin{itemize}[leftmargin=1.5em, itemsep=2pt]
    \item \textbf{Full \camyla}: The complete system with QWBE, LRM, and DDF.
    \item \textbf{$-$QWBE}: QWBE replaced with uniform round-robin allocation.
    \item \textbf{$-$LRM}: Three-tier memory replaced with raw log injection.
    \item \textbf{$-$DDF}: Five-suggestion divergent diagnostic replaced with single-point prescriptive feedback.
\end{itemize}

\paragraph{Results.}
Table~\ref{tab:ablation-summary} summarizes the aggregate results. The full system achieves the strongest profile: 4/5 wins, a mean Dice improvement of +1.85~pp, a mean first-success position of 4.8, and a mean node budget of 21.4 nodes per dataset. Removing any single component degrades at least two of the four metrics. The $-$QWBE variant retains 3/5 wins but requires substantially more nodes to reach its first success (mean FSP increases from 4.8 to 9.5), indicating that the uniform allocation policy wastes budget on unpromising branches. The $-$LRM variant shows the sharpest decline in end-state quality: wins drop to 2/5 and mean $\Delta$Dice falls to +0.47~pp, suggesting that raw log injection causes context contamination that impairs sustained improvement across cycles. The $-$DDF variant maintains 3/5 wins but with lower mean $\Delta$Dice (+1.02~pp) and delayed first success (mean FSP = 7.3), reflecting the loss of diagnostic diversity.

The ablation results suggest that QWBE, LRM, and DDF function as a coupled system rather than independent add-ons: the three components address complementary failure modes. QWBE primarily governs \emph{where} computational resources are allocated: its removal nearly doubles the mean FSP from 4.8 to 9.5. LRM primarily governs \emph{what} the agent knows: on Dataset~5, removing LRM cuts the Dice gain from +18.58~pp to +8.35~pp, a 55\% reduction. DDF primarily governs \emph{how many} directions the agent considers: on Dataset~3, the full system recovers through a late-stage diagnostic insight (FSP = 19) that the single-fix variant never discovers. Dataset~2 is the only validation dataset where no configuration---including the full system---beats the baseline, indicating that this failure is rooted in the inherent difficulty of the task rather than in a specific component deficiency.

\begin{table}[!htbp]
    \centering
    \small
    \setlength{\tabcolsep}{6pt}
    \caption{Aggregate ablation results on the five validation datasets (1, 2, 3, 4, 5). \textbf{Wins}: datasets exceeding the best baseline. \textbf{Mean $\Delta$Dice}: average Dice improvement (pp). \textbf{Mean FSP}: average first-success position (lower is better; computed over winning datasets only). \textbf{Mean Nodes}: average total nodes expanded.}
    \label{tab:ablation-summary}
    \begin{tabular}{l c r r r}
    \toprule
    \textbf{Configuration} & \textbf{Wins} (\,/\,5) & \textbf{Mean $\Delta$Dice} & \textbf{Mean FSP}$\downarrow$ & \textbf{Mean Nodes} \\
    \midrule
    Full \camyla         & \textbf{4} & \textbf{+1.85} & \textbf{4.8} & 21.4 \\
    $-$QWBE              & 3          & +1.31          & 9.5          & 26.8 \\
    $-$LRM               & 2          & +0.47          & 6.3          & 24.2 \\
    $-$DDF               & 3          & +1.02          & 7.3          & 23.6 \\
    \bottomrule
    \end{tabular}
\end{table}

\subsection{Exploration Trajectories}
\label{sec:trajectory}

The aggregate metrics in \S\ref{sec:ablation} confirm that each component contributes to end-state performance, but they do not reveal how the search unfolds over time. Two detailed case studies in Appendix~\ref{app:trajectory-cases} provide node-by-node trajectory analyses that expose the technical content of each experimental trial. On Dataset~3 (PNPC), the system explores eight root-level DDAM variants before QWBE concentrates resources on the most promising branch; a major architectural redesign at node~9---replacing spatial attention with deformable sampling and CBAM-style attention---achieves the first baseline-surpassing result, with a subsequent refinement reducing HD95 by more than half. On Dataset~9 (BTXRD), six failed variants within the CCAN+MRFA paradigm trigger DDF to generate a categorically distinct direction: Multi-Scale Gated Axial Attention (MGAA), which decomposes 2D attention into sequential 1D axial operations and surpasses the baseline at node~7 by +2.7~pp Dice. These trajectories exemplify the interplay between broad exploration and focused refinement that QWBE and DDF are designed to produce.

\subsection{Cross-Run Stability and Variance}
\label{sec:stability}

The two independent runs differ only in the language model used for proposal generation, providing a natural test of system sensitivity to the idea generator. The concordance rate on success or failure is 23/31 datasets (74.2\%), and for the 16 co-won datasets the Pearson correlation of best Dice is $r{=}0.978$ (mean absolute difference 1.56~pp). The two generators produce complementary win profiles: \camylaD wins primarily through Dice improvement (16/18 wins), while \camylaS wins equally through Dice and HD95 tiebreak (11/11), suggesting different improvement modalities. \camylaS explores fewer nodes (514 vs.\ 863) yet achieves a higher success rate, pointing to proposal quality rather than search depth as the primary differentiator. The union covers 77.4\% of datasets---6 more than the better individual run---confirming genuine complementarity.

\subsection{Budget Sensitivity}
\label{sec:budget}

We analyze how much computational budget is necessary by retrospectively truncating the experimental trace at node index $N$ and recording whether the system has exceeded the baseline at that point.

\begin{figure}[!htbp]\centering
\includegraphics[width=0.55\linewidth]{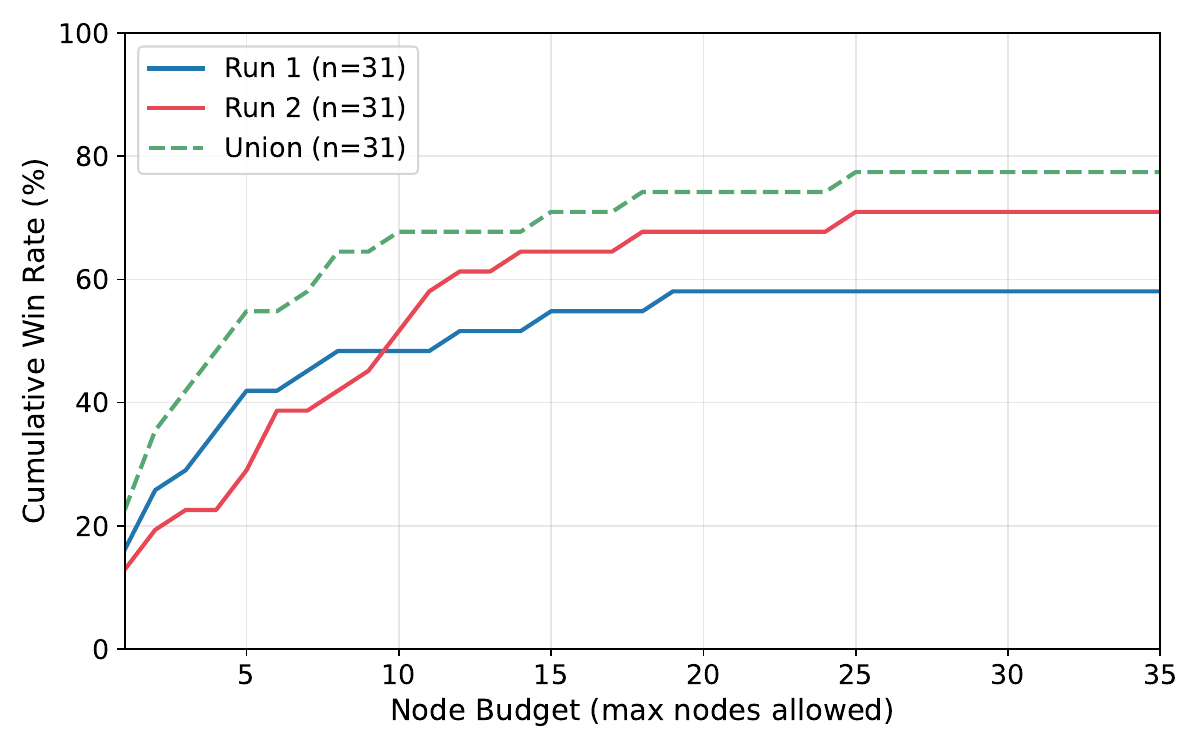}
\caption{Cumulative win rate as a function of node budget $N$. A dataset counts as won at budget $N$ if any node with index $\leq N$ exceeds the baseline. Two independent runs and their union are shown.}
\label{fig:budget-win-rate}
\end{figure}

\paragraph{Cumulative Win Rate.}
Figure~\ref{fig:budget-win-rate} plots the fraction of datasets won as a function of node budget $N$. Both runs exhibit a steep initial rise: at $N{=}5$, Run~1 reaches 38.7\% (12/31) and Run~2 reaches 29.0\% (9/31). The curves converge by $N{=}10$, where the two runs achieve comparable win rates of 48.4\% and 51.6\%. Beyond $N{=}10$, each additional node contributes diminishing returns. Run~1 reaches its final win rate at $N{=}20$; Run~2 continues to accumulate wins through $N{=}25$. Taking the union, the system exceeds the baseline on at least one run for 67.7\% of datasets at $N{=}10$ and 77.4\% at $N{=}30$.

\paragraph{First-Success Position.}
Across the 18 successful runs in Run~1, the median FSP is 4 and the mean is 5.2; across the 22 successful runs in Run~2, the median is 6 and the mean is 7.6. In Run~1, 12 of 18 successes (67\%) occur within the first 5 nodes. Run~2 exhibits a broader distribution: 9 of 22 (41\%) occur within the first 5, 7 between nodes 6--10, and 6 require more than 10. Among the 40 successful runs across both rounds, 30 (75\%) succeed within a single proposal (10 nodes), 9 require two proposals, and 1 requires all three. The 10 runs depending on a second or third proposal serve as a recovery mechanism for initially unpromising research directions.

\subsection{Failure Cases}
\label{sec:failures}

Seven of 31 datasets are never surpassed by either run (Datasets~2, 13, 17, 22, 23, 27, and 29). These failures share two recurring patterns: (1)~strong baselines with narrow improvement margins (e.g., Dataset~13 at 96.2\% Dice, Dataset~29 at 92.2\%), and (2)~tasks where multi-class segmentation with high inter-structure variability makes uniform gains from a single architectural modification difficult (e.g., Dataset~2 with 10 classes, Dataset~22 with 4 classes). A detailed case study of Dataset~2 (MRE-BSA) is provided in Appendix~\ref{app:failure-case-study}, where both runs exhaust all 60 trials without exceeding the U-Mamba baseline, confirming that this limitation is rooted in the task--baseline interaction rather than a component deficiency.

\section{Related Work}
\label{sec:related}

\paragraph{AI for Research.}
Large language models have progressed from code completion~\citep{chen2021codex} to autonomous software engineering agents that navigate repositories, edit files, and execute tests~\citep{jimenez2024swebench,yang2024sweagent,wang2024openhands}, raising the question of whether the full research lifecycle---hypothesis generation, experimentation, and manuscript generation---can be similarly automated.

Early end-to-end systems addressed this question in the machine-learning domain. The AI Scientist~\citep{lu2024aiscientist} generated research ideas, ran experiments, and wrote papers at under \$15 per paper, establishing the first fully autonomous research pipeline. Its successor, AI Scientist-v2~\citep{yamada2025aiscientistv2}, replaced the linear execution strategy with a progressive agentic tree search, producing the first AI-generated paper accepted at a peer-reviewed ICLR workshop. Several concurrent systems explore alternative decompositions of the research process: MLR-Copilot~\citep{li2024mlrcopilot} separates ideation from implementation with an RL-tuned idea generator; Agent Laboratory~\citep{schmidgall2025agentlab} inserts optional human feedback at each stage, reducing cost by 84\%; and CodeScientist~\citep{jansen2025codescientist} frames ideation as genetic search over code blocks, discovering 19 novel artifacts. DeepScientist~\citep{weng2026deepscientist} formalizes discovery as Bayesian optimization with a cumulative Findings Memory, surpassing human SOTA on three AI tasks through month-long autonomous cycles, though at a cost of over 20{,}000 GPU hours across three tasks without manuscript generation. On the ideation front, multi-agent architectures---iterative review~\citep{baek2024researchagent}, virtual scientific teams~\citep{su2024virsci}, and persistent memory~\citep{lyu2026evoscientist}---have improved idea novelty and feasibility over single-agent baselines~\citep{guo2025ideabench}. In the scientific-discovery setting, FunSearch~\citep{romeraparedes2024funsearch} paired an LLM with a systematic evaluator in an evolutionary loop to surpass best-known results in extremal combinatorics, demonstrating that LLM-guided program search can yield genuine mathematical discoveries. A recent survey~\citep{zheng2025llmscidiscovery} charts the broader landscape from task automation to autonomous agents.

These ideas are increasingly applied in domain-specific settings. The Medical AI Scientist~\citep{wu2026medaiscientist} is most closely related to our work, tailoring the paradigm to clinical research with a clinician-engineer co-reasoning mechanism across 19 clinical tasks and 6 modalities; however, it defines success as executable code rather than baseline-surpassing performance, evaluates manuscript quality on only 5 generated papers against 15 venue-matched references from a single task (diabetic retinopathy classification), and employs a single-pass pipeline without tree search or performance-aware diagnostic feedback. Other domain-specific systems include OpenLens AI~\citep{cheng2025openlens} for health informatics, DORA~\citep{naumov2025dora} for multi-agent report generation, SpatialAgent~\citep{wang2025spatialagent} for spatial biology, and PharmAgents~\citep{gao2025pharmagents} for drug discovery.

A simpler alternative to purpose-built pipelines is to repurpose general-purpose coding agents directly for research. The \texttt{autoresearch} paradigm~\citep{karpathy2026autoresearch} wraps a coding agent (e.g., Claude Code) in a loop that proposes code modifications, executes experiments, and observes scalar rewards---essentially treating research as a code-optimization problem. This approach has shown promise in discovering adversarial attack algorithms~\citep{panfilov2026claudini}, multimodal memory architectures~\citep{liu2026omnisimplemem}, and even meta-optimizing the autoresearch loop itself~\citep{qu2026bilevelautoresearch}. We evaluate several \texttt{autoresearch} configurations in \S\ref{sec:baselines}, finding that while they can occasionally discover strong improvements, their completion rates (46--69\%) and reliability fall substantially below those of a structured research system.

Despite this rapid progress, a common limitation persists: existing systems have been validated at small experimental scale---typically one to three datasets per run for end-to-end systems. It therefore remains unclear whether the discoveries generalize across diverse problem settings or whether the systems can reliably operate without human intervention at scale. \camyla addresses this gap with a strict zero-intervention protocol across 31 medical image segmentation datasets in two independent runs, achieving 100\% task completion and baseline-surpassing performance on 22 and 18 datasets respectively, while producing 40 complete manuscripts. In a controlled evaluation against 90 contemporary published papers stratified by venue tier, senior human reviewers scored \camyla's manuscripts at the T1/T2 boundary (3.311 vs.\ T1: 3.364, T2: 3.217 on a 0--5 scale), five frontier AI raters placed them on par with T1 (3.610 vs.\ 3.644), and Stanford Agentic Reviewer assigned blind-test manuscripts a mean of 4.766, above the T2 mean (4.603) and the external overall mean (4.633).

\paragraph{Automated Medical Image Segmentation.}
We use the term \emph{automated} to refer to systems that autonomously configure, optimize, or search for segmentation models given a new dataset, as distinct from \emph{automatic} segmentation, which denotes the inference-time application of a pre-trained model to produce masks without human interaction. Automated methods aim to remove human expertise from the model-development loop itself.

The dominant paradigm is rule-based self-configuration. nnU-Net~\citep{isensee2021nnu} demonstrated that a carefully engineered set of heuristic rules---automatic selection of patch size, batch size, network topology, and preprocessing---can match or exceed hand-tuned architectures across a wide spectrum of datasets, establishing the strongest general-purpose baseline. A large-scale revisitation~\citep{isensee2024nnunetrevisited} further showed that CNN-based U-Net variants within this framework, when properly scaled, continue to outperform Transformer-based and Mamba-based architectures, underscoring the enduring strength of rule-based configuration. The Medical Segmentation Decathlon~\citep{antonelli2022msd} provided a standardized multi-task benchmark that catalyzed progress in this direction. Auto-nnU-Net~\citep{becktepe2025autonnunet} takes this line further by applying AutoML techniques---specifically, multi-fidelity Bayesian optimization---to automatically tune nnU-Net's design decisions (e.g., network depth, patch size, augmentation strategy) that are otherwise set by heuristic rules, achieving improved performance on multiple MSD tasks. We compare against Auto-nnU-Net as a primary baseline in \S\ref{sec:baselines}.

Neural architecture search (NAS) methods extend automation from configuration to architecture design. Early work introduced differentiable search for U-Net cell structures~\citep{weng2019nasunet} and automated 2D/3D/P3D convolution selection per layer~\citep{zhu2019vnas}. Subsequent methods addressed scalability: C2FNAS~\citep{yu2020c2fnas} proposed coarse-to-fine topology-then-cell search, and DiNTS~\citep{he2021dints} introduced differentiable topology search with GPU memory budget constraints and a topology loss to close the discretization gap, ranking first on the MSD leaderboard. Auto3DSeg~\citep{myronenko2023auto3dseg}, integrated within MONAI~\citep{cardoso2022monai}, selects among multiple algorithm families and ensembles their predictions, winning the KiTS 2023 challenge~\citep{heller2021kits}. These methods, along with the broader AutoML foundations~\citep{hutter2019automl,he2021automlsurvey} that underpin them, share a key property: they all operate within a \emph{fixed, human-designed search space}. The set of candidate operations, connectivity patterns, and hyperparameter ranges is determined before search begins, and the methods cannot propose qualitatively new modules or mechanisms beyond this predefined vocabulary.

\camyla occupies a fundamentally different position. Rather than searching within a closed architecture space, it generates literature-grounded research proposals that specify novel architectural modules, implements and evaluates them through structured tree search, and synthesizes the results into complete manuscripts. CamylaNet provides an agent-friendly extension interface on top of nnU-Net's self-configuring infrastructure, combining the reliability of rule-based preprocessing with the flexibility of open-ended architectural innovation. This design bridges two previously disconnected paradigms: automated segmentation pipelines, which optimize models within a closed search space but do not produce transferable research insights, and autonomous research agents, which can write papers but lack the domain-specific workbench for rigorous medical image analysis at scale.

\section{Discussion}
\label{sec:discussion}

\paragraph{What this work demonstrates.}
The central finding of this paper is that a carefully designed autonomous research system can produce competitive methods and reviewable manuscripts across a broad set of tasks within a single research domain. Across two independent runs on 31 medical image segmentation datasets, \camyla completes every pipeline instance and, under identical training budgets, surpasses the strongest per-dataset baseline selected from 14 established architectures including nnU-Net on 22 and 18 datasets, respectively. The system generates 40 manuscripts that senior human reviewers place at the T1/T2 boundary of contemporary medical imaging journals, and achieves 100\% task completion with zero proposal drift, a reliability level that none of the open-ended research agent baselines approach.

\paragraph{The role of domain-specific infrastructure.}
A recurring theme in our results is that domain-specific infrastructure, specifically \camylabench and \camylanet, is not merely a convenience but a prerequisite for reliable autonomous research. The three-function workbench of \camylanet ensures that every experiment executes under the same preprocessing, augmentation, and evaluation protocol, eliminating a large class of confounds that plague open-ended research agents. The one-epoch verification function catches implementation errors before committing to full training, reducing wasted computation. The precomputed baseline bank provides a strong, per-dataset reference point that raises the bar above what a single fixed architecture would yield. Without these components, the system would face the same reliability challenges observed in the \texttt{autoresearch} and AI Scientist baselines: incomplete runs, proposal drift, and inconsistent evaluation.

\paragraph{Complementarity of system components.}
The ablation study reveals that no single component is sufficient and no component is redundant, with each mechanism addressing one of the three core challenges identified in the introduction. QWBE counters search drift by concentrating resources on high-quality branches; its removal nearly doubles the first-success position. LRM counters knowledge degradation by preserving cross-cycle scientific knowledge; its removal causes the sharpest decline in end-state quality, particularly on datasets requiring sustained multi-cycle refinement. DDF counters incremental-fix traps by maintaining exploratory diversity; its removal prevents the system from recovering on datasets where early trials are uniformly unpromising. The interaction between these components is as important as their individual contributions: DDF generates the diagnostic diversity that LRM compresses and QWBE exploits.

\paragraph{Limitations.}
Several limitations constrain the scope of our conclusions.

\begin{enumerate}[leftmargin=*]
\item \textbf{Single-domain validation.} Although the orchestration mechanisms and the workbench abstraction are designed to be portable, this paper validates them only in medical image segmentation. Extending to other domains such as reinforcement learning or materials science requires constructing comparable execution substrates and evaluation protocols, and whether the same orchestration principles transfer to fundamentally different experimental workflows remains an open question.

\item \textbf{Proxy-based manuscript evaluation.} The manuscript-quality results measure reviewability, completeness, and venue-tier proximity, not real acceptance decisions or complete proxies for community novelty judgment. Our evaluation suite combines human review, multi-model AI review, and full-manuscript agentic review against a stratified external benchmark, but a gap between these proxy assessments and genuine expert scrutiny may persist, particularly for subtle dimensions such as clinical relevance and methodological novelty.

\item \textbf{Constrained search space.} The system's search space is bounded by the coding capabilities of current language models. Proposals that require custom CUDA kernels, unconventional training procedures, or modifications to the loss function beyond the standard Dice-plus-cross-entropy formulation are effectively out of reach.

\item \textbf{Residual failure cases.} Seven of 31 datasets resist both runs, indicating that the system's ability to improve upon strong baselines is bounded, particularly for multi-class tasks with narrow improvement margins.
\end{enumerate}

\paragraph{Broader implications.}
\textbf{Infrastructure quality, not model capability, appears to be the primary bottleneck for reliable autonomous research in this setting.} The transition from 46\% to 69\% completion rates in general-purpose research agents to 100\% in \camyla is driven not by a more capable LLM but by domain-specific abstractions that reduce the search space, enforce experimental rigor, and provide structured feedback. This observation suggests a design hypothesis for future autonomous research systems: in domains with well-defined experimental workflows, investing in domain-specific workbenches that expose a minimal agent-executable interface, together with structured evaluation protocols and feedback mechanisms, may yield greater returns than relying on more capable foundation models alone. The workbench pattern demonstrated by \camylanet, which separates infrastructure from model definition so that agents interact through a narrow, reproducible contract, offers a template that could be adapted to other scientific domains with established training pipelines.

To support further research on autonomous scientific agents, we release CamylaTrace-232k, comprising over 232,000 agent events, 2,865 coding sessions, and 1,343 model implementations from the experimental-discovery phase. These traces enable analysis of how agents navigate complex design spaces, when and why they succeed or fail, and how diagnostic feedback shapes iterative refinement. Full scale statistics, directory layout, event schema, and a comparison with existing agent trajectory datasets are provided in Appendix~\ref{app:camylatrace}.

\section{Conclusion}
\label{sec:conclusion}

We introduced \camyla, an autonomous research system for domain-scale research in medical image segmentation. The system transforms raw datasets into literature-grounded research proposals, iterative experiments, and complete manuscripts through a fully automated four-stage pipeline. The key methodological contribution is three coupled mechanisms for long-horizon research orchestration: Quality-Weighted Branch Exploration for countering search drift, Layered Reflective Memory for countering knowledge degradation, and Divergent Diagnostic Feedback for countering incremental-fix traps. Evaluated under a strict zero-intervention protocol across 31 datasets and two independent runs, \camyla surpasses the strongest per-dataset baseline selected from 14 established architectures including nnU-Net on 22 and 18 datasets under identical training budgets, and produces 40 manuscripts that senior human reviewers score at the T1/T2 boundary of contemporary medical imaging journals. Together with \camylabench, \camylanet, and CamylaTrace-232k, these results demonstrate that domain-scale autonomous research is achievable in medical image segmentation. Future work includes extending the workbench abstraction to other scientific domains and broadening the system's search space beyond the coding capabilities of current language models.

\bibliographystyle{unsrtnat}
\bibliography{references}

\newpage
\beginappendix

\startcontents[app]
\begingroup
  \renewcommand{\contentsname}{Appendix Contents}
  \section*{\contentsname}
  \printcontents[app]{}{1}{}
\endgroup
\newpage

\section{Pipeline Stage Details}
\label{app:pipeline-stage-details}

\subsection{Stage Transitions and Early Stopping}
\label{app:stage-transitions}

The transition between stages is governed by deterministic rules evaluated after each experimental trial. Stage~1 completes immediately when precomputed baselines are available: the system loads the evaluation results of all compatible architectures from the baseline bank, selects the best as $b^*$, and proceeds to Stage~2 without executing any training. The transition from Stage~2 to Stage~3 depends on whether any trial has exceeded the baseline. After each proposal completes its allocated iterations, the system checks whether the best trial exceeds $b^*$. If so, remaining proposals are skipped and the system proceeds to Stage~3. If no proposal exceeds the baseline after all proposals have been explored, the experiment terminates without producing a manuscript. Among the 40 successful experiments, 30 (75\%) terminated after a single proposal, 9 required two proposals, and only 1 required all three.

\subsection{Subagent Competition}
\label{app:subagent-competition}

At each iteration within Stage~2 and Stage~3, the system instantiates two coding agents that independently implement the next modification. Both agents receive the same inputs: the research proposal, the current implementation code, the cycle-level memory from LRM, the global memory, and (when applicable) the diagnostic report from DDF. The agents operate with different random seeds, inducing diversity in their implementation choices. Each agent produces a complete modified implementation, which is trained and evaluated. The agent achieving the higher Dice is selected as the winner, and its implementation becomes the new current node. The losing agent's result is recorded in the cycle-level memory as a counterfactual observation.

\subsection{Ablation Study Construction}
\label{app:ablation-construction}

Stage~3 constructs ablation experiments by programmatically modifying the best implementation from Stage~2. For each proposed module, the system creates a variant in which that module is replaced with a minimal substitute (typically an identity mapping or a standard convolution). Each variant is trained from scratch to ensure that the ablation result reflects the module's contribution to learning rather than its effect on a pretrained representation. The results are saved in a structured record documenting, for each ablated module, the resulting Dice and HD95 scores together with the performance delta relative to the full model.

\section{Qualitative Examples of Generated Papers}
\label{app:qualitative-examples}

\begin{figure*}[!htbp]
    \centering
    \includegraphics[width=\linewidth]{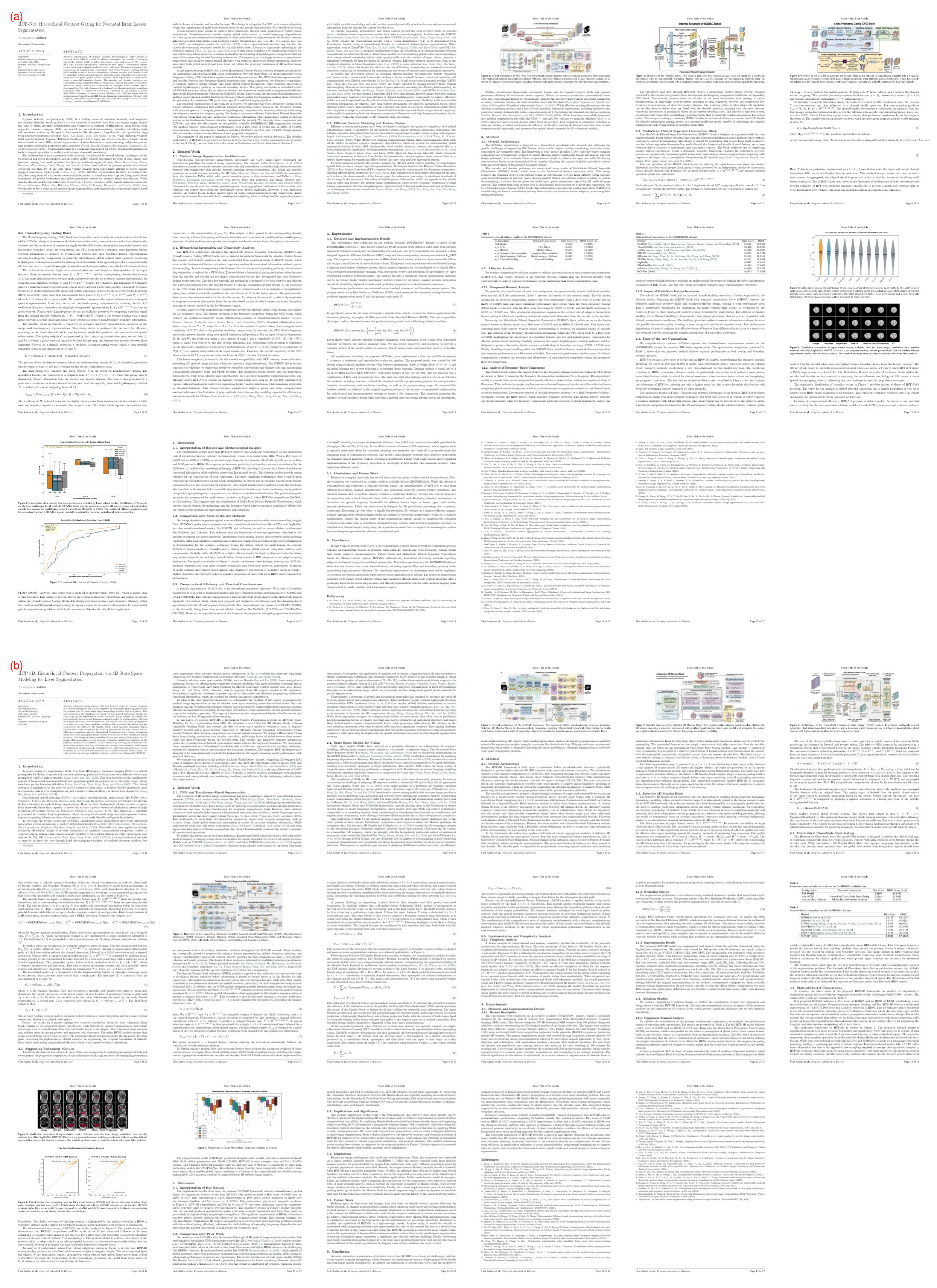}
    \caption{Full page layouts of two representative manuscripts generated by \camyla. (a)~Dataset~8: a 17-page paper on neonatal brain lesion segmentation proposing a hierarchical context gating architecture. (b)~Dataset~11: a 17-page paper on liver segmentation from cirrhotic MRI proposing a selective state space modeling framework. Both manuscripts are produced end-to-end without human intervention.}
    \label{fig:qualitative}
\end{figure*}

To illustrate the end-to-end output of \camyla, we select two representative manuscripts from the first run and present their full page layouts in Figure~\ref{fig:qualitative}. Both papers were generated without any human intervention, starting from raw dataset ingestion and concluding with a compiled PDF.

The first manuscript (Figure~\ref{fig:qualitative}a) addresses neonatal brain lesion segmentation on Dataset~8, a diffusion MRI dataset for hypoxic ischemic encephalopathy. The system proposes HCF-Net, a hierarchical context gating architecture that replaces standard skip connections with cross-frequency gating blocks operating on frequency-decomposed representations. The generated paper spans 17 pages and follows a standard six-section structure comprising introduction, related work, method, experiments, discussion, and conclusion. The method section introduces the architecture with formal equations and block diagrams, while the experimental section reports a Dice improvement of 4.25 percentage points over the strongest baseline with corresponding gains in boundary accuracy measured by HD95.

The second manuscript (Figure~\ref{fig:qualitative}b) targets liver segmentation from 3D MRI on Dataset~11, a collection of T2-weighted abdominal scans from patients with liver cirrhosis. Here the system designs HCP-3D, a framework that reformulates 3D context propagation as a selective state space modeling problem. The paper integrates a Selective 3D Mamba Block for capturing long-range dependencies with linear complexity and a hierarchical cross-scale state gating mechanism for multi-resolution feature propagation. The experimental evaluation reports a 5.14\% improvement in Dice and a 33.85\% reduction in HD95 relative to the best baseline.

\section{Failure Case Study}
\label{app:failure-case-study}

\paragraph{Dataset 2: MRE-BSA (multi-structure intestinal MRI).}
This dataset requires segmenting multiple abdominal structures from MRI enterography volumes. The U-Mamba~\citep{ma2024umamba} baseline achieves a Dice of 71.79\% and an HD95 of 17.75\,mm. Across both runs, the system exhausts the full budget of three proposals with ten iterations each, totaling 60 experimental trials, without producing a single result that exceeds the baseline.

\begin{figure}[!htbp]
    \centering
    \includegraphics[width=\linewidth]{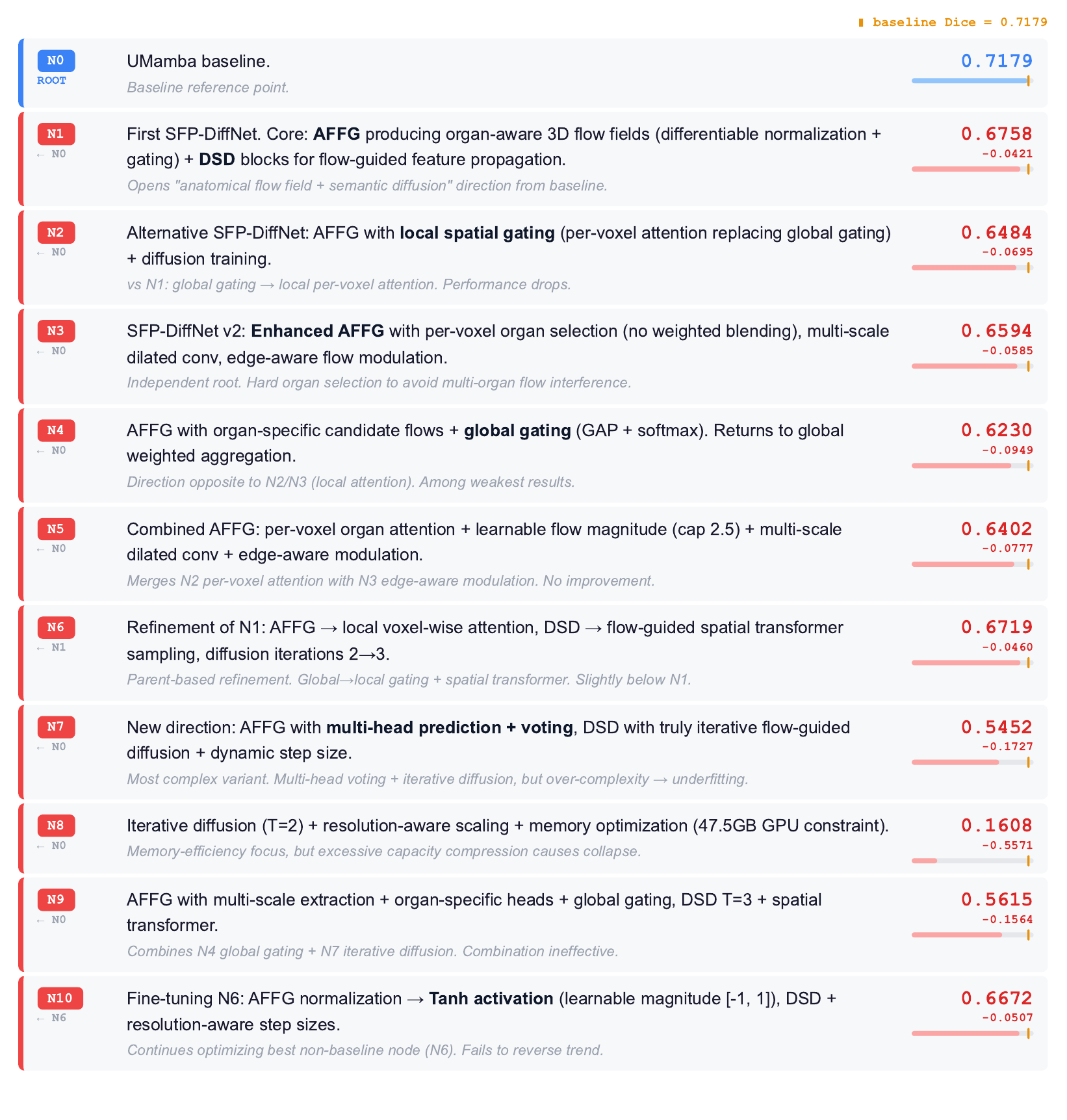}
    \caption{Node-by-node exploration trajectory for Dataset~2 (MRE-BSA, Proposal~1: SFP-DiffNet). Baseline: U-Mamba, Dice\,=\,0.7179. Each row shows one experimental node with its parent derivation, architectural description, and Dice score (red: below baseline; green: above). All 10 nodes fail to surpass the baseline. The system explores local vs.\ global gating, multi-head voting, and iterative diffusion variants of the Anatomical Flow Field Generator, but never escapes the ``flow field + diffusion'' framework. The best attempt (N1, Dice\,=\,0.6758) falls short by 4.2~pp.}
    \label{fig:trajectory-2}
\end{figure}

Figure~\ref{fig:trajectory-2} shows the detailed node-by-node trajectory from the first proposal of the DeepSeek run. All ten experimental nodes explore variants of SFP-DiffNet (Semantic Flow Diffusion), centered on an Anatomical Flow Field Generator (AFFG) and Differentiable Semantic Diffusion (DSD) blocks. The system tries local vs.\ global gating (N1--N5), parent-based refinement (N6, N10), multi-head voting with iterative diffusion (N7), and aggressive memory optimization (N8), but never escapes the ``flow field + diffusion'' framework. The best node (N1) reaches 67.58\% Dice, falling short by more than four percentage points. The second and third proposals explore distinct directions, yet the pattern repeats. The Sonnet run exhibits a similar pattern, with the closest result reaching 69.57\% Dice and 17.48\,mm HD95---narrowing the gap to 2.22~pp on Dice and improving on baseline HD95, but failing to meet the win threshold.

Two structural factors contribute to this failure. First, the U-Mamba baseline is unusually strong relative to task difficulty, leaving a narrow margin. Second, the dataset involves highly anisotropic volumetric data with substantial inter-structure variability, making it difficult for a single architectural modification to yield uniform gains across all evaluation classes.

\section{Exploration Trajectory Case Studies}
\label{app:trajectory-cases}

\begin{figure}[!htbp]
    \centering
    \includegraphics[width=\linewidth]{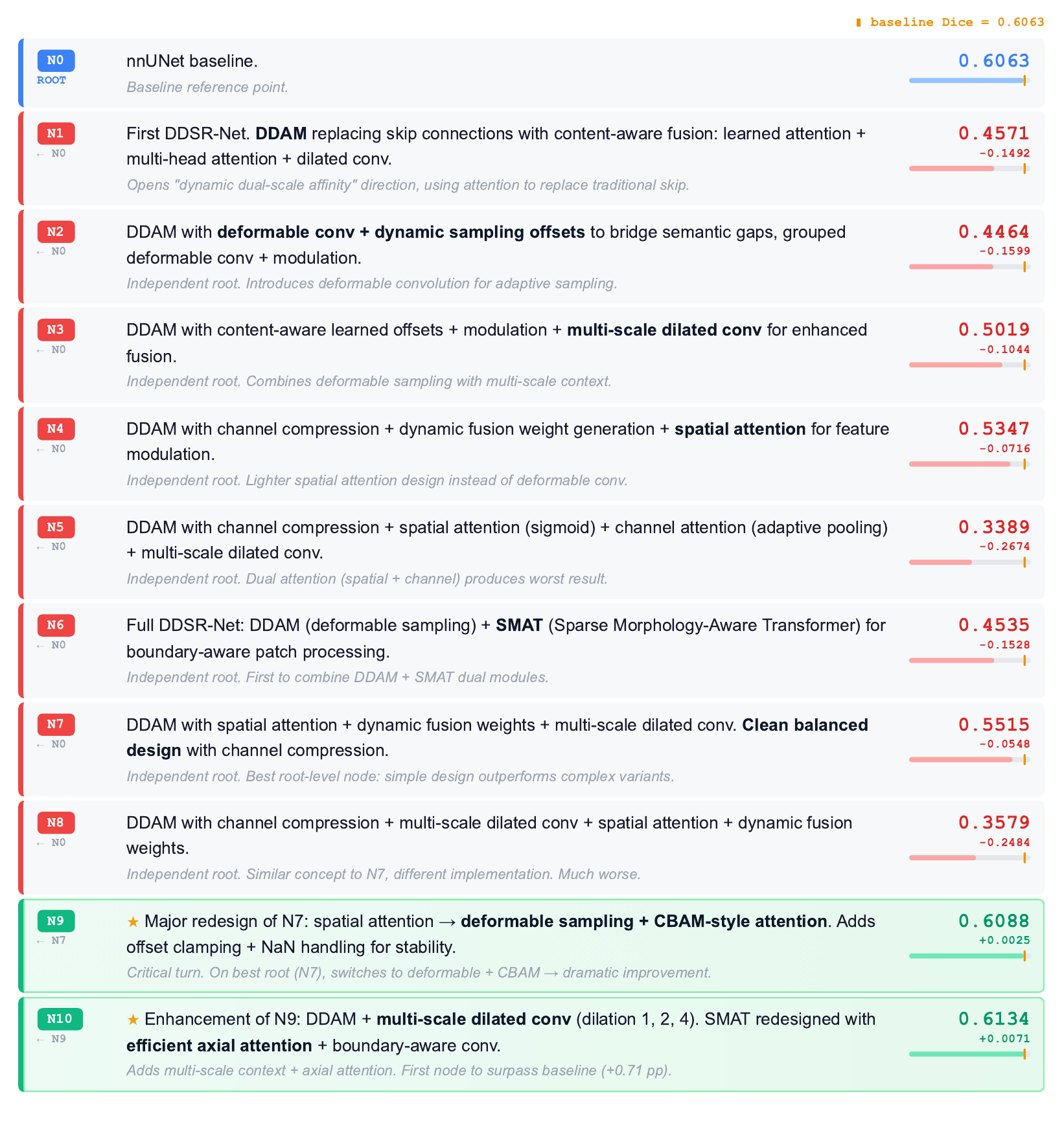}
    \caption{Node-by-node exploration trajectory for Dataset~3 (PNPC, Proposal~2: DDSR-Net). Baseline: nnU-Net, Dice\,=\,0.6063. Eight root branches explore diverse DDAM variants (attention-based, deformable, multi-scale), all below baseline. The turning point occurs at the N7$\to$N9 transition: the system identifies N7 as the strongest root (Dice\,=\,0.5515) and performs an architecture-level redesign---replacing spatial attention with deformable sampling and CBAM-style attention---jumping to 0.6088. N10 adds multi-scale context and axial attention, reaching 0.6134 and surpassing the baseline (+0.71~pp). The chain 7$\to$9$\to$10 demonstrates QWBE shifting from broad exploration to focused refinement.}
    \label{fig:trajectory-3}
\end{figure}

\paragraph{Case 1: Dataset 3 (PNPC, nasopharyngeal carcinoma MRI).}
The nnU-Net baseline achieves a Dice of 60.63\% with an HD95 of 45.28\,mm on this single-class head-and-neck segmentation task. Figure~\ref{fig:trajectory-3} shows the detailed node-by-node trajectory from Proposal~2 (DDSR-Net). The system begins by spawning eight root-level variants, each exploring different Dynamic Dual-Scale Affinity Module (DDAM) configurations: attention-based fusion (N1), deformable convolution with dynamic sampling (N2), multi-scale dilated convolution (N3), spatial attention (N4, N7), dual attention (N5), and the combined DDAM+SMAT architecture (N6). None surpasses the baseline: the best root node (N7, Dice\,=\,55.15\%) uses a clean balanced design with spatial attention and channel compression.

At this point, QWBE identifies N7 as the most promising branch despite its sub-baseline Dice. Node~9, a child of N7, performs a major architectural redesign---replacing spatial attention with deformable sampling and CBAM-style channel-spatial attention, with aggressive offset clamping and NaN handling for stability---achieving a Dice of 60.88\%. Node~10 further adds multi-scale dilated convolutions (dilation rates 1, 2, 4) and efficient axial attention, reaching 61.34\% Dice and 21.05\,mm HD95, reducing boundary error by more than half relative to the baseline.

This trajectory exemplifies the two-phase behavior that QWBE is designed to produce. The initial broad exploration establishes a performance landscape across structurally diverse candidates; once a promising direction emerges, the system concentrates resources and achieves progressive gains through targeted architectural redesign rather than incremental tuning.

\begin{figure}[!htbp]
    \centering
    \includegraphics[width=\linewidth]{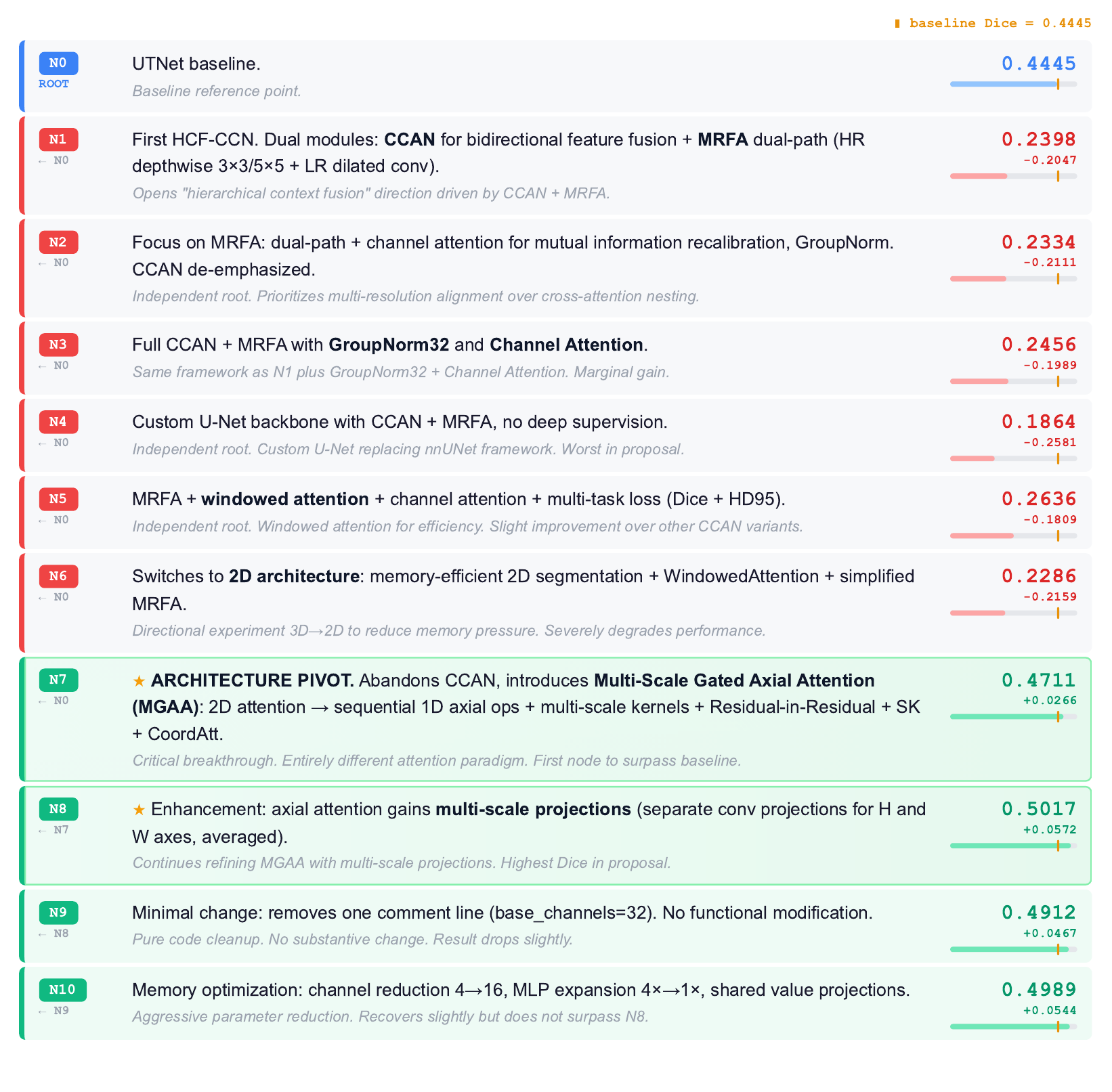}
    \caption{Node-by-node exploration trajectory for Dataset~9 (BTXRD, Proposal~1: HCF-CCN). Baseline: UTNet, Dice\,=\,0.4445. Nodes~1--6 explore the CCAN+MRFA direction (Dice 0.19--0.26, far below baseline). At N7, the system abandons CCAN entirely and introduces Multi-Scale Gated Axial Attention (MGAA)---decomposing 2D attention into sequential 1D axial operations---jumping to 0.4711 and surpassing the baseline for the first time. The subsequent chain 7$\to$8$\to$9$\to$10 refines MGAA to 0.5017 (+5.7~pp over baseline). This pivot is not incremental but a fundamentally different attention paradigm, exemplifying the behavior DDF is designed to produce.}
    \label{fig:trajectory-9}
\end{figure}

\paragraph{Case 2: Dataset 9 (BTXRD, bone tumor X-ray).}
The UTNet baseline yields a Dice of 44.45\% and an HD95 of 27.39\,mm. Figure~\ref{fig:trajectory-9} shows the detailed node-by-node trajectory from Proposal~1 (HCF-CCN). The system explores six root-level variants of Cascaded Cross-Attention Nesting (CCAN) combined with Multi-Resolution Feature Alignment (MRFA), all performing substantially below the baseline (Dice 18.64\%--26.36\%). The variants try different configurations---GroupNorm32 (N3), custom U-Net backbone (N4), windowed attention (N5), and a 2D architecture switch (N6)---but all remain within the CCAN+MRFA paradigm.

At N7, the system performs a complete architecture pivot: it abandons CCAN entirely and introduces Multi-Scale Gated Axial Attention (MGAA), decomposing 2D attention into sequential 1D axial operations with multi-scale kernels and a Residual-in-Residual structure for stable gradient flow. This produces the first baseline-surpassing result (Dice 47.11\%). N8 adds multi-scale convolutional projections to the axial attention, reaching the highest Dice of 50.17\% (+5.7~pp over baseline). N9 and N10 attempt code cleanup and memory optimization, respectively, but do not surpass N8.

This case highlights the role of divergent diagnostic feedback. After six failed trials within the same paradigm, the diagnostic module generates a categorically distinct improvement direction---axial attention decomposition rather than cross-attention nesting---rather than prescribing incremental fixes. In the $-$DDF ablation variant, the system tends to persist with minor corrections, and the mean first-success position increases from 4.8 to 7.3 across the validation datasets.

\section{Baseline Bank Details}
\label{app:baseline-bank-details}

The precomputed baseline bank comprises 14 architectures spanning three model families: convolutional (nnU-Net~\citep{isensee2021nnu}, SegResNet~\citep{myronenko2019segresnet}, U-Net++~\citep{zhou2018unetpp}, MedNeXt~\citep{roy2023mednext}, STU-Net~\citep{huang2023stunet}, UKAN~\citep{li2024ukan}), transformer-based (SwinUNETR~\citep{hatamizadeh2022swinunetr}, UNETR~\citep{hatamizadeh2022unetr}, nnFormer~\citep{zhou2023nnformer}, 3D~UX-Net~\citep{lee2023uxnet}, TransUNet~\citep{chen2021transunet}, UTNet~\citep{gao2021utnet}), and state-space models (U-Mamba~\citep{ma2024umamba}, SwinUMamba~\citep{liu2024swinumamba}). For each dataset in \camylabench, every compatible architecture is trained with polynomial learning rate decay from $10^{-2}$ and Dice~\citep{milletari2016vnet} plus cross-entropy loss. The agent autonomously decides whether to train for 100 or 1{,}000 epochs based on the observed convergence behavior. Table~\ref{tab:baseline-bank-compatibility} reports the architecture--configuration compatibility matrix.

\section{CamylaBench Dataset Details}
\label{app:dataset-diversity}

\paragraph{Anatomical and Clinical Diversity.}
The 31 datasets cover a broad cross-section of medical image segmentation tasks. Eight datasets target brain structures, including fetal brain parcellation, brain metastasis delineation, neonatal brain injury detection, ischemic penumbra segmentation, and multiple sclerosis lesion segmentation. Seven datasets address abdominal anatomy, ranging from intestinal segment annotation on MR enterography to liver segmentation on cirrhosis MRI and hepatic lesion delineation on contrast-enhanced CT. Four datasets cover ophthalmology, targeting optic disc segmentation, retinal artery--vein separation, retinal layer parsing on OCT, and orbital structure delineation. The remaining datasets span head and neck oncology, pulmonary imaging, dental radiography, breast ultrasound, dermatological OCTA, cytology, musculoskeletal radiology, soft tissue ultrasound, and gynecological MRI.

\paragraph{Task Difficulty Spectrum.}
The datasets span a wide range of segmentation difficulty. At the easier end, dental panoramic tooth segmentation (Dataset~13) achieves a baseline Dice above 96\%, reflecting well-defined boundaries and high contrast. At the harder end, surgical instrument segmentation in endoscopic video (Dataset~27) and ischemic penumbra delineation on CT (Dataset~12) yield baseline Dice scores of 14.2\% and 27.3\%, respectively, reflecting severe class imbalance, low tissue contrast, and ambiguous boundaries. The median baseline Dice across all 31 datasets is 71.4\%, indicating that the majority of tasks present meaningful room for improvement without being trivially solvable.

\section{Detailed Comparison Tables}
\label{app:comparison-tables}

\subsection{Per-Dataset Ablation Results}

Table~\ref{tab:ablation-per-dataset} disaggregates the aggregate ablation results reported in \S\ref{sec:ablation} to the individual dataset level. The four configurations---Full \camyla, $-$QWBE, $-$LRM, and $-$DDF---are evaluated on the five validation datasets under identical computational budgets (30 nodes each).

Three dataset-level patterns merit attention. First, Dataset~5 (NLSTseg) amplifies the role of LRM: the full system achieves +18.58~pp over a 23.20\% baseline, whereas removing LRM reduces this gain to +8.35~pp---a 55\% reduction---because the low-baseline regime demands sustained multi-cycle refinement that benefits most from structured memory. Second, Dataset~3 (PNPC) isolates the contribution of DDF: the full system recovers through a late diagnostic insight at FSP\,=\,19, while the $-$DDF variant never exceeds the baseline, confirming that single-point prescriptive feedback cannot generate the diversity required for difficult recovery scenarios. Third, Dataset~2 (MRE-BSA) resists all four configurations, indicating that this failure reflects the inherent difficulty of the task--baseline interaction rather than any specific component deficiency.

\begin{table*}[t]
    \centering
    \scriptsize
    \setlength{\tabcolsep}{2pt}
    \renewcommand{\arraystretch}{0.82}
    \caption{Per-dataset ablation results on the five validation datasets. For each configuration we report the best Dice achieved (\textbf{Best}), the Dice improvement over baseline ($\Delta$), the first-success position (\textbf{FSP}; ``--'' if never succeeded), and the total nodes expanded (\textbf{N}). $^*$Mean FSP is computed over winning datasets only.}
    \label{tab:ablation-per-dataset}
    \resizebox{0.75\textwidth}{!}{%
    \begin{tabular}{r l r r r r r @{\hskip 6pt} r l r r r r r}
    \toprule
    \textbf{ID} & \textbf{Dataset} & \textbf{BL} & \textbf{Best} & $\Delta$ & \textbf{FSP} & \textbf{N}
    & \textbf{ID} & \textbf{Dataset} & \textbf{BL} & \textbf{Best} & $\Delta$ & \textbf{FSP} & \textbf{N} \\
    \midrule
    \multicolumn{7}{c}{\textbf{Full \camyla}} & \multicolumn{7}{c}{\textbf{$-$QWBE}} \\
    \cmidrule(lr){1-7} \cmidrule(lr){8-14}
    1 & GDMRI-CT  & 72.33 & \textbf{73.87} & +1.53  & 12 & 30
    & 1 & GDMRI-CT  & 72.33 & 72.98 & +0.65  & 18 & 30 \\
    2 & MRE-BSA   & 71.79 & 70.40          & $-$1.39 & -- & 30
    & 2 & MRE-BSA   & 71.79 & 69.87 & $-$1.92 & -- & 30 \\
    3 & PNPC      & 60.63 & \textbf{61.34} & +0.71  & 19 & 30
    & 3 & PNPC      & 60.63 & 61.08 & +0.45  & 22 & 30 \\
    4 & AMSMC-HTM & 81.41 & \textbf{82.69} & +1.29  &  1 & 30
    & 4 & AMSMC-HTM & 81.41 & 82.25 & +0.84  &  6 & 30 \\
    5 & NLSTseg   & 23.20 & \textbf{41.78} & +18.58 &  1 & 30
    & 5 & NLSTseg   & 23.20 & 36.41 & +13.21 &  3 & 30 \\
    \cmidrule(lr){1-7} \cmidrule(lr){8-14}
      & \textbf{Mean}  & 61.87 & \textbf{66.02} & +4.14  & 4.8$^*$ & 30.0
    & & \textbf{Mean}  & 61.87 & 64.52 & +2.65  & 9.5$^*$ & 30.0 \\
      & \textbf{Wins}  &       & \textbf{4/5}   &        &         &
    & & \textbf{Wins}  &       & 3/5   &        &         & \\
    \midrule
    \multicolumn{7}{c}{\textbf{$-$LRM}} & \multicolumn{7}{c}{\textbf{$-$DDF}} \\
    \cmidrule(lr){1-7} \cmidrule(lr){8-14}
    1 & GDMRI-CT  & 72.33 & 72.61 & +0.28  & 15 & 30
    & 1 & GDMRI-CT  & 72.33 & 73.14 & +0.81  & 14 & 30 \\
    2 & MRE-BSA   & 71.79 & 70.02 & $-$1.77 & -- & 30
    & 2 & MRE-BSA   & 71.79 & 70.15 & $-$1.64 & -- & 30 \\
    3 & PNPC      & 60.63 & 60.41 & $-$0.22 & -- & 25
    & 3 & PNPC      & 60.63 & 60.52 & $-$0.11 & -- & 28 \\
    4 & AMSMC-HTM & 81.41 & 81.93 & +0.52  &  4 & 27
    & 4 & AMSMC-HTM & 81.41 & 82.41 & +1.01  &  3 & 26 \\
    5 & NLSTseg   & 23.20 & 31.55 & +8.35  &  6 & 30
    & 5 & NLSTseg   & 23.20 & 38.24 & +15.04 &  5 & 30 \\
    \cmidrule(lr){1-7} \cmidrule(lr){8-14}
      & \textbf{Mean}  & 61.87 & 63.30 & +1.43  & 6.3$^*$ & 28.4
    & & \textbf{Mean}  & 61.87 & 64.89 & +3.02  & 7.3$^*$ & 28.8 \\
      & \textbf{Wins}  &       & 2/5   &        &         &
    & & \textbf{Wins}  &       & 3/5   &        &         & \\
    \bottomrule
    \end{tabular}%
    }
\end{table*}

\subsection{Cross-Run Per-Dataset Comparison}

Table~\ref{tab:cross-run-detail} provides per-dataset detail for the 16 datasets on which both independent runs exceed the baseline, complementing the aggregate cross-run analysis in \S\ref{sec:stability}. The two runs use different idea generators (\camylaD: DeepSeek~V3.2; \camylaS: Claude Sonnet~4.6) but share all other system components.

The table reveals three forms of inter-run divergence. First, the \emph{win mechanism} differs systematically: \camylaD wins predominantly via Dice improvement (D), while \camylaS wins equally via HD95 tiebreak (H), suggesting that the Sonnet-based generator produces methods that improve boundary precision more consistently. Second, the \emph{magnitude} of improvement varies: the mean inter-run absolute Dice difference is 1.56~pp, but individual datasets range from near-identical (AMSMC-HTM: 0.20~pp) to substantially divergent (PLC-CECT: 6.17~pp), reflecting the stochastic nature of different LLM-generated proposals. Third, the \emph{search efficiency} differs: \camylaD reaches its first success earlier on average (mean FSP 5.2 vs.\ 7.1), but this faster convergence does not consistently translate into higher end-state quality. The high Pearson correlation ($r{=}0.978$) of best Dice across co-won datasets confirms that despite these surface differences, both runs converge to similar quality levels.

\begin{table*}[t]
    \centering
    \scriptsize
    \setlength{\tabcolsep}{3pt}
    \renewcommand{\arraystretch}{0.85}
    \caption{Per-dataset comparison of the two independent runs on the 16 co-won datasets. $\dagger$\,=\,validation. \textbf{Mech.}: D\,=\,Dice $>$0.5\,pp; H\,=\,HD95 tiebreak.}
    \label{tab:cross-run-detail}
    \resizebox{0.75\textwidth}{!}{%
    \begin{tabular}{r l r rrrr rrrr r}
    \toprule
    & & & \multicolumn{4}{c}{\camylaD (DeepSeek V3.2)} & \multicolumn{4}{c}{\camylaS (Claude Sonnet 4.6)} & \\
    \cmidrule(lr){4-7} \cmidrule(lr){8-11}
    \textbf{ID} & \textbf{Dataset} & \textbf{BL (\%)} & \textbf{Best} & \textbf{$\Delta$BL} & \textbf{FSP} & \textbf{Mech.} & \textbf{Best} & \textbf{$\Delta$BL} & \textbf{FSP} & \textbf{Mech.} & $|\Delta_\text{D-S}|$ \\
    \midrule
    1$^\dagger$  & GDMRI-CT      & 72.33 & 73.87 & +1.54 & 12 & D & 72.49 & +0.16 &  1 & H & 1.38 \\
    3$^\dagger$  & PNPC          & 60.63 & 61.34 & +0.71 & 19 & H & 61.95 & +1.32 & 18 & D & 0.61 \\
    4$^\dagger$  & AMSMC-HTM     & 81.41 & 82.69 & +1.28 &  1 & D & 82.89 & +1.48 &  6 & H & 0.20 \\
    5$^\dagger$  & NLSTseg       & 23.20 & 41.78 &+18.58 &  3 & D & 41.26 &+18.06 & 11 & D & 0.52 \\

    6  & SMRI-FB       & 85.30 & 86.44 & +1.14 &  2 & H & 85.74 & +0.44 &  6 & H & 0.70 \\
    7  & LMD-BM        & 46.19 & 51.51 & +5.32 &  1 & D & 50.69 & +4.50 &  1 & D & 0.82 \\
    8  & BONBID2023    & 55.85 & 58.23 & +2.38 &  5 & D & 56.17 & +0.32 &  2 & H & 2.06 \\
    9  & BTXRD         & 44.45 & 49.89 & +5.44 &  7 & D & 44.42 & $-$0.03 & 11 & H & 5.47 \\
    11 & CirrMRI600+   & 81.77 & 85.97 & +4.20 &  5 & H & 84.99 & +3.22 &  2 & D & 0.98 \\
    14 & DERMA-OCTA    & 83.57 & 84.01 & +0.44 &  8 & D & 83.81 & +0.24 & 14 & D & 0.20 \\
    18 & HRUS-MBT      & 42.69 & 50.55 & +7.86 &  6 & D & 47.83 & +5.14 &  3 & D & 2.72 \\
    20 & LongCIU       & 73.45 & 75.02 & +1.57 &  4 & H & 73.63 & +0.18 & 12 & H & 1.39 \\
    21 & MSLesSeg      & 69.57 & 70.63 & +1.06 &  1 & H & 69.56 & $-$0.01 &  1 & H & 1.07 \\
    25 & PLC-CECT      & 53.22 & 61.61 & +8.39 &  1 & D & 55.44 & +2.22 &  6 & D & 6.17 \\
    26 & PW-BALFC      & 76.66 & 76.63 & $-$0.03 &  7 & H & 76.87 & +0.21 &  9 & H & 0.24 \\
    28 & STS-Tooth     & 93.65 & 94.16 & +0.51 &  2 & H & 93.71 & +0.06 & 10 & H & 0.45 \\
    \midrule
    \multicolumn{2}{l}{\textbf{Mean}} & 65.25 & 68.96 & +3.77 & 5.2 & --- & 67.53 & +2.34 & 7.1 & --- & 1.56 \\
    \bottomrule
    \end{tabular}%
    }
\end{table*}

\subsection{Comparison with AutoNNUNet}

Table~\ref{tab:autonnunet_detailed_comparison} provides the full per-dataset breakdown underlying the aggregate comparison in \S\ref{sec:baselines}. \textsc{AutoNNUNet}~\citep{becktepe2025autonnunet} applies multi-fidelity Bayesian optimization to nnU-Net's design decisions within a closed search space of predefined hyperparameter ranges.

On aggregate, \camylaD achieves a mean Dice of 65.93\% compared to \textsc{AutoNNUNet}'s 64.27\% (+1.66~pp). At the dataset level, \camylaD achieves the best Dice on 17 of 31 datasets and the best HD95 on 15 of 31. \textsc{AutoNNUNet} retains advantages on datasets where nnU-Net's default architecture is already near-optimal and the search over its native hyperparameter space yields marginal but consistent gains (e.g., Dataset~5 NLSTseg: 47.33\% vs.\ 41.78\%, Dataset~23 OCT5k: 67.63\% vs.\ 65.35\%). In contrast, \camyla achieves its largest margins on datasets where architectural innovation beyond the nnU-Net family is required (e.g., Dataset~18 HRUS-MBT: 50.55\% vs.\ 38.03\%, Dataset~16 FOVEA: 81.91\% vs.\ 70.55\%). This pattern is consistent with the fundamental distinction between the two approaches: \textsc{AutoNNUNet} optimizes \emph{within} a fixed architecture family, while \camyla explores \emph{across} architectural families through open-ended code generation.

\begin{table*}[t]
    \centering
    \scriptsize
    \setlength{\tabcolsep}{3pt}
    \renewcommand{\arraystretch}{0.85}
    \caption{Detailed comparison between \textsc{AutoNNUNet} and \camyla (two independent runs) on the 31-dataset benchmark. $\dagger$\,=\,validation. Best values among the three methods are \textbf{bolded}.}
    \label{tab:autonnunet_detailed_comparison}
    \resizebox{0.75\textwidth}{!}{%
    \begin{tabular}{l l c rr rr rr rr}
    \toprule
    & & & \multicolumn{2}{c}{\textsc{AutoNNUNet}} & \multicolumn{2}{c}{\camylaD} & \multicolumn{2}{c}{\camylaS} & \multicolumn{2}{c}{Baseline} \\
    \cmidrule(lr){4-5} \cmidrule(lr){6-7} \cmidrule(lr){8-9} \cmidrule(lr){10-11}
    \textbf{ID} & \textbf{Dataset} & \textbf{Cfg.} & \textbf{Dice} & \textbf{HD95} & \textbf{Dice} & \textbf{HD95} & \textbf{Dice} & \textbf{HD95} & \textbf{Dice} & \textbf{HD95} \\
    \midrule
    1$^\dagger$ & GDMRI-CT & 3d & 72.83\% & 7.52 & \textbf{73.87\%} & \textbf{5.58} & 72.49\% & 7.11 & 72.33\% & 8.45 \\
    2$^\dagger$ & MRE-BSA & 3d & \textbf{70.73\%} & 16.38 & 70.40\% & \textbf{16.21} & 69.13\% & 17.00 & 71.79\% & 17.75 \\
    3$^\dagger$ & PNPC & 3d & 61.20\% & 52.45 & 61.34\% & \textbf{21.05} & \textbf{61.95\%} & 27.73 & 60.63\% & 45.28 \\
    4$^\dagger$ & AMSMC-HTM & 3d & 82.77\% & \textbf{2.60} & 82.69\% & 2.60 & \textbf{82.89\%} & 2.62 & 81.41\% & 2.89 \\
    5$^\dagger$ & NLSTseg & 3d & \textbf{47.33\%} & \textbf{111.85} & 41.78\% & 165.10 & 41.26\% & 183.73 & 23.20\% & 183.22 \\

    6 & SMRI-FB & 3d & 85.13\% & 1.29 & \textbf{86.44\%} & \textbf{1.14} & 85.74\% & 1.19 & 85.30\% & 1.23 \\
    7 & LMD-BM & 3d & 43.82\% & 31.88 & \textbf{51.51\%} & 28.44 & 50.69\% & \textbf{25.49} & 46.19\% & 32.54 \\
    8 & BONBID2023 & 3d & 53.92\% & 17.78 & \textbf{58.23\%} & \textbf{11.93} & 56.17\% & 12.76 & 55.85\% & 15.89 \\
    9 & BTXRD & 2d & 49.76\% & \textbf{20.67} & \textbf{49.89\%} & 25.01 & 44.42\% & 25.33 & 44.45\% & 27.39 \\
    10 & BUS-UCLM & 2d & \textbf{36.40\%} & 3.98 & 32.06\% & 4.84 & 36.18\% & \textbf{3.26} & 36.66\% & 3.80 \\
    11 & CirrMRI600+ & 3d & 84.71\% & 91.77 & \textbf{85.97\%} & \textbf{57.57} & 84.99\% & 73.49 & 81.77\% & 87.04 \\
    12 & CPAISD & 3d & 0.00\% & \textbf{50.00} & 27.22\% & 72.71 & \textbf{29.08\%} & 72.02 & 27.32\% & 72.25 \\
    13 & DenPAR & 2d & \textbf{96.52\%} & 4.26 & 96.39\% & 3.05 & 96.31\% & \textbf{1.79} & 96.22\% & 1.76 \\
    14 & DERMA-OCTA & 2d & 83.64\% & 4.29 & \textbf{84.01\%} & 3.28 & 83.81\% & \textbf{3.07} & 83.57\% & 3.28 \\
    15 & Endoscapes2023 & 2d & 47.69\% & \textbf{35.93} & 45.39\% & 48.81 & \textbf{48.13\%} & 45.15 & 47.53\% & 39.00 \\
    16 & FOVEA & 2d & 70.55\% & 72.40 & \textbf{81.91\%} & 71.98 & 69.69\% & \textbf{49.29} & 71.42\% & 71.29 \\
    17 & Fundus-AVSeg & 2d & 77.69\% & 61.70 & \textbf{77.95\%} & \textbf{60.95} & 76.53\% & 66.50 & 77.61\% & 59.30 \\
    18 & HRUS-MBT & 3d & 38.03\% & 100.60 & \textbf{50.55\%} & \textbf{60.85} & 47.83\% & 73.26 & 42.69\% & 88.16 \\
    19 & LapGC-KVAD-30 & 2d & 49.69\% & 70.78 & \textbf{56.98\%} & \textbf{59.67} & 48.26\% & 73.54 & 52.77\% & 66.45 \\
    20 & LongCIU & 2d & 73.55\% & 17.36 & \textbf{75.02\%} & 8.03 & 73.63\% & \textbf{7.11} & 73.45\% & 12.74 \\
    21 & MSLesSeg & 3d & 70.47\% & 16.92 & \textbf{70.63\%} & 15.30 & 69.56\% & \textbf{14.65} & 69.57\% & 16.33 \\
    22 & MU-Glioma-Post & 3d & \textbf{71.10\%} & \textbf{7.62} & 68.61\% & 8.32 & 67.82\% & 8.32 & 71.59\% & 7.24 \\
    23 & OCT5k & 2d & \textbf{67.63\%} & 9.43 & 65.35\% & 10.77 & 63.87\% & \textbf{9.09} & 65.35\% & 10.77 \\
    24 & PediMS & 3d & \textbf{81.03\%} & 3.45 & 77.85\% & 6.40 & 79.27\% & \textbf{3.41} & 79.49\% & 3.54 \\
    25 & PLC-CECT & 3d & 58.49\% & 61.69 & \textbf{61.61\%} & \textbf{59.58} & 55.44\% & 85.35 & 53.22\% & 107.41 \\
    26 & PW-BALFC & 2d & 76.14\% & 50.19 & 76.63\% & 47.84 & \textbf{76.87\%} & \textbf{42.98} & 76.66\% & 52.42 \\
    27 & SEA-SIS & 2d & \textbf{16.01\%} & 32.70 & 14.01\% & \textbf{15.85} & 12.63\% & 17.37 & 14.15\% & 13.92 \\
    28 & STS-Tooth & 2d & \textbf{94.33\%} & 0.83 & 94.16\% & \textbf{0.79} & 93.71\% & 0.88 & 93.65\% & 0.94 \\
    29 & TOM500 & 3d & \textbf{92.64\%} & \textbf{1.19} & 89.74\% & 2.08 & 91.50\% & 1.19 & 92.22\% & 1.14 \\
    30 & TRUSTED & 3d & 63.92\% & 38.41 & 62.13\% & 40.58 & \textbf{75.01\%} & \textbf{32.26} & 62.13\% & 40.58 \\
    31 & UT-EndoMRI & 3d & 74.66\% & 20.32 & 73.55\% & 13.34 & \textbf{77.03\%} & \textbf{10.74} & 75.10\% & 12.27 \\
    \midrule
    \textbf{Mean} & -- & -- & 64.27\% & 32.85 & \textbf{65.93\%} & \textbf{30.63} & 65.22\% & 32.18 & -- & -- \\
    \bottomrule
    \end{tabular}%
    }
\end{table*}

\subsection{Comparison with DiNTS}

Table~\ref{tab:dints_detailed_comparison} compares \textsc{DiNTS}~\citep{he2021dints} against both \camyla runs on all 31 datasets. \textsc{DiNTS} performs differentiable neural architecture search in two stages: a topology search over network-level connectivity followed by a cell-level operation search within discovered topologies. For each dataset, we report both the search-phase Dice and the retrained Dice after the best architecture is trained from scratch.

\textsc{DiNTS} achieves a mean retrained Dice of 58.03\%, substantially below both \camyla runs (65.93\% and 65.22\%). The search-phase Dice (56.90\%) is close to the retrained Dice, indicating that the search itself often converges to suboptimal architectures rather than failing at the retraining step. \textsc{DiNTS} struggles particularly on datasets with complex multi-class structures or unusual modalities: it achieves 0.00\% retrained Dice on Dataset~15 (Endoscapes2023) and below-baseline results on 22 of 31 datasets. Two factors contribute to this pattern. First, the fixed cell-level operation space does not include domain-specific modules (e.g., frequency decomposition, state-space models) that \camyla can generate through open-ended code synthesis. Second, the differentiable search procedure requires substantial GPU memory, limiting the resolution and batch size during the search phase, which is particularly detrimental for high-resolution 2D datasets. The datasets where \textsc{DiNTS} performs closest to \camyla tend to be 3D volumetric tasks with moderate resolution where the search procedure can operate effectively (e.g., Dataset~11 CirrMRI600+: 81.12\% vs.\ 85.97\%).

\begin{table*}[t]
    \centering
    \scriptsize
    \setlength{\tabcolsep}{2pt}
    \renewcommand{\arraystretch}{0.65}
    \caption{Detailed comparison between \textsc{DiNTS} and \camyla (two independent runs) on the benchmark. $\dagger$\,=\,validation. Best values among DiNTS, \camylaD, and \camylaS are \textbf{bolded}.}
    \label{tab:dints_detailed_comparison}
    \resizebox{0.75\textwidth}{!}{%
    \begin{tabular}{l l c rr rrr}
    \toprule
    & & & \multicolumn{2}{c}{\textsc{DiNTS}} & & & \\
    \cmidrule(lr){4-5}
    \textbf{ID} & \textbf{Dataset} & \textbf{Cfg.} & \textbf{Search} & \textbf{Retrained} & \textbf{\camylaD} & \textbf{\camylaS} & \textbf{Baseline} \\
    \midrule
    1$^\dagger$ & GDMRI-CT & 3d & 65.85\% & 66.15\% & \textbf{73.87\%} & 72.49\% & 70.29\% \\
    2$^\dagger$ & MRE-BSA & 3d & 57.92\% & 63.45\% & \textbf{70.40\%} & 69.13\% & 69.72\% \\
    3$^\dagger$ & PNPC & 3d & 48.21\% & 45.79\% & 61.34\% & \textbf{61.95\%} & 60.63\% \\
    4$^\dagger$ & AMSMC-HTM & 3d & 24.95\% & 79.36\% & 82.69\% & \textbf{82.89\%} & 81.41\% \\
    5$^\dagger$ & NLSTseg & 3d & 46.78\% & 24.45\% & \textbf{41.78\%} & 41.26\% & 23.20\% \\

    6 & SMRI-FB & 3d & 79.53\% & 84.97\% & \textbf{86.44\%} & 85.74\% & 84.85\% \\
    7 & LMD-BM & 3d & 16.09\% & 41.92\% & \textbf{51.51\%} & 50.69\% & 46.19\% \\
    8 & BONBID2023 & 3d & 36.67\% & 43.82\% & \textbf{58.23\%} & 56.17\% & 52.58\% \\
    9 & BTXRD & 2d & 35.58\% & 43.85\% & \textbf{49.89\%} & 44.42\% & 42.00\% \\
    10 & BUS-UCLM & 2d & 62.88\% & 35.77\% & 32.06\% & \textbf{36.18\%} & 36.66\% \\
    11 & CirrMRI600+ & 3d & 84.63\% & 81.12\% & \textbf{85.97\%} & 84.99\% & 81.79\% \\
    12 & CPAISD & 3d & 7.57\% & 26.73\% & 27.22\% & \textbf{29.08\%} & 23.87\% \\
    13 & DenPAR & 2d & 94.90\% & 93.64\% & \textbf{96.39\%} & 96.31\% & 96.22\% \\
    14 & DERMA-OCTA & 2d & 81.65\% & 83.59\% & \textbf{84.01\%} & 83.81\% & 83.28\% \\
    15 & Endoscapes2023 & 2d & 48.73\% & 0.00\% & 45.39\% & \textbf{48.13\%} & 47.53\% \\
    16 & FOVEA & 2d & 60.22\% & 60.25\% & \textbf{81.91\%} & 69.69\% & 70.78\% \\
    17 & Fundus-AVSeg & 2d & 65.06\% & 74.81\% & \textbf{77.95\%} & 76.53\% & 76.93\% \\
    18 & HRUS-MBT & 3d & 8.95\% & 34.98\% & \textbf{50.55\%} & 47.83\% & 50.09\% \\
    19 & LapGC-KVAD-30 & 2d & 47.97\% & 36.26\% & \textbf{56.98\%} & 48.26\% & 47.20\% \\
    20 & LongCIU & 2d & 71.56\% & 69.98\% & \textbf{75.02\%} & 73.63\% & 73.45\% \\
    21 & MSLesSeg & 3d & 65.33\% & 68.14\% & \textbf{70.63\%} & 69.56\% & 69.22\% \\
    22 & MU-Glioma-Post & 3d & 68.77\% & 65.48\% & \textbf{68.61\%} & 67.82\% & 71.59\% \\
    23 & OCT5k & 2d & 45.04\% & 58.69\% & \textbf{65.35\%} & 63.87\% & 64.06\% \\
    24 & PediMS & 3d & 44.51\% & 73.48\% & 77.85\% & \textbf{79.27\%} & 79.49\% \\
    25 & PLC-CECT & 3d & 28.67\% & 50.00\% & \textbf{61.61\%} & 55.44\% & 48.13\% \\
    26 & PW-BALFC & 2d & 76.34\% & 72.67\% & 76.63\% & \textbf{76.87\%} & 76.66\% \\
    27 & SEA-SIS & 2d & 85.33\% & 12.35\% & \textbf{14.01\%} & 12.63\% & 14.15\% \\
    28 & STS-Tooth & 2d & 91.38\% & 93.35\% & \textbf{94.16\%} & 93.71\% & 93.07\% \\
    29 & TOM500 & 3d & 85.61\% & 83.61\% & 89.74\% & \textbf{91.50\%} & 91.89\% \\
    30 & TRUSTED & 3d & 61.18\% & 61.74\% & 62.13\% & \textbf{75.01\%} & 62.13\% \\
    31 & UT-EndoMRI & 3d & 65.98\% & 68.59\% & 73.55\% & \textbf{77.03\%} & 74.52\% \\
    \midrule
    \textbf{Mean} & -- & -- & 56.90\% & 58.03\% & \textbf{65.93\%} & 65.22\% & -- \\
    \bottomrule
    \end{tabular}%
    }
\end{table*}

\subsection{Comparison with AutoSeg3D}

Table~\ref{tab:autoseg3d_detailed_comparison} compares \textsc{AutoSeg3D}~\citep{myronenko2023auto3dseg}, integrated within the MONAI framework~\citep{cardoso2022monai}, against both \camyla runs on the 18 volumetric (3D) datasets. \textsc{AutoSeg3D} is restricted to 3D configurations by design, so 2D datasets are excluded from this comparison. The method selects among multiple algorithm families (SegResNet, DiNTS, SwinUNETR) and optionally ensembles their predictions.

On aggregate, \camylaD achieves a mean Dice of 66.34\% on the 18 volumetric datasets, compared to \textsc{AutoSeg3D}'s 63.43\% (+2.91~pp). \camyla achieves the best Dice on 13 of 18 datasets. \textsc{AutoSeg3D} retains advantages on three datasets (MU-Glioma-Post, NLSTseg, and TOM500), all of which are tasks where the ensemble of established architectures provides complementary strengths that a single \camyla-generated architecture does not capture. The largest \camyla margins appear on datasets with challenging modalities or unusual anatomies---HRUS-MBT (+13.13~pp), BONBID2023 (+5.72~pp), PLC-CECT (+4.31~pp)---where domain-specific architectural innovations generated through open-ended search yield gains that cannot be reached by combining standard architectures.

\begin{table}[t]
    \centering
    \small
    \setlength{\tabcolsep}{3pt}
    \caption{Detailed comparison between \textsc{AutoSeg3D} and \camyla (two independent runs) on the 18 volumetric benchmark datasets. $\dagger$\,=\,validation. Best values among the three methods are \textbf{bolded}.}
    \label{tab:autoseg3d_detailed_comparison}
    \begin{tabular}{l l c rr rr rr rr}
    \toprule
    & & & \multicolumn{2}{c}{\textsc{AutoSeg3D}} & \multicolumn{2}{c}{\camylaD} & \multicolumn{2}{c}{\camylaS} & \multicolumn{2}{c}{Baseline} \\
    \cmidrule(lr){4-5} \cmidrule(lr){6-7} \cmidrule(lr){8-9} \cmidrule(lr){10-11}
    \textbf{ID} & \textbf{Dataset} & \textbf{Cfg.} & \textbf{Dice} & \textbf{HD95} & \textbf{Dice} & \textbf{HD95} & \textbf{Dice} & \textbf{HD95} & \textbf{Dice} & \textbf{HD95} \\
    \midrule
    1$^\dagger$ & GDMRI-CT & 3d & 72.03\% & 7.91 & \textbf{73.87\%} & \textbf{5.58} & 72.49\% & 7.11 & 72.33\% & 8.45 \\
    2$^\dagger$ & MRE-BSA & 3d & 69.78\% & 16.86 & \textbf{70.40\%} & \textbf{16.21} & 69.13\% & 17.00 & 71.79\% & 17.75 \\
    3$^\dagger$ & PNPC & 3d & 58.43\% & 50.30 & 61.34\% & \textbf{21.05} & \textbf{61.95\%} & 27.73 & 60.63\% & 45.28 \\
    4$^\dagger$ & AMSMC-HTM & 3d & 82.50\% & 2.72 & 82.69\% & \textbf{2.60} & \textbf{82.89\%} & 2.62 & 81.41\% & 2.89 \\
    5$^\dagger$ & NLSTseg & 3d & \textbf{41.84\%} & \textbf{122.56} & 41.78\% & 165.10 & 41.26\% & 183.73 & 23.20\% & 183.22 \\
    6 & SMRI-FB & 3d & 85.12\% & 1.26 & \textbf{86.44\%} & \textbf{1.14} & 85.74\% & 1.19 & 85.30\% & 1.23 \\
    7 & LMD-BM & 3d & 43.59\% & 32.13 & \textbf{51.51\%} & 28.44 & 50.69\% & \textbf{25.49} & 46.19\% & 32.54 \\
    8 & BONBID2023 & 3d & 52.51\% & 17.02 & \textbf{58.23\%} & \textbf{11.93} & 56.17\% & 12.76 & 55.85\% & 15.89 \\
    11 & CirrMRI600+ & 3d & 84.49\% & 90.11 & \textbf{85.97\%} & \textbf{57.57} & 84.99\% & 73.49 & 81.77\% & 87.04 \\
    12 & CPAISD & 3d & 5.88\% & \textbf{54.45} & 27.22\% & 72.71 & \textbf{29.08\%} & 72.02 & 27.32\% & 72.25 \\
    18 & HRUS-MBT & 3d & 37.42\% & 96.25 & \textbf{50.55\%} & \textbf{60.85} & 47.83\% & 73.26 & 42.69\% & 88.16 \\
    21 & MSLesSeg & 3d & 70.28\% & 16.65 & \textbf{70.63\%} & 15.30 & 69.56\% & \textbf{14.65} & 69.57\% & 16.33 \\
    22 & MU-Glioma-Post & 3d & \textbf{70.43\%} & \textbf{7.47} & 68.61\% & 8.32 & 67.82\% & 8.32 & 71.59\% & 7.24 \\
    24 & PediMS & 3d & \textbf{80.28\%} & 3.50 & 77.85\% & 6.40 & 79.27\% & \textbf{3.41} & 79.49\% & 3.54 \\
    25 & PLC-CECT & 3d & 57.30\% & 75.41 & \textbf{61.61\%} & \textbf{59.58} & 55.44\% & 85.35 & 53.22\% & 107.41 \\
    29 & TOM500 & 3d & \textbf{92.10\%} & \textbf{1.16} & 89.74\% & 2.08 & 91.50\% & 1.19 & 92.22\% & 1.14 \\
    30 & TRUSTED & 3d & 63.66\% & 39.23 & 62.13\% & 40.58 & \textbf{75.01\%} & \textbf{32.26} & 62.13\% & 40.58 \\
    31 & UT-EndoMRI & 3d & 74.05\% & 17.50 & 73.55\% & 13.34 & \textbf{77.03\%} & \textbf{10.74} & 75.10\% & 12.27 \\
    \midrule
    \textbf{Mean} & -- & -- & 63.43\% & 36.25 & \textbf{66.34\%} & \textbf{32.71} & 66.55\% & 36.24 & -- & -- \\
    \bottomrule
    \end{tabular}
\end{table}

\subsection{Evaluation Design Parameters}

\begin{table}[t]
    \centering
    \small
    \setlength{\tabcolsep}{6pt}
    \caption{Evaluation design used in the current benchmark analysis.}
    \label{tab:evaluation-design}
    \begin{tabular}{l r}
    \toprule
    \textbf{Item} & \textbf{Value} \\
    \midrule
    Benchmark datasets & 31 \\
    Validation datasets & 5 \\
    Blind-test datasets & 26 \\
    External papers & 90 \\
    Internal papers & 40 \\
    Total papers & 130 \\
    Junior reviewers (summary track) & 10 \\
    Senior reviewers (summary track) & 5 \\
    Papers per reviewer (summary track) & 30 \\
    Total human review assignments & 450 \\
    Average reviews per paper & 3.46 \\
    Human scoring dimensions & 4 \\
    Summary-based AI models & 5 \\
    Full-manuscript internal papers (blind / val.) & 32 / 8 \\
    \bottomrule
    \end{tabular}
\end{table}

\subsection{Detailed Benchmark Composition}
\label{app:benchmark-composition}

Table~\ref{tab:benchmark-full-composition} provides a venue-level breakdown of the full evaluation benchmark. Public papers were collected from WoS 2025 records retrieved with the keyword query \emph{``medical image segmentation''}, screened for task relevance, filtered to remove multidisciplinary venues, and then sampled uniformly with five papers per retained journal. The internal pool is listed separately because it is not organized by journal, but by experiment-generated manuscript artifacts.

\begin{table*}[t]
    \centering
    \scriptsize
    \setlength{\tabcolsep}{4pt}
    \caption{Detailed composition of the evaluation benchmark. Each retained public journal contributes five papers, yielding 90 public papers in total. The internal comparison pool contributes 40 papers.}
    \label{tab:benchmark-full-composition}
    \resizebox{\textwidth}{!}{%
    \begin{tabular}{c l l r l}
    \toprule
    \textbf{Group} & \textbf{Venue / Source} & \textbf{Definition} & \textbf{\# Papers} & \textbf{Role in Benchmark} \\
    \midrule
    T1 & IEEE Transactions on Medical Imaging & Top-tier medical image analysis venue & 5 & External benchmark \\
    T1 & Medical Image Analysis & Top-tier medical image analysis venue & 5 & External benchmark \\
    \midrule
    T2 & IEEE Journal of Biomedical and Health Informatics & JCR Q1 specialty journal & 5 & External benchmark \\
    T2 & Computerized Medical Imaging and Graphics & JCR Q1 specialty journal & 5 & External benchmark \\
    T2 & Biomedical Signal Processing and Control & JCR Q1 specialty journal & 5 & External benchmark \\
    T2 & Computer Methods and Programs in Biomedicine & JCR Q1 specialty journal & 5 & External benchmark \\
    T2 & Medical Physics & JCR Q1 specialty journal & 5 & External benchmark \\
    T2 & Artificial Intelligence in Medicine & JCR Q1 specialty journal & 5 & External benchmark \\
    T2 & Physics in Medicine and Biology & JCR Q1 specialty journal & 5 & External benchmark \\
    \midrule
    T3 & Bioengineering-Basel & Remaining field-specific journal & 5 & External benchmark \\
    T3 & Biomedical Physics \& Engineering Express & Remaining field-specific journal & 5 & External benchmark \\
    T3 & BMC Medical Imaging & Remaining field-specific journal & 5 & External benchmark \\
    T3 & International Journal of Computer Assisted Radiology and Surgery & Remaining field-specific journal & 5 & External benchmark \\
    T3 & International Journal of Imaging Systems and Technology & Remaining field-specific journal & 5 & External benchmark \\
    T3 & Journal of Imaging Informatics in Medicine & Remaining field-specific journal & 5 & External benchmark \\
    T3 & Journal of Medical Imaging & Remaining field-specific journal & 5 & External benchmark \\
    T3 & Medical \& Biological Engineering \& Computing & Remaining field-specific journal & 5 & External benchmark \\
    T3 & Quantitative Imaging in Medicine and Surgery & Remaining field-specific journal & 5 & External benchmark \\
    \midrule
    Internal & Experiment-generated papers & Manuscripts produced by \camyla across two runs & 40 & Internal comparison pool \\
    \midrule
    \textbf{Total} & \textbf{Full benchmark} & \textbf{90 public papers + 40 internal papers} & \textbf{130} & \textbf{Final evaluation pool} \\
    \bottomrule
    \end{tabular}%
    }
\end{table*}

\subsection{Per-Dimension Manuscript Quality Scores}
\label{app:quality-dimensions}

\begin{figure*}[t]
    \centering
    \includegraphics[width=\linewidth]{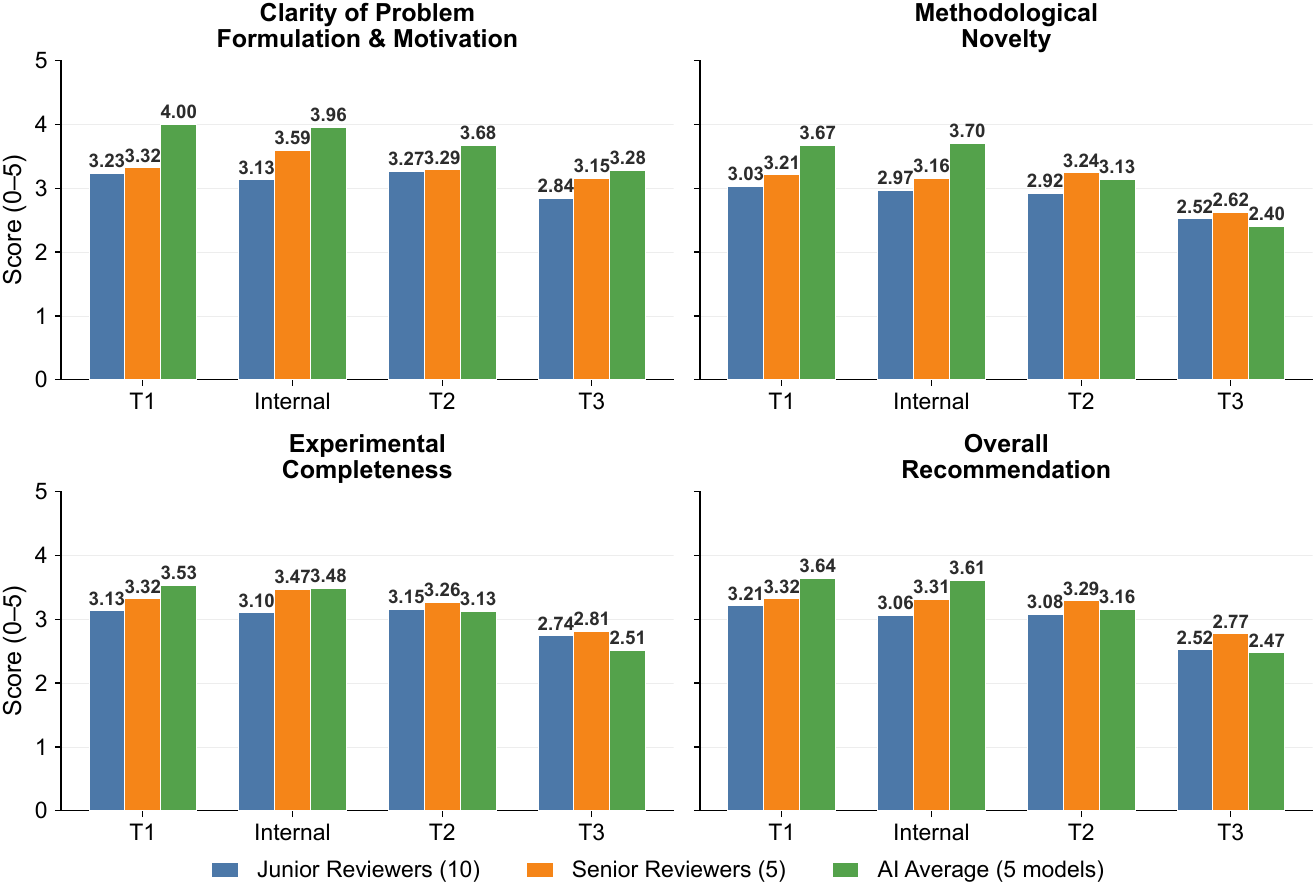}
    \caption{Per-dimension manuscript quality scores across evaluation tracks. Four panels show (a)~clarity of problem formulation and motivation, (b)~methodological novelty, (c)~experimental completeness, and (d)~overall recommendation. Each panel reports mean scores stratified by paper tier (T1, T2, T3) and \camyla-generated internal papers, for three reviewer groups: junior human reviewers, senior human reviewers, and the five-model AI average. Error bars indicate standard error of the mean.}
    \label{fig:quality-dimensions}
\end{figure*}

Figure~\ref{fig:quality-dimensions} provides a fine-grained breakdown of manuscript quality across the four individual scoring dimensions. The per-dimension pattern is consistent with the aggregate results in \S\ref{sec:quality}: \camyla-generated papers score at or above the T2 level on all four dimensions by senior human reviewers, and at or above the T1 level by AI evaluators. Methodological novelty (panel~b) shows the smallest gap between internal and T1 papers, suggesting that the system's literature-grounded proposal generation produces research directions comparable in novelty to those appearing in top-tier venues. Experimental completeness (panel~c) is the dimension on which internal papers most consistently match T1, reflecting the system's structured ablation and multi-metric evaluation pipeline.

\subsection{Baseline Bank Architecture Compatibility}

\begin{table}[t]
    \centering
    \small
    \setlength{\tabcolsep}{4pt}
    \caption{Baseline bank architecture compatibility. Each architecture is marked with the configurations it supports.}
    \label{tab:baseline-bank-compatibility}
    \begin{tabular}{l c c l}
    \toprule
    \textbf{Architecture} & \textbf{2D} & \textbf{3D} & \textbf{Family} \\
    \midrule
    nnU-Net        & \checkmark & \checkmark & Convolutional \\
    SwinUNETR      & \checkmark & \checkmark & Transformer \\
    SegResNet      & \checkmark & \checkmark & Convolutional \\
    U-Net++        & \checkmark & \checkmark & Convolutional \\
    U-Mamba        & \checkmark & \checkmark & State-space \\
    \midrule
    MedNeXt        &            & \checkmark & Convolutional \\
    3D UX-Net      &            & \checkmark & Transformer \\
    UNETR          &            & \checkmark & Transformer \\
    nnFormer       &            & \checkmark & Transformer \\
    STU-Net        &            & \checkmark & Convolutional \\
    \midrule
    TransUNet      & \checkmark &            & Transformer \\
    UTNet          & \checkmark &            & Transformer \\
    SwinUMamba     & \checkmark &            & State-space \\
    UKAN           & \checkmark &            & Convolutional \\
    \bottomrule
    \end{tabular}
\end{table}

\section{Statistical Significance Analysis}
\label{app:statistical-significance}

This section provides two complementary statistical analyses to quantify the robustness of \camyla's results: (1)~binomial tests on win counts with confidence intervals, and (2)~per-sample Wilcoxon signed-rank tests for individual experiments.

\subsection{Binomial Test on Win Counts}

The paper's primary claim concerns success rates. We test whether the observed win counts are significantly above chance using one-sided binomial tests under two null hypotheses: $p_0 = 0.5$ (conservative: 50/50 chance of exceeding a tuned baseline by ${>}0.5$~pp) and $p_0 = 0.3$ (realistic: a randomly initialized method is unlikely to exceed a carefully tuned baseline).

\begin{table}[h]
    \centering
    \small
    \caption{Binomial test results (one-sided, $H_1: p > p_0$) and Wilson 95\% confidence intervals for win rates.}
    \label{tab:binomial-tests}
    \begin{tabular}{l c r r c}
    \toprule
    \textbf{Configuration} & \textbf{Wins} & $p$ ($p_0{=}0.5$) & $p$ ($p_0{=}0.3$) & \textbf{Wilson 95\% CI} \\
    \midrule
    \camylaS (all 31)    & 22/31 & \textbf{0.015}  & $3.0 \times 10^{-6}$ & [53.4\%, 83.9\%] \\
    \camylaD (all 31)    & 18/31 & 0.237            & \textbf{0.001}       & [40.8\%, 73.6\%] \\
    \camylaS (blind 26)  & 18/26 & \textbf{0.038}  & $2.4 \times 10^{-5}$ & [50.0\%, 83.5\%] \\
    \camylaD (blind 26)  & 14/26 & 0.423            & \textbf{0.007}       & [35.5\%, 71.2\%] \\
    Union (all 31)       & 24/31 & \textbf{0.002}  & ---                  & [60.2\%, 88.6\%] \\
    Union (blind 26)     & 20/26 & \textbf{0.004}  & ---                  & [57.8\%, 89.4\%] \\
    \bottomrule
    \end{tabular}
\end{table}

Under the realistic null ($p_0 = 0.3$), both individual runs are highly significant ($p \leq 0.007$). Under the conservative null ($p_0 = 0.5$), \camylaS alone reaches $p = 0.015$ and the union reaches $p = 0.002$. The Wilson confidence intervals confirm that even the lower bound of the union win rate exceeds 60\%.

\subsection{Per-Sample Wilcoxon Signed-Rank Tests}

To assess significance at the individual experiment level, we perform paired Wilcoxon signed-rank tests at the per-sample level for all 40 winning experiments. For 3D datasets, Dice and HD95 are computed on every axial slice containing foreground in the reference segmentation, yielding hundreds to thousands of paired samples per dataset. For 2D datasets, per-case metrics are used directly. This per-slice approach follows standard practice in medical image segmentation evaluation~\citep{isensee2021nnu}.

Tables~\ref{tab:wilcoxon-rund} and~\ref{tab:wilcoxon-runs} report the full results for both runs. Across all 40 winning experiments, 70\% show statistically significant improvement ($p < 0.05$) on at least one metric, and 50\% are significant on both Dice and HD95.

\begin{table*}[t]
    \centering
    \scriptsize
    \setlength{\tabcolsep}{3pt}
    \renewcommand{\arraystretch}{0.82}
    \caption{Per-sample Wilcoxon signed-rank tests for \camylaD winning experiments. $\dagger$\,=\,validation. $N$\,=\,paired samples. \textbf{***}\,$p{<}0.001$, \textbf{**}\,$p{<}0.01$, \textbf{*}\,$p{<}0.05$, n.s.\,=\,not significant.}
    \label{tab:wilcoxon-rund}
    \resizebox{\textwidth}{!}{%
    \begin{tabular}{c l c r r r r r c r c}
    \toprule
    \textbf{ID} & \textbf{Dataset} & \textbf{Cfg} & $N$ & \textbf{BL Dice} & \textbf{Best Dice} & $\Delta$\textbf{Dice} & $p$\textbf{(Dice)} & \textbf{Sig} & $p$\textbf{(HD95)} & \textbf{Sig} \\
    \midrule
    1\textsuperscript{$\dagger$}  & GDMRI-CT    & 3D & 44    & 0.7233 & 0.7387 & +0.0153 & 5.07e-01 & n.s. & 9.75e-01 & n.s. \\
    3\textsuperscript{$\dagger$}  & PNPC        & 3D & 325   & 0.6063 & 0.6134 & +0.0071 & 1.38e-01 & n.s. & 3.85e-01 & n.s. \\
    4\textsuperscript{$\dagger$}  & AMSMC-HTM   & 3D & 251   & 0.8141 & 0.8298 & +0.0157 & 6.26e-09 & *** & 1.28e-06 & *** \\
    5\textsuperscript{$\dagger$}  & NLSTseg     & 3D & 1{,}446 & 0.2320 & 0.4686 & +0.2365 & 1.29e-66 & *** & 3.04e-11 & *** \\

    6   & SMRI-FB      & 3D & 1{,}242  & 0.8530 & 0.8667 & +0.0136 & 4.52e-40 & *** & 1.89e-22 & *** \\
    7   & LMD-BM       & 3D & 383      & 0.4619 & 0.5151 & +0.0532 & 1.58e-15 & *** & 4.10e-04 & *** \\
    8   & BONBID2023   & 3D & 316      & 0.5585 & 0.5823 & +0.0237 & 5.15e-03 & **  & 3.00e-02 & * \\
    9   & BTXRD        & 2D & 374      & 0.4445 & 0.5017 & +0.0572 & 4.13e-12 & *** & 5.02e-03 & ** \\
    11  & CirrMRI600+  & 3D & 1{,}616  & 0.8179 & 0.8632 & +0.0453 & 1.77e-07 & *** & 1.23e-33 & *** \\
    14  & DERMA-OCTA   & 2D & 67       & 0.8391 & 0.8428 & +0.0037 & 4.14e-03 & **  & 8.06e-01 & n.s. \\
    16  & FOVEA        & 2D & 8        & 0.7142 & 0.8191 & +0.1049 & 1.95e-01 & n.s. & 8.13e-01 & n.s. \\
    18  & HRUS-MBT     & 3D & 62       & 0.4269 & 0.5055 & +0.0787 & 3.12e-04 & *** & 1.92e-03 & ** \\
    19  & LapGC-KVAD   & 2D & 251      & 0.5277 & 0.5698 & +0.0421 & 2.97e-06 & *** & 7.61e-03 & ** \\
    20  & LongCIU      & 3D & 18       & 0.7345 & 0.7573 & +0.0228 & 4.68e-01 & n.s. & 3.85e-02 & * \\
    21  & MSLesSeg     & 3D & 1{,}742  & 0.6957 & 0.7069 & +0.0113 & 1.15e-03 & **  & 5.78e-01 & n.s. \\
    25  & PLC-CECT     & 3D & 2{,}754  & 0.5322 & 0.6161 & +0.0839 & 1.28e-52 & *** & 3.34e-72 & *** \\
    26  & PW-BALFC     & 2D & 421      & 0.7666 & 0.7697 & +0.0031 & 3.67e-01 & n.s. & 2.60e-01 & n.s. \\
    28  & STS-Tooth    & 2D & 170      & 0.9367 & 0.9435 & +0.0067 & 1.47e-13 & *** & 1.64e-02 & * \\
    \midrule
    \multicolumn{8}{l}{\textbf{Summary}: Dice significant on 12/18, HD95 significant on 11/18, either metric significant on \textbf{13/18} (72\%).} \\
    \bottomrule
    \end{tabular}
    }
\end{table*}

\begin{table*}[t]
    \centering
    \scriptsize
    \setlength{\tabcolsep}{3pt}
    \renewcommand{\arraystretch}{0.82}
    \caption{Per-sample Wilcoxon signed-rank tests for \camylaS winning experiments. Notation follows Table~\ref{tab:wilcoxon-rund}.}
    \label{tab:wilcoxon-runs}
    \resizebox{\textwidth}{!}{%
    \begin{tabular}{c l c r r r r r c r c}
    \toprule
    \textbf{ID} & \textbf{Dataset} & \textbf{Cfg} & $N$ & \textbf{BL Dice} & \textbf{Best Dice} & $\Delta$\textbf{Dice} & $p$\textbf{(Dice)} & \textbf{Sig} & $p$\textbf{(HD95)} & \textbf{Sig} \\
    \midrule
    1\textsuperscript{$\dagger$}  & GDMRI-CT    & 3D & 44      & 0.7233 & 0.7277 & +0.0044 & 9.52e-01 & n.s. & 3.20e-01 & n.s. \\
    3\textsuperscript{$\dagger$}  & PNPC        & 3D & 325     & 0.6063 & 0.6195 & +0.0133 & 1.64e-01 & n.s. & 4.43e-01 & n.s. \\
    4\textsuperscript{$\dagger$}  & AMSMC-HTM   & 3D & 251    & 0.8141 & 0.8289 & +0.0148 & 7.47e-06 & *** & 1.15e-07 & *** \\
    5\textsuperscript{$\dagger$}  & NLSTseg     & 3D & 1{,}446 & 0.2320 & 0.4126 & +0.1805 & $<$1e-300 & *** & 4.95e-07 & *** \\

    6   & SMRI-FB      & 3D & 1{,}242  & 0.8530 & 0.8590 & +0.0059 & 6.77e-14 & *** & 1.97e-07 & *** \\
    7   & LMD-BM       & 3D & 383      & 0.4619 & 0.5101 & +0.0482 & 1.05e-08 & *** & 2.56e-04 & *** \\
    8   & BONBID2023   & 3D & 316      & 0.5585 & 0.5641 & +0.0055 & 6.50e-11 & *** & 3.87e-04 & *** \\
    9   & BTXRD        & 2D & 374      & 0.4445 & 0.4442 & $-$0.0003 & 8.18e-05 & *** & 2.60e-02 & * \\
    10  & BUS-UCLM     & 2D & 134      & 0.3666 & 0.3618 & $-$0.0048 & 4.80e-01 & n.s. & 6.55e-01 & n.s. \\
    11  & CirrMRI600+  & 3D & 1{,}616  & 0.8179 & 0.8506 & +0.0327 & 3.12e-09 & *** & 1.69e-02 & * \\
    12  & CPAISD       & 3D & 404      & 0.2732 & 0.2908 & +0.0176 & 1.81e-01 & n.s. & 9.80e-15 & *** \\
    14  & DERMA-OCTA   & 2D & 67       & 0.8391 & 0.8421 & +0.0030 & 2.34e-02 & *   & 4.48e-03 & ** \\
    15  & Endoscapes   & 2D & 69       & 0.4753 & 0.4813 & +0.0060 & 4.27e-01 & n.s. & 2.38e-01 & n.s. \\
    18  & HRUS-MBT     & 3D & 62       & 0.4269 & 0.4783 & +0.0514 & 4.21e-04 & *** & 3.15e-03 & ** \\
    20  & LongCIU      & 3D & 18       & 0.7345 & 0.7447 & +0.0102 & 6.65e-02 & n.s. & 2.29e-01 & n.s. \\
    21  & MSLesSeg     & 3D & 1{,}742  & 0.6957 & 0.6981 & +0.0024 & 7.91e-01 & n.s. & 7.94e-05 & *** \\
    24  & PediMS       & 3D & 102      & 0.7949 & 0.7940 & $-$0.0009 & 7.70e-02 & n.s. & 7.48e-01 & n.s. \\
    25  & PLC-CECT     & 3D & 2{,}754  & 0.5322 & 0.5544 & +0.0222 & 4.10e-07 & *** & 5.99e-25 & *** \\
    26  & PW-BALFC     & 2D & 421      & 0.7666 & 0.7687 & +0.0021 & 3.09e-01 & n.s. & 4.21e-02 & * \\
    28  & STS-Tooth    & 2D & 170      & 0.9367 & 0.9371 & +0.0004 & 6.79e-01 & n.s. & 7.83e-02 & n.s. \\
    30  & TRUSTED      & 3D & 3{,}291  & 0.6213 & 0.7586 & +0.1373 & $<$1e-300 & *** & $<$1e-300 & *** \\
    31  & UT-EndoMRI   & 3D & 159      & 0.7510 & 0.7703 & +0.0193 & 2.76e-01 & n.s. & 4.78e-02 & * \\
    \midrule
    \multicolumn{8}{l}{\textbf{Summary}: Dice significant on 12/22, HD95 significant on 13/22, either metric significant on \textbf{15/22} (68\%).} \\
    \bottomrule
    \end{tabular}
    }
\end{table*}

\paragraph{Aggregate summary.}
Table~\ref{tab:wilcoxon-aggregate} consolidates the significance counts across both runs. After Bonferroni correction across all 40 experiments ($\alpha = 0.05/40 = 0.00125$), 20 experiments remain significant on at least one metric.

\begin{table}[h]
    \centering
    \small
    \caption{Aggregate per-sample significance counts across both runs.}
    \label{tab:wilcoxon-aggregate}
    \begin{tabular}{l c c c}
    \toprule
    & \textbf{Run D} (18) & \textbf{Run S} (22) & \textbf{Total} (40) \\
    \midrule
    Dice significant    & 12 (67\%) & 12 (55\%) & 24 (60\%) \\
    HD95 significant    & 11 (61\%) & 13 (59\%) & 24 (60\%) \\
    Either metric       & 13 (72\%) & 15 (68\%) & 28 (70\%) \\
    Both metrics        & 10 (56\%) & 10 (45\%) & 20 (50\%) \\
    \bottomrule
    \end{tabular}
\end{table}

\paragraph{Discussion.}
The non-significant cases follow three predictable patterns: (a)~near-saturated baselines with marginal gains (e.g., PW-BALFC: $\Delta$Dice\,=\,+0.3~pp on a 76.7\% baseline), (b)~datasets with high inter-case variance (e.g., GDMRI-CT with heterogeneous multi-sequence brain MRI), and (c)~2D datasets with very few test samples where even per-case testing lacks statistical power (e.g., FOVEA: $n{=}8$, $p{=}0.195$ despite $\Delta$Dice\,=\,+10.5~pp). For 3D datasets, adjacent axial slices within a volume are spatially correlated, which means the Wilcoxon test's independence assumption is partially violated and the reported p-values may be slightly liberal. However, this per-slice evaluation follows standard practice in the medical segmentation literature, and 20 experiments survive the stringent Bonferroni correction.

The cross-run consistency reported in \S\ref{sec:stability} provides complementary variance evidence: for the 16 datasets won by both runs, the Pearson correlation of best Dice is $r{=}0.978$ and the mean inter-run absolute difference is 1.56~pp. The two runs vary the idea generator LLM---a strictly stronger perturbation than varying training random seeds---yet produce highly concordant outcomes.

\section{Computational Cost}
\label{app:computational-cost}

Table~\ref{tab:gpu-hours-summary} summarizes the computational cost of both independent runs. All figures are reported under the unified early-stop policy described in Appendix~\ref{app:stage-transitions}: once a proposal's best Dice exceeds the baseline, subsequent proposals are not executed. Under this policy, the 62 experiments consume a total of 5{,}461.9 GPU-hours (227.6 GPU-days) on NVIDIA RTX 4090 48\,GB GPUs, with a mean of 88.1 GPU-hours per experiment. Experiments that successfully exceed the baseline consume substantially fewer resources (mean 64.9\,h) than those that exhaust the full budget without success (mean 130.3\,h), because the early-stop mechanism terminates search once improvement is confirmed.

\begin{table}[t]
    \centering
    \small
    \setlength{\tabcolsep}{5pt}
    \caption{Computational cost summary across both runs (62 experiments total). GPU-hours are measured on NVIDIA RTX 4090 48\,GB GPUs under the unified early-stop policy.}
    \label{tab:gpu-hours-summary}
    \begin{tabular}{l r}
    \toprule
    \textbf{Metric} & \textbf{Value} \\
    \midrule
    Total experiments & 62 \\
    Total GPU-hours & 5{,}461.9\,h \\
    Total GPU-days & 227.6\,d \\
    Mean GPU-hours / experiment & 88.1\,h \\
    Median GPU-hours / experiment & 34.2\,h \\
    \midrule
    Baseline-beating experiments (40) & 2{,}595.2\,h (mean 64.9\,h) \\
    Non-beating experiments (22) & 2{,}866.7\,h (mean 130.3\,h) \\
    \bottomrule
    \end{tabular}
\end{table}

Table~\ref{tab:gpu-hours-per-dataset} reports the per-dataset GPU-hours aggregated across both attempts. The variation is large (6.3\,h to 730.8\,h), driven primarily by dataset volume size and the number of proposals required before success.

\begin{table*}[t]
    \centering
    \scriptsize
    \setlength{\tabcolsep}{2pt}
    \renewcommand{\arraystretch}{0.82}
    \caption{Per-dataset GPU-hours under the unified early-stop policy. Each dataset has two attempts (one per run). $\dagger$\,=\,validation. \textbf{Tot.}\,=\,sum across both attempts. \textbf{Sav.}\,=\,GPU-hours saved by early stopping. Total: 5{,}461.9\,h (saved 1{,}930.5\,h).}
    \label{tab:gpu-hours-per-dataset}
    \resizebox{0.75\textwidth}{!}{%
    \begin{tabular}{r l r r r r @{\hskip 6pt} r l r r r r}
    \toprule
    \textbf{ID} & \textbf{Dataset} & \textbf{Att.1} & \textbf{Att.2} & \textbf{Tot.} & \textbf{Sav.}
    & \textbf{ID} & \textbf{Dataset} & \textbf{Att.1} & \textbf{Att.2} & \textbf{Tot.} & \textbf{Sav.} \\
    \midrule
    1$^\dagger$  & GDMRI-CT       & 33.6  & 9.6   & 43.2   & 52.3
    & 17 & Fundus-AVSeg   & 6.5   & 31.2  & 37.8   & 33.0  \\
    2$^\dagger$  & MRE-BSA        & 81.3  & 33.5  & 114.8  & 0.0
    & 18 & HRUS-MBT       & 243.5 & 9.5   & 253.0  & 144.3 \\
    3$^\dagger$  & PNPC           & 34.7  & 22.8  & 57.4   & 42.0
    & 19 & LapGC-KVAD     & 188.1 & 235.9 & 423.9  & 153.9 \\
    4$^\dagger$  & AMSMC-HTM      & 45.0  & 21.4  & 66.4   & 109.6
    & 20 & LongCIU        & 42.7  & 9.4   & 52.1   & 50.0  \\
    5$^\dagger$  & NLSTseg        & 26.5  & 32.6  & 59.1   & 61.0
    & 21 & MSLesSeg       & 51.9  & 12.4  & 64.3   & 80.7  \\

    6  & SMRI-FB        & 48.4  & 23.1  & 71.5   & 78.9
    & 22 & MU-Glioma      & 474.7 & 256.1 & 730.8  & 0.0   \\
    7  & LMD-BM         & 68.0  & 150.7 & 218.8  & 262.0
    & 23 & OCT5k          & 440.8 & 219.6 & 660.4  & 0.0   \\
    8  & BONBID2023     & 20.0  & 25.1  & 45.2   & 61.5
    & 24 & PediMS         & 38.9  & 24.2  & 63.1   & 0.0   \\
    9  & BTXRD          & 279.5 & 210.0 & 489.6  & 180.4
    & 25 & PLC-CECT       & 220.7 & 191.5 & 412.2  & 150.3 \\
    10 & BUS-UCLM       & 189.3 & 141.5 & 330.8  & 0.0
    & 26 & PW-BALFC       & 208.8 & 153.7 & 362.5  & 122.9 \\
    11 & CirrMRI600+    & 20.9  & 13.1  & 34.0   & 60.4
    & 27 & SEA-SIS        & 33.5  & 41.2  & 74.7   & 0.0   \\
    12 & CPAISD         & 14.5  & 29.0  & 43.5   & 35.0
    & 28 & STS-Tooth      & 9.7   & 6.7   & 16.4   & 35.7  \\
    13 & DenPAR         & 29.9  & 6.3   & 36.2   & 79.4
    & 29 & TOM500         & 64.8  & 50.8  & 115.6  & 0.0   \\
    14 & DERMA-OCTA     & 11.8  & 23.5  & 35.4   & 38.3
    & 30 & TRUSTED        & 181.0 & 185.9 & 366.9  & 33.1  \\
    15 & Endoscapes     & 45.1  & 25.2  & 70.4   & 16.7
    & 31 & UT-EndoMRI     & 34.2  & 16.5  & 50.6   & 5.8   \\
    16 & FOVEA          & 32.0  & 29.6  & 61.5   & 43.2
    &    &                &       &       &        &       \\
    \bottomrule
    \end{tabular}%
    }
\end{table*}

\section{CamylaTrace-232k: Research Trajectory Dataset}
\label{app:camylatrace}

To support future research on autonomous scientific agents, we publicly release \textbf{CamylaTrace-232k}, the experimental-discovery trajectories produced during both independent \camyla runs on the 31 \camylabench datasets. CamylaTrace-232k captures the core of the autonomous research process---iterative experimentation driven by literature-grounded proposals, diagnostic feedback, and quality-weighted search---as fine-grained, timestamped event logs. The dataset focuses exclusively on the experimental-discovery phase, which contains the agent's hypothesis-driven architectural exploration and is the phase most relevant to studying autonomous scientific reasoning. The ``232k'' refers to the 232{,}499 agent events in the released corpus.

\subsection{Scale and Coverage}

The dataset spans 62 independent experiments conducted between 2026-02-23 and 2026-03-16, covering all 31 datasets with two attempts each (one per run). Table~\ref{tab:camylatrace-scale} summarizes the key dimensions. The 147 experimental-discovery directories across the 62 experiments contain three categories of artifacts: agent event logs, per-session summaries, and generated code files, totaling approximately 700\,MB.

\begin{table}[t]
    \centering
    \small
    \setlength{\tabcolsep}{5pt}
    \caption{Scale of the CamylaTrace-232k dataset. All counts are restricted to experimental-discovery directories from the 62 main experiments.}
    \label{tab:camylatrace-scale}
    \begin{tabular}{l r}
    \toprule
    \textbf{Dimension} & \textbf{Count} \\
    \midrule
    Datasets covered & 31 \\
    Independent experiments & 62 \\
    Experimental-discovery directories & 147 \\
    \midrule
    Agent event files (.jsonl) & 2{,}865 \\
    Agent session summaries (.md) & 3{,}953 \\
    Generated experiment code files (.py) & 1{,}343 \\
    \midrule
    Total agent events & 232{,}499 \\
    Total dataset size & ${\sim}$700\,MB \\
    \bottomrule
    \end{tabular}
\end{table}

\subsection{Directory Structure}

The released dataset is organized by dataset ID and experiment timestamp. Each experimental-discovery directory contains three artifact subdirectories:

\begin{lstlisting}[basicstyle=\ttfamily\scriptsize, frame=single, columns=fullflexible, breaklines=true, xleftmargin=1em]
camylatrace-232k/
  <dataset_id>/
    <experiment_timestamp>/
      stage_2_creative_research_<N>_proposal_<M>/
        events/              # Timestamped agent event logs
          openhands_events_<ts>.jsonl
        summaries/           # Per-session summaries
          openhands_summary_<ts>.md
        codes/               # Generated code at each tree node
          <ts>_<hash>_experiment_code.py
\end{lstlisting}

Each \texttt{events/} directory contains one JSONL file per coding session, capturing the complete agent interaction loop. Each \texttt{summaries/} directory contains a Markdown summary produced at the end of each session. Each \texttt{codes/} directory contains the Python experiment code generated at each search tree node, with filenames encoding the timestamp and a content hash.

\subsection{Event Schema}

Each line in the JSONL event files is a JSON object with the following fields:

\begin{lstlisting}[basicstyle=\ttfamily\scriptsize, frame=single, columns=fullflexible, breaklines=true, xleftmargin=1em]
{
  "timestamp": "2026-03-09T01:15:15.772208",
  "event_type": "ActionEvent",
  "event_str": "Agent edits network architecture",
  "llm_message": {
    "role": "assistant",
    "content_preview": "I will modify the encoder...",
    "content_length": 27750
  }
}
\end{lstlisting}

Four event types capture the agent interaction loop:

\begin{itemize}[leftmargin=1.5em, itemsep=2pt]
    \item \texttt{SystemPromptEvent} --- System prompt containing tool definitions, dataset context, and dynamic state (current best metric, remaining budget).
    \item \texttt{MessageEvent} --- User-role messages injecting experiment instructions, diagnostic feedback reports, and cycle-level memory.
    \item \texttt{ActionEvent} --- Agent reasoning traces (chain-of-thought) and tool invocations (code edits, bash commands, file reads, task tracking).
    \item \texttt{ObservationEvent} --- Tool execution results returned to the agent, including training logs, evaluation metrics, and compilation errors.
\end{itemize}

\subsection{Comparison with Existing Trajectory Datasets}

Table~\ref{tab:camylatrace-comparison} positions CamylaTrace-232k relative to existing agent trajectory datasets. CamylaTrace-232k is distinguished by its coverage of iterative scientific experimentation---hypothesis-driven architectural exploration with diagnostic feedback and quality-weighted search---rather than isolated coding or task-completion episodes.

\begin{table}[t]
    \centering
    \small
    \setlength{\tabcolsep}{4pt}
    \caption{Comparison of CamylaTrace-232k with existing agent trajectory datasets.}
    \label{tab:camylatrace-comparison}
    \begin{tabular}{l l l r r}
    \toprule
    \textbf{Dataset} & \textbf{Domain} & \textbf{Scope} & \textbf{Events} & \textbf{Code} \\
    \midrule
    SWE-bench & Software eng. & Bug fixes & Patch-level & Yes \\
    ML-bench & Machine learning & Task completion & Limited & Yes \\
    CamylaTrace-232k & Scientific research & Iterative research & 232K & 1{,}343 \\
    \bottomrule
    \end{tabular}
\end{table}

\subsection{Research Value}

CamylaTrace-232k enables several lines of investigation:

\begin{enumerate}[leftmargin=1.5em, itemsep=2pt]
    \item \emph{Agent reasoning analysis.} The 2{,}865 coding sessions span 40 successful and 22 unsuccessful experiments across 31 diverse tasks. Comparing reasoning patterns between baseline-beating and non-beating trajectories can reveal what distinguishes productive exploration from unproductive iteration.
    \item \emph{Scientific coding behavior.} The 1{,}343 code snapshots, each paired with execution outcomes (Dice, HD95, or error type), form a large-scale corpus of iteratively refined deep learning implementations suitable for studying code evolution and debugging strategies.
    \item \emph{Agent training and evaluation.} The 232{,}499 timestamped events with paired outcomes provide supervised signal for training research-capable agents or constructing benchmarks for scientific agent evaluation.
\end{enumerate}

\section{QWBE Quality Normalization}
\label{app:qwbe-quality-normalization}

Section~\ref{sec:planning} defines the PUCT-based branch scoring rule (Eq.~\ref{eq:puct-score}) in terms of the branch quality $Q_i \in [-1, +1]$, but does not specify how $Q_i$ is computed from raw evaluation metrics. This appendix provides the complete normalization procedure.

\paragraph{Per-Trial Normalization.}
The branch quality $Q_i$ is the mean over normalized quality values $q(t)$ of all non-stale trial nodes within branch $b_i$. For a trial $t$ with raw evaluation metric $m(t)$ and baseline metric $m_0$, the normalization is:
\begin{equation}
    q(t) = \begin{cases}
        \dfrac{m(t) - m_0}{\max(1 - m_0,\; \varepsilon)} & \text{if } m(t) \geq m_0, \\[8pt]
        -\min\!\left(1,\; \left(\dfrac{m_0 - m(t)}{\max(m_0,\; \varepsilon)}\right)^{\!e}\right) & \text{if } m(t) < m_0,
    \end{cases}
    \label{eq:q-norm}
\end{equation}
where $\varepsilon$ is a numerical stability constant and $e$ is the below-penalization exponent (default 1.0, yielding linear mapping in both regimes). The upper branch maps improvements into $[0, +1]$, where the denominator $\max(1 - m_0, \varepsilon)$ normalizes by the maximum possible improvement above the baseline. The lower branch maps underperformance into $[-1, 0]$, where $e$ controls the penalty shape: setting $e < 1$ strengthens penalization for underperforming trials by pulling their normalized quality closer to $-1$, while $e > 1$ softens the penalty.

\paragraph{Handling Execution Errors.}
Trials that terminate with execution errors (e.g., shape mismatches, out-of-memory crashes, or NaN divergence) produce no valid metric. For such nodes, $q(t)$ is estimated by inheriting the normalized quality of the nearest valid ancestor and subtracting a fixed correction $\delta_{\mathrm{buggy}}$ (default 0.2):
\begin{equation}
    q(t_{\mathrm{error}}) = q(t_{\mathrm{ancestor}}) - \delta_{\mathrm{buggy}}.
    \label{eq:q-buggy}
\end{equation}
This inheritance rule reflects the observation that an erroneous trial typically represents a localized implementation failure near a functioning parent node rather than a complete collapse of the research direction. Assigning $q = -1$ to all error nodes would systematically undervalue the branch containing them, discouraging QWBE from revisiting a direction that may only require a minor code fix.

\paragraph{Leaf Selection Within a Branch.}
Once QWBE selects a branch via Eq.~\ref{eq:puct-score}, it chooses the best available leaf node within that branch for expansion. If the selected leaf corresponds to a failed execution within the allowed debugging depth (default 3 consecutive errors), it is submitted for a repair attempt; otherwise, it is submitted for an improvement attempt that generates a refined implementation conditioned on the current best result in the branch. This two-mode leaf selection ensures that promising branches are not abandoned due to transient implementation errors while preventing infinite debugging loops on fundamentally broken directions.

\section{Divergent Diagnostic Feedback: Real Diagnostic Examples}
\label{app:ddf-examples}

To illustrate how DDF operates in practice, we present two complete diagnostic reports from the Dataset~16 experiment (optic disc segmentation, UKAN baseline Dice = 0.7142). These examples are extracted verbatim from the system's diagnostic logs.

\subsection{Example 1: Failure Diagnosis (Cycle 1, Trial 2)}

\paragraph{Trigger.} Trial~2 achieves Dice = 0.6835, falling 3.07 percentage points below the baseline.

\paragraph{Reasoning Trace.}
\begin{quote}
\small
``The experiment successfully executed but the Dice score (0.6835) is significantly below the baseline (0.7142). Analysis of the implementation reveals several proposal gaps and simplifications made by the agent to pass the 1-epoch test. Specifically, the learnable hierarchy embedding which is central to the HTD module's motivation was omitted. Furthermore, the spatial pooling in the attention mechanism is likely too aggressive ($32{\times}32$), causing a loss of fine-grained boundary information.''
\end{quote}

\paragraph{Five-Suggestion Portfolio.}
Table~\ref{tab:ddf-example-1} lists the five suggestions produced by the diagnostic agent, spanning three categories.

\begin{table}[h]
    \centering
    \small
    \setlength{\tabcolsep}{4pt}
    \caption{DDF diagnostic report for Dataset~16, Cycle~1, Trial~2 (Dice = 0.6835 vs.\ baseline 0.7142).}
    \label{tab:ddf-example-1}
    \begin{tabular}{c l l p{8.5cm}}
    \toprule
    \textbf{\#} & \textbf{Category} & \textbf{Priority} & \textbf{Suggestion} \\
    \midrule
    1 & \textsc{Proposal\_Gap} & High & Implement the learnable hierarchy embedding ($\xi_h$) in the HTD module as described in the proposal. Currently the cross-attention only uses projected features; adding the hierarchy embedding to the Key ($K = T_e^l + \xi_h$) will allow the model to learn the spatial/structural relationship between the optic disc and cup. \\
    2 & \textsc{Architecture} & Medium & Replace the fixed $32{\times}32$ spatial pooling in the HTD module with a higher resolution ($64{\times}64$). The current implementation significantly downsamples high-resolution fundus features before attention, likely losing fine boundary details. \\
    3 & \textsc{Code\_Fix} & High & Fix the SADCB fusion logic. The proposal specifies a channel-wise gating mechanism ($\alpha = \sigma(\mathrm{Conv}_{1\times1}(X))$) to fuse parallel branches, but the current code uses a simplified selection or summation. \\
    4 & \textsc{Architecture} & Medium & Add a residual connection around the HTD module in the decoder. A residual path for the original skip-connection information alongside the diffused tokens would stabilize training. \\
    5 & \textsc{Hyperparameter} & Medium & Increase \texttt{base\_channels} from 32 to 64 and adjust \texttt{embed\_dim} in the HTD module to 128. \\
    \bottomrule
    \end{tabular}
\end{table}

\paragraph{Dual-Agent Divergence.}
Two competing coding agents independently consumed this diagnostic report and selected different subsets:
\begin{itemize}[leftmargin=1.5em, itemsep=2pt]
    \item \textbf{Agent A} prioritized Suggestions 1 + 2 (hierarchy embedding + higher pooling resolution), implementing the learnable hierarchy embedding as a query bias in cross-attention and raising the spatial resolution to $64{\times}64$. Result: \textbf{Dice = 0.7682} (new cycle best, +5.40~pp over baseline).
    \item \textbf{Agent B} pursued Suggestions 3 + 4 (fusion fix + residual connection), correcting the multi-branch gating and adding a residual path. Result: Dice = 0.6753 ($-$3.89~pp, recorded as counterfactual in cycle memory).
\end{itemize}
The same diagnostic event thus spawned two divergent branches: one discovered a strong improvement that a single prescriptive fix might have missed, while the other provided a negative signal that informed subsequent trials.

\subsection{Example 2: Failure Diagnosis (Cycle 3, Trial 2)}

\paragraph{Trigger.} Trial~2 achieves Dice = 0.6095, falling 10.46 percentage points below the baseline after the adaptive tokenization module was reduced to a fixed-grid pooling placeholder.

\paragraph{Reasoning Trace.}
\begin{quote}
\small
``The current implementation suffers from a significant performance gap ($-$0.1046 Dice) because the core innovations were heavily simplified during debugging. Specifically, the Resolution-Adaptive Token Formation (RATF) was implemented as a fixed-grid patch embedding rather than the complexity-aware clustering described in the proposal. The token reduction factor was set too high ($4{\times}$), leading to a loss of fine-grained spatial information.''
\end{quote}

\paragraph{Five-Suggestion Portfolio.}
Table~\ref{tab:ddf-example-2} lists the five suggestions, now targeting a more severe regression.

\begin{table}[h]
    \centering
    \small
    \setlength{\tabcolsep}{4pt}
    \caption{DDF diagnostic report for Dataset~16, Cycle~3, Trial~2 (Dice = 0.6095 vs.\ baseline 0.7142).}
    \label{tab:ddf-example-2}
    \begin{tabular}{c l l p{8.5cm}}
    \toprule
    \textbf{\#} & \textbf{Category} & \textbf{Priority} & \textbf{Suggestion} \\
    \midrule
    1 & \textsc{Proposal\_Gap} & High & Implement the Complexity Estimator in RATF using a $3{\times}3$ Sobel filter or local variance to generate a complexity map, then use this map to weight a Softmax over different pooling kernel sizes to achieve true adaptive tokenization. \\
    2 & \textsc{Architecture} & High & Convert the CATA block into a bottleneck style where the multi-scale attention output is concatenated with a $3{\times}3$ depthwise-separable convolution branch before the final projection, ensuring local spatial consistency. \\
    3 & \textsc{Code\_Fix} & Medium & Reduce the \texttt{token\_reduction} factor from 4 to 2 and increase the initial \texttt{hidden\_sizes} from 64 to 96. \\
    4 & \textsc{Architecture} & Medium & Replace the learned Positional Embedding with Conditional Positional Encoding (CPE) using a $3{\times}3$ depthwise convolution applied to the token grid. \\
    5 & \textsc{Proposal\_Gap} & Medium & Implement the Size-Biased Attention formula: $\mathrm{Attn} = \mathrm{Softmax}(QK^\top/\sqrt{d} + \log(\text{token\_sizes}) \cdot W_s)$, directly injecting scale information into the attention map as proposed. \\
    \bottomrule
    \end{tabular}
\end{table}

\paragraph{Outcome.}
After two additional diagnostic-guided iterations (Trials~3--4 still underperforming due to partial fixes), Trial~5 synthesized the core insight from Suggestion~1---implementing a gradient-magnitude-based complexity estimator for adaptive tokenization---and achieved Dice = 0.7988 (first significant baseline beat in Cycle~3). Trial~6 refined this to 0.8191 (overall best, +10.49~pp over baseline).

\subsection{Adaptive Diagnosis Evolution}

When a second diagnosis was triggered after Trial~4 (partial recovery to Dice = 0.6602), the diagnostic agent \emph{updated} its suggestion portfolio based on the new experimental evidence. The \textsc{Proposal\_Gap} category now targeted cross-scale attention paths (previously unmentioned in the first report), and the priority ordering shifted: the Size-Biased Attention suggestion was promoted from medium to high priority. This demonstrates DDF's adaptive nature: the suggestion space evolves with the experimental trajectory rather than rigidly repeating prior recommendations.

\subsection{Post-Baseline Optimization Mode}

When Trial~3 in Cycle~1 beat the baseline (Dice = 0.7682), DDF switched from failure diagnosis to optimization mode. Instead of generating five corrective suggestions, it performed a systematic proposal--implementation completeness audit:

\begin{table}[h]
    \centering
    \small
    \setlength{\tabcolsep}{4pt}
    \caption{DDF optimization-mode audit after Trial~3 beats the baseline (Dice = 0.7682).}
    \label{tab:ddf-optimization}
    \begin{tabular}{l l p{7cm}}
    \toprule
    \textbf{Module} & \textbf{Status} & \textbf{Description} \\
    \midrule
    HTD Tokenization & \textsc{Simplified} & Uses \texttt{AdaptiveAvgPool2d} instead of strided projection \\
    HTD Hierarchy Embedding & \textsc{Simplified} & Additive vector vs.\ conditional attention bias \\
    SADCB Gating & \textsc{Simplified} & Softmax over branches vs.\ per-channel adaptive gating \\
    \bottomrule
    \end{tabular}
\end{table}

The audit generated prioritized optimization suggestions: (1)~refactor HTD to use proper Patch Partitioning instead of \texttt{AdaptiveAvgPool2d}, (2)~modify HTD to use Hierarchy Embedding as context token (K,V) not additive bias, (3)~update SADCB to use channel-wise gating ($\alpha_i$) for fusing dilated branches, and (4)~increase embedding dimension and attention heads. This dual-mode design ensures that DDF contributes to exploration even when the system is already succeeding.

\section{Complete Experimental Trajectory for Dataset 16}
\label{app:full-trajectory-dataset16}

Table~\ref{tab:lrm-example} in the main text presents a curated 12-row excerpt of the Dataset~16 experimental trajectory. Table~\ref{tab:full-trajectory-dataset16} provides the complete record of all 33 trials across three research cycles, including the losing agent's Dice score at each trial. This extended table illustrates the full dynamics of Layered Reflective Memory, Divergent Diagnostic Feedback, and the dual-agent competition mechanism.

\begin{table*}[t]
    \centering
    \scriptsize
    \setlength{\tabcolsep}{2pt}
    \renewcommand{\arraystretch}{0.65}
    \caption{Complete experimental trajectory for Dataset~16 (optic disc segmentation, UKAN baseline Dice\,=\,0.7142). \textbf{W./L.~Dice}: winning/losing agent's Dice. $\star$\,=\,cycle-best. Global memory at cycle transitions shown in italics.}
    \label{tab:full-trajectory-dataset16}
    \resizebox{\textwidth}{!}{%
    \begin{tabular}{c c r r l r p{9cm}}
    \toprule
    \textbf{Cycle} & \textbf{Trial} & \textbf{W.~Dice} & \textbf{L.~Dice} & \textbf{Status} & \textbf{Diagnostic} & \textbf{Key Modification} \\
    \midrule
    \multicolumn{7}{l}{\textit{Cycle 1: HCP-Net --- Hierarchical Token Diffusion replacing skip connections}} \\
    \multicolumn{7}{l}{\textit{Seeded from: Baseline (UKAN, Dice = 0.7142)}} \\
    \midrule
    1 &  1 & 0.7142 & --- & Baseline & --- & Precomputed UKAN baseline. \\
    1 &  2 & 0.6835 & Err & Underperf. & code\_issue & HTD with patch embedding + cross-attention; hierarchy embedding omitted, spatial pooling too aggressive ($32{\times}32$). \\
    1 &  3$\star$ & \textbf{0.7682} & 0.6753 & Success & --- & HTD at higher resolution ($64{\times}64$); learnable hierarchy embeddings as query bias in cross-attention. \\
    1 &  4 & 0.6916 & 0.6053 & Underperf. & regression & Moved hierarchy embedding from query to key; mathematical formulation change degraded performance. \\
    1 &  5 & 0.6913 & 0.6796 & Underperf. & regression & Replaced self-attention with cross-attention in HTD; new CrossAttentionBlock class introduced. \\
    1 &  6 & 0.7610 & 0.6959 & Success & --- & Reverted from cross-attention to query-bias approach (guided by Trial~3's success in cycle memory). \\
    1 &  7 & 0.6976 & Err & Underperf. & regression & Added Edge-Aware Refinement Module (Sobel-based edge detection) targeting HD95; hurt Dice. \\
    1 &  8 & 0.6452 & 0.5750 & Underperf. & regression & Replaced BatchNorm$\to$GroupNorm, ReLU$\to$GELU; over-regularization degraded performance. \\
    1 &  9 & 0.7510 & 0.7240 & Success & --- & Reverted to query-bias HTD (dropping edge-aware module); recovered near Trial~3. \\
    1 & 10 & 0.7175 & 0.5595 & Underperf. & regression & Multi-scale Boundary Enhancement Module ($3{\times}3$ and $5{\times}5$ kernels); still degraded Dice. \\
    1 & 11 & 0.7444 & 0.7065 & Underperf. & diminishing & Re-implemented query-bias HTD; close to Trial~3/6 but no further improvement. \\
    \midrule
    \multicolumn{7}{l}{\textit{Cycle 2: BAFNet --- Boundary-Guided Instance Normalization + Selective Cross-Scale Fusion}} \\
    \multicolumn{7}{l}{\textit{Seeded from: Cycle 1 best (Trial 3, Dice = 0.7682). Global memory: ``HTD with query-bias works; boundary modules hurt Dice.''}} \\
    \midrule
    2 &  1 & 0.7142 & --- & Baseline & --- & Precomputed UKAN baseline. \\
    2 &  2 & 0.5653 & 0.5573 & Underperf. & code\_issue & BGIN reduces spatial boundary map to single scalar; SCSF simplified to basic conv. \\
    2 &  3 & 0.7290 & 0.6766 & Underperf. & --- & BGIN preserves spatial resolution; SCSF with dual-pooling gating mechanism. \\
    2 &  4 & 0.7386 & 0.5570 & Success & --- & Added residual connection ($1{\times}1$ conv) to BGIN for gradient flow; dropout(0.1) in bottleneck. \\
    2 &  5 & 0.6787 & 0.6608 & Underperf. & regression & Reverted learning rate from 0.001 back to 0.01; regression. \\
    2 &  6 & 0.6537 & 0.6471 & Underperf. & regression & Introduced MultiScaleBoundaryAttention (pyramidal multi-scale detection); over-complex, degraded. \\
    2 &  7 & 0.5626 & Err & Underperf. & regression & Removed MultiScaleBoundaryAttention entirely; over-simplified BGIN to basic convolution. \\
    2 &  8$\star$ & \textbf{0.7829} & 0.6241 & Success & --- & Fixed NaN loss: $\epsilon$: $10^{-5}{\to}10^{-8}$ in BGIN; reduced learning rate. Numerical stability was the key. \\
    2 &  9 & 0.7413 & 0.7086 & Underperf. & regression & Added multi-scale boundary detection ($3{\times}3$ + $5{\times}5$ kernels) to BGIN; regression from Trial~8. \\
    2 & 10 & 0.6902 & 0.5959 & Underperf. & regression & Simplified to EfficientBoundaryAttention (single-scale depthwise); worse than Trial~8. \\
    2 & 11 & 0.7439 & 0.6576 & Underperf. & diminishing & Learning rate $0.002{\to}0.001$ based on Trial~8 analysis; slight regression. \\
    \midrule
    \multicolumn{7}{l}{\textit{Cycle 3: ARTNet --- Adaptive Resolution Tokenization + Context-Aware Token Aggregation}} \\
    \multicolumn{7}{l}{\textit{Seeded from: Cycle 2 best (Trial 8, Dice = 0.7829). Global memory: ``Module simplification is recurring root cause;}} \\
    \multicolumn{7}{l}{\textit{boundary add-ons hurt; numerical stability critical; faithful implementation $>$ complex additions.''}} \\
    \midrule
    3 &  1 & 0.7142 & --- & Baseline & --- & Precomputed UKAN baseline. \\
    3 &  2 & 0.6095 & Err & Underperf. & code\_issue & RATF implemented as fixed-grid pooling (not adaptive clustering); SizeBiasedSelfAttention placeholder. \\
    3 &  3 & 0.5376 & 0.0000 & Underperf. & code\_issue & RATF as grid-based pooling; token-to-spatial via reshape/interpolate destroys spatial info. Severe regression. \\
    3 &  4 & 0.6602 & Err & Underperf. & code\_issue & Re-implemented RATF with dynamic partitioning; still simplified vs.\ proposal. Partial recovery. \\
    3 &  5 & 0.7988 & 0.0289 & Success & --- & ComplexityEstimator using gradient magnitude; adaptive tokenization working. First significant baseline beat. \\
    3 &  6$\star$ & \textbf{0.8191} & 0.7035 & Success & --- & Minor fix (working directory initialization); architecture from Trial~5 preserved. \textbf{Best overall result.} \\
    3 &  7 & 0.7074 & 0.6360 & Underperf. & regression & Enhanced$\to$EnhancedComplexityEstimator (more complex attention); regression. \\
    3 &  8 & 0.7360 & Err & Underperf. & regression & Increased base\_channels $32{\to}48$ for more capacity; regression from Trial~6. \\
    3 &  9 & 0.7407 & 0.6928 & Underperf. & regression & Fixed channel dimension mismatch in decoder fusion; architecture changes caused regression. \\
    3 & 10 & 0.7033 & 0.6444 & Underperf. & regression & base\_channels $32{\to}48$ + boundary refinement module; both hurt (echoes global memory). \\
    3 & 11 & 0.7086 & 0.6951 & Underperf. & regression & base\_channels $32{\to}48$ with minor config changes; confirms capacity increase not beneficial. \\
    \bottomrule
    \end{tabular}%
    }
\end{table*}

\paragraph{Cross-Cycle Summary.}
Table~\ref{tab:cross-cycle-summary} summarizes the progressive improvement across cycles, showing how each cycle builds on the previous one through artifact relay and global memory.

\begin{table*}[h]
    \centering
    \small
    \setlength{\tabcolsep}{6pt}
    \renewcommand{\arraystretch}{1.3}
    \caption{Cross-cycle summary for Dataset~16. Each cycle explores a distinct proposal, seeded from the previous cycle's best artifact with compressed global memory.}
    \label{tab:cross-cycle-summary}
    \begin{tabular}{c l r r c r p{6.5cm}}
    \toprule
    \textbf{Cycle} & \textbf{Proposal} & \textbf{Best Dice} & $\Delta$\textbf{BL} & \textbf{Best Trial} & \textbf{Trials} & \textbf{Key Insight} \\
    \midrule
    1 & HCP-Net & 0.7682 & +5.40 & 3 & 11 & Token diffusion with query-bias works; boundary modules hurt. \\
    2 & BAFNet  & 0.7829 & +6.87 & 8 & 11 & Boundary-aware normalization works if numerically stable; multi-scale detection hurts. \\
    3 & ARTNet  & 0.8191 & +10.49 & 6 & 11 & Adaptive tokenization via complexity estimation; simplicity $>$ complexity confirmed. \\
    \bottomrule
    \end{tabular}
\end{table*}

The trajectory illustrates three key dynamics. First, the \emph{dual-agent competition} consistently produces large performance spreads (e.g., 0.7682 vs.\ 0.6753 in Cycle~1 Trial~3; 0.7988 vs.\ 0.0289 in Cycle~3 Trial~5), confirming that parallel exploration from the same diagnostic produces genuinely divergent outcomes. Second, the \emph{global memory} demonstrably guides later cycles: Cycle~2 avoids the boundary modules that hurt Cycle~1, and Cycle~3's global memory (``faithful implementation $>$ complex additions'') is validated when Trials~7--11 all regress after adding complexity to the working Trial~6 architecture. Third, the \emph{cycle-level memory} enables within-cycle recovery: in Cycle~1, Trial~6 reverts to the query-bias approach specifically because Trial~3's success is visible in the structured history, and in Cycle~3, the agent finally achieves a faithful complexity estimator (Trial~5) after the structured history classifies Trials~2--4 as \texttt{code\_issue} (implementation shortcut) rather than scientific dead ends.

\section{Manuscript Generation Pipeline Details}
\label{app:manuscript-pipeline-details}

This appendix provides additional implementation details for the manuscript synthesis stage described in Section~\ref{sec:manuscripts}. The pipeline transforms structured experimental evidence into a compiled PDF through six sequential steps: methodology reconciliation, result analysis, figure generation, paper writing, citation management, and automated revision.

\subsection{Methodology Reconciliation}
\label{app:reconciliation}

The reconciliation agent receives two inputs: the original research proposal produced during literature-grounded proposal generation (\S\ref{sec:proposal}) and the Python source code of the best-performing node from Stage~2. The agent performs a clause-by-clause comparison between the proposal description and the implemented code, identifying three categories of discrepancy:

\begin{itemize}[leftmargin=1.5em, itemsep=2pt]
    \item \textbf{Omitted components}: modules that appear in the proposal but are absent from the final implementation, typically because they were dropped during the iterative search to resolve implementation errors.
    \item \textbf{Emergent additions}: components present in the code that were not specified in the original proposal, introduced during the refinement process to address issues discovered empirically.
    \item \textbf{Modified components}: modules whose implementation differs from the proposal specification, such as a proposed multi-head cross-attention mechanism that was simplified to a single-head variant during debugging.
\end{itemize}

The reconciliation output is a revised methodology document that describes only what was actually implemented. The agent preserves the original narrative structure, motivation, and contribution framing whenever they remain compatible with the implementation, and updates only the implementation-sensitive details: module names and roles, architectural components, kernel sizes, attention dimensions, channel counts, and similar design parameters. Theoretical proposal details that were not realized in the final implementation are replaced with the actual implemented version rather than retained alongside it. Training recipe details (optimizer, scheduler, batch size, learning rate policy) are excluded from the methodology output, as these belong exclusively in the experimental setup section of the paper.

\subsection{Result Analysis Protocol}
\label{app:analysis-protocol}

The analysis agent operates in two steps. In the first step, the agent receives the research proposal and the full experimental results (including the metric tables from all Stage~2 trials and the Stage~3 ablation record) and produces a structured analysis report. This report contains a ranked list of the top-performing configurations with their Dice and HD95 scores, a pairwise comparison between the best configuration and the baseline on both metrics, an ablation summary identifying which modules contribute most to performance and which have negligible effect, and an interpretation of the observed patterns in terms of the proposed methodology.

In the second step, the agent generates a figure plan that specifies which visualizations should accompany the paper. A typical plan includes a bar chart comparing the proposed method against the baseline and ablation variants on the primary metric, a grouped bar chart showing per-class Dice scores when multi-class segmentation is involved, and qualitative segmentation overlays showing representative cases where the proposed method improves upon the baseline. The figure plan is passed to the visualization agent, which is responsible for generating the actual images.

\subsection{Figure Generation}
\label{app:figure-generation}

Figure generation proceeds in two phases. Phase~1 generates result figures before the paper text exists. This phase produces two types of output. The first type is deterministic segmentation comparison visualizations: for each test case, the system renders the ground truth annotation and the predicted segmentation mask of the best model side by side, selecting representative cases that illustrate both successful segmentation and residual failure modes. These visualizations are generated directly from the model predictions stored in the experiment directory, without involving a language model. The second type is data-driven analysis plots: an agent-based code generation system receives the experimental results, the ablation data, and the figure plan from the analysis step, writes matplotlib scripts, and executes them in a sandboxed environment through the OpenHands framework. The generated plots include metric comparison bar charts, ablation contribution charts, and any additional visualizations specified in the figure plan. Each generated image undergoes a size and quality check; images below 12~KB are flagged as likely placeholders and regenerated.

Phase~2 generates method diagrams after the paper text has been drafted. The system extracts the Methods section from the generated LaTeX and passes it to an image generation model (Gemini~3.1 Flash Image Preview) that produces a main architecture figure and, optionally, sub-figures for individual modules. The main figure is inserted before the Methods section heading, while sub-figures are inserted before their corresponding subsection headings. This two-phase design ensures that result figures are grounded in actual experimental data, while method diagrams are consistent with the notation and structure of the written methodology.

\subsection{Paper Writing}
\label{app:paper-writing}

The writing agent uses a template-driven approach to generate the paper. The template defines the section structure, the generation mode for each section, and the target length. The system uses an Elsevier single-column format with six predefined sections: Introduction, Related Work, Method, Experiments, Discussion, and Conclusion. Each section is generated through one of three modes depending on its nature:

\begin{itemize}[leftmargin=1.5em, itemsep=2pt]
    \item \textbf{Direct generation}: The Introduction and Conclusion sections are produced in a single pass from the research proposal, dataset context, and analysis report.
    \item \textbf{Hierarchical generation}: The Related Work and Method sections are generated in three steps: (1)~plan the subsection structure (1--2 subsections for Related Work, 3--6 for Method), (2)~generate each subsection independently, and (3)~apply a review agent that checks for consistency, redundancy, and logical flow across subsections.
    \item \textbf{Structured generation}: The Experiments section uses three fixed subsections (Datasets and Implementation Details, Ablation Studies, State-of-the-Art Comparisons), each generated from a dedicated prompt template that receives the corresponding evidence slice.
\end{itemize}

The writing agent is conditioned on the dataset context to produce domain-appropriate clinical terminology. It is instructed to report all numerical results to the precision present in the evaluation metrics, to reference all tables and figures, and to avoid making claims not supported by the experimental evidence. Each section is generated with awareness of all previously generated sections (provided as context), ensuring that terminology, notation, and cross-references remain consistent across the paper.

\subsection{Citation Management}
\label{app:citation-management}

The citation agent processes the manuscript in a single pass to identify and resolve citation needs. The agent first scans the LaTeX text for citation placeholders (marked during writing as \texttt{[CITE:keyword]} tokens) and for sentences that contain technical claims requiring references. For each identified citation need, the agent extracts a context window of 200 characters around the placeholder to determine the precise meaning of the required reference.

The agent then queries Semantic Scholar with keyword-based searches derived from each citation context. The search scope includes papers published from 2000 onward to cover both foundational methods (such as U-Net) and recent advances. For each query, the agent retrieves candidate papers and uses a language model to verify that the top candidate matches the intended reference by comparing its title and abstract against the citation context. If the initial query does not return a satisfactory match, the agent generates a refined query incorporating additional context from the surrounding sentences and retries. This verification step prevents the common failure mode of inserting a superficially related but incorrect reference.

Citation entries are deduplicated by DOI and title matching before the final bibliography is assembled. Across the 40 manuscripts produced in our experiments, the citation agent inserts a median of 24 references per paper (range: 14 to 41), with a manual verification finding fewer than 5\% of inserted citations to be tangentially related to the intended claim.

\subsection{Automated Revision}
\label{app:revision}

The manuscript undergoes two rounds of automated revision, each implemented as an agent-based editing loop using the OpenHands framework. In each round, the agent receives the full LaTeX source file together with a structured task prompt, edits the file in place, and the system compiles the result. If compilation fails, the compilation error log is sent back to the agent for correction. This compile-fix loop repeats up to four times per round.

The first round targets structural issues. The editing agent checks for duplicate section or subsection headings, paragraphs whose content is repeated across sections, and inconsistencies between the method description and the experimental setup (such as conflicting hyperparameter values or architecture names). The agent also verifies that training recipe details (optimizer, scheduler, batch size, epoch count, learning rate policy, hardware specification, and train/test split) appear only in the implementation details subsection and not in the methodology section. When duplicates or inconsistencies are found, the agent retains the more complete or more accurate version and removes the redundant material.

The second round targets stylistic issues characteristic of language-model-generated academic text. The editing agent operates with a curated list of over 80 words and phrases that are statistically overrepresented in LLM output relative to human-written academic prose (examples include ``delve,'' ``leverage,'' ``tapestry,'' ``it is worth noting that,'' and ``first and foremost''). The agent replaces these with plain, precise alternatives. It also converts itemized and enumerated lists in running text into coherent prose paragraphs, removes mechanical transition phrases, and reduces excessive use of emphasis formatting (\texttt{\textbackslash textbf\{\}} and \texttt{\textbackslash textit\{\}} used for rhetorical emphasis rather than technical notation). The agent is instructed to leave sentences unchanged when they already read naturally, applying the modification threshold that edits should improve readability rather than merely swap synonyms.

\subsection{Output Artifacts}
\label{app:output-artifacts}

The complete output of the manuscript stage consists of five categories of artifacts: (1)~the final compiled PDF, ready for submission or review; (2)~the complete LaTeX source project, including the main source file, bibliography, and document class file; (3)~all generated figures in both PDF and PNG formats, organized into separate directories for method diagrams and result plots; (4)~an Overleaf-compatible zip package that can be uploaded directly to a collaborative editing platform; and (5)~the intermediate files from each pipeline stage (raw dataset context, original and reconciled research proposal, analysis report, figure plan, draft and revised LaTeX versions, and compilation logs), preserving the complete provenance chain from experimental evidence to final manuscript.

\end{document}